\documentclass{article} 
\usepackage{iclr2025_conference,times}

\usepackage{amsmath,amsfonts,bm}

\def\eqref#1{equation~\ref{#1}}

\def\1{\bm{1}}

\DeclareMathAlphabet{\mathsfit}{\encodingdefault}{\sfdefault}{m}{sl}
\SetMathAlphabet{\mathsfit}{bold}{\encodingdefault}{\sfdefault}{bx}{n}

\usepackage{graphicx}
\usepackage{booktabs}
\usepackage{comment}

\usepackage[pagebackref,breaklinks,colorlinks,citecolor=blue]{hyperref}
\usepackage{cleveref}

\usepackage{times}
\usepackage{algorithmic}
\usepackage{latexsym}

\usepackage{url}

\usepackage{floatrow}

\usepackage{bm}
\usepackage[utf8]{inputenc} 
\usepackage{amsfonts}       
\usepackage{amsmath}
\usepackage{amssymb}
\usepackage{booktabs}
\usepackage[graphicx]{realboxes}
\usepackage{tikz}
\usepackage{tabto}
\usetikzlibrary{positioning}
\usetikzlibrary{calc}
\usetikzlibrary{matrix}
\usepackage{overpic}
\usepackage{nicefrac}       
\usepackage{bbm}
\usepackage{enumitem}
\usepackage{multirow}
\usepackage{caption}
\usepackage{xcolor,colortbl}
\usepackage{subcaption}
\usepackage{verbatim}
\usepackage{tcolorbox}
\usepackage[final]{listings}
\usepackage{xspace}
\usepackage{wrapfig}
\usepackage{adjustbox} 
\usepackage[symbol]{footmisc}

\usepackage{rotating}
\usepackage{tabularx}

\usepackage{pifont}
\newcommand{\cmark}{\ding{51}}%
\newcommand{\xmark}{\ding{55}}%
\usepackage{arydshln}
\usepackage{titletoc}
\definecolor{lightblue}{rgb}{0.95, 1., 1.}

\usepackage[framemethod=tikz]{mdframed}
\mdfdefinestyle{mystyle}{%
  linecolor=black!80,
  fontcolor=black!,
  leftline=true,
  innerleftmargin=3,
  innerrightmargin=0,
  outerlinewidth=2pt,
  topline=false,
  rightline=false,
  bottomline=false,
  skipabove=\topsep,
  skipbelow=\topsep,
  backgroundcolor=lightblue
}

\usepackage[normalem]{ulem} 
\newcommand{\eg}{e.g.}
\newcommand{\ie}{i.e.}

\newfloatcommand{capbtabbox}{table}[][\FBwidth]

\definecolor{OliveGreen}{rgb}{0.33,0.42,0.184}
\definecolor{ForestGreen}{RGB}{34,139,34}
\definecolor{Green}{rgb}{0,1.,0.}
\definecolor{Red}{rgb}{1.,0.,0.}
\definecolor{Yellow}{rgb}{1,1,0.}
\definecolor{codegreen}{rgb}{0,0.6,0}
\definecolor{codegray}{rgb}{0.5,0.5,0.5}
\definecolor{codepurple}{rgb}{0.58,0,0.82}
\definecolor{backcolour}{rgb}{0.95,0.95,0.92}

\newcommand{\Table}[1]{Table~\ref{#1}}

\newcommand{\Figure}[1]{Fig.~\ref{#1}}
\newcommand{\Listing}[1]{Listing~\ref{#1}}
\newcommand{\Equation}[1]{Eq.~(\ref{#1})}

\newcommand{\Section}[1]{Sec.~\ref{#1}}
\newcommand{\Appendix}[1]{App.~\ref{#1}}
\newcommand{\ignore}[1]{}
\newcommand{\model}{Gecko}

\newcommand{\pointscore}{point-wise instance scoring}
\newcommand{\pairscore}{pair-wise instance scoring}
\newcommand{\paircompare}{model ordering}
\newcommand{\Pointscore}{Point-wise instance scoring}
\newcommand{\Pairscore}{Pair-wise instance scoring}
\newcommand{\Paircompare}{Model ordering}

\newcolumntype{P}[1]{>{\centering\arraybackslash}p{#1}}

\title{Formatting Instructions for ICLR 2025 \\ Conference Submissions}

\title{Revisiting text-to-image evaluation with Gecko: on metrics, prompts and human rating}

\author{
    Olivia Wiles{\hypersetup{hidelinks}\thanks{Equal contribution. Correspondence to: \textcolor{blue}{oawiles@google.com}; \textcolor{blue}{nematzadeh@google.com.}} \textsuperscript{ ,$\dagger$}} 
    \And Chuhan Zhang\textsuperscript{*,$\dagger$} 
    \And Isabela Albuquerque\textsuperscript{*,$\dagger$} 
    \And Ivana Kaji\'c\textsuperscript{$\dagger$} 
    \And Su Wang\textsuperscript{$\dagger$} \\
    \And Emanuele Bugliarello\textsuperscript{$\dagger$} 
    \And Yasumasa Onoe\textsuperscript{$\dagger$} 
    \And Pinelopi Papalampidi\textsuperscript{$\dagger$} 
    \And Ira Ktena\textsuperscript{$\dagger$} 
    \And Chris Knutsen\textsuperscript{$\dagger$} \\
    \And Cyrus Rashtchian\textsuperscript{$\ddagger$} 
    \And Anant Nawalgaria\textsuperscript{\S} 
    \And Jordi Pont-Tuset{\hypersetup{hidelinks}\thanks{Google DeepMind, \textsuperscript{$\ddagger$}Google Research, \textsuperscript{\S}Google Cloud. \\ Github link: \url{https://github.com/google-deepmind/gecko_benchmark_t2i}}} 
    \And Aida Nematzadeh\textsuperscript{$\dagger$}
}

 \iclrfinalcopy 

\begin{document}

\maketitle

\begin{abstract}
\vspace{-1mm}


While text-to-image (T2I) generative models have become ubiquitous, they do not necessarily generate images that align with a given prompt. 
While many metrics and benchmarks have been proposed to evaluate T2I models for alignment, the impact of the evaluation components (prompt sets, human annotations, evaluation task) has not been systematically measured.
We find that looking at only \textit{one slice of data}, \ie~one set of skills or human annotations, is not enough to obtain stable conclusions that generalise to new conditions or slices when evaluating T2I models or alignment metrics. 
We address this by introducing an evaluation suite of $>$100K annotations across four human annotation templates that comprehensively evaluates models' capabilities across a a range of methods for gathering human annotations and comparing models.
In particular, we propose (1) a carefully curated set of prompts -- {\em Gecko2K}; (2) a statistically grounded method of comparing T2I models; and (3) a framework to systematically evaluate metrics under three {\em evaluation tasks} -- {\em \paircompare, \pairscore, \pointscore}.
Using this evaluation suite, we compare a wide range of metrics and find that a given metric may do better in one setting but worse in another.
As a result, we introduce a new, interpretable auto-eval metric that is consistently better correlated with human ratings than existing ones on our evaluation suite--across different human templates and evaluation settings--and on TIFA160. 
\end{abstract}

\section{Introduction}
\label{sec:intro}



Text-to-image (T2I) models \citep{saharia2022photorealistic,yu2022scaling,betker2023improving,rombach2022high} generate images of impressive quality, but the  images are not necessarily aligned with the prompt.
The key to comparing T2I models is in the dataset of prompts and human annotations we collect.
Human annotation is slow and expensive, motivating the creation~\citep{hu2023tifa, cho2023davidsonian} of automatic-evaluation (auto-eval) metrics as a replacement.
To evaluate both metrics and models, human annotation is the gold standard.
However, \cite{clark-etal-2021-thats} show that the template design and annotator knowledge can significantly impact results in the text domain. 
In this work, we create a comprehensive benchmark to answer the question: {\em how do the choices around prompts and human annotation templates impact our metric and modelling decisions?}

\begin{figure}[t]
    \scriptsize
    \begin{tabular}{ccccc}
         & {\bf Likert} & {\bf Word Level} & {\bf DSG(H)} & {\bf Side-by-Side} \\ \toprule
        {\bf Model 1 (M1)} \\
        \adjustbox{valign=c}{\includegraphics[width=0.14\linewidth,height=0.14\linewidth]{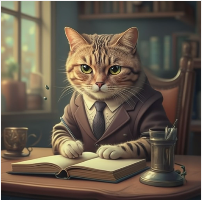}} & \adjustbox{valign=c}{1-2-{\bf \underline 3}-4-5} & \begin{minipage}{3.5cm} {\scriptsize \colorbox{green!30}{A} \colorbox{red!30}{cartoon} \colorbox{green!30}{cat} \colorbox{green!30}{in} \colorbox{green!30}{a} \colorbox{green!30}{professor} \colorbox{green!30}{outfit}, \colorbox{green!30}{writing} \colorbox{green!30}{a} \colorbox{green!30}{book} \colorbox{yellow!30}{with} \colorbox{yellow!30}{the}  \colorbox{yellow!30}{title}  ``\colorbox{yellow!30}{what}  \colorbox{yellow!30}{if}  \colorbox{yellow!30}{a}  \colorbox{yellow!30}{cat}  \colorbox{yellow!30}{wrote}  \colorbox{yellow!30}{a}  \colorbox{yellow!30}{book}.''} \end{minipage}  & \begin{minipage}{3.8cm}{\scriptsize Q1: Is there a cat?
        {\color{OliveGreen} \small \ding{52}} \\
        Q2: Is the cat a cartoon? 
        {\color{red} \small \xmark} \\
        Q3: Is the cat in a professor outfit?
        {\color{OliveGreen} \small \ding{52}}  \\
        Q4: Is the cat writing a book? 
        {\color{OliveGreen} \small \ding{52}}  \\
        Q5: Is the book title “what if a cat
        wrote a book?” 
        {\color{red} \small \xmark} \\

        } \end{minipage} & \adjustbox{valign=c}{\large =} \\
        {\bf Model 2 (M2)} \\
        \begin{minipage}{.14\textwidth}
        \includegraphics[width=\linewidth]{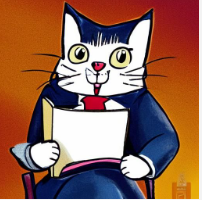} \end{minipage}
        & 1-2-3-{\bf \underline 4}-5 & \begin{minipage}{3.5cm} {\scriptsize \colorbox{green!30}{A} \colorbox{green!30}{cartoon} \colorbox{green!30}{cat} \colorbox{green!30}{in} \colorbox{green!30}{a} \colorbox{green!30}{professor} \colorbox{green!30}{outfit}, \colorbox{red!30}{writing} \colorbox{green!30}{a} \colorbox{green!30}{book} \colorbox{red!30}{with} \colorbox{red!30}{the}  \colorbox{red!30}{title}  ``\colorbox{red!30}{what}  \colorbox{red!30}{if}  \colorbox{red!30}{a}  \colorbox{red!30}{cat}  \colorbox{red!30}{wrote}  \colorbox{red!30}{a}  \colorbox{red!30}{book}.''} \end{minipage}  & \begin{minipage}{3.8cm}{\scriptsize Q1: Is there a cat?
        {\color{OliveGreen} \small \ding{52}} \\
        Q2: Is the cat a cartoon? 
        {\color{red} \small \xmark} \\
        Q3: Is the cat in a professor outfit?
        {\color{OliveGreen} \small \ding{52}}  \\
        Q4: Is the cat writing a book? 
        {\color{OliveGreen} \small \ding{52}}  \\
        Q5: Is the book title “what if a cat
        wrote a book?” 
        {\color{red} \small \xmark} \\

        } \end{minipage} & \adjustbox{valign=c}{\large =}  \\ \midrule
        {\bf Model ordering} & M1 {\color{red}{$<$}} M2 & M1 {\color{red}{$>$}} M2 & M1 {\color{red}{$=$}} M2 & M1 {\color{red}{$=$}} M2
        \\ \bottomrule
    \end{tabular}
    \caption{{\bf Model ordering outcomes for one annotation template do not necessarily generalise to other templates.} We generate images for two models using the prompt: {\color{blue} \em A cartoon cat in a professor outfit, writing a book with the title “what if a cat wrote a book.”}. By collecting extensive human evaluation, we expose disparities across templates: outcomes between T2I models or auto-eval metric obtained for one template  may not generalise to others.} 
    \label{fig:overview}
    \vspace{-0.5em}
\end{figure}

There has been limited work analysing the impact on model and metric ranking due to these choices.
Previous work builds a benchmark by collecting annotations across {\em one} template and prompts that cover a limited distribution of skills (see \Table{tab:humanevalcomparisons} for a comparison).
A skill refers to a generation challenge, such as text rendering or generating different colors and shapes and a sub-skill refers to sub-challenges (e.g.~generating longer text or Gibberish).
Other work does not systematically gather prompts to ensure a wide coverage of skills and properties with the exception of \cite{zhu2023contrastive}. 
\cite{cho2023davidsonian} does consider different lengths of prompts and \cite{zhu2023contrastive} some skills but this is done for a specific template and does not consider varied challenges for a given skill.

By looking at {\em one slice of data} (\eg, too few or too specific prompts or one human annotation template), we are at risk of drawing conclusions that are {\em specific to that slice and do not generalise}.
As a result, we collect a comprehensive dataset (\Table{tab:humanevalcomparisons}) systematically over different prompt sources and different skills/subskills.
Given this prompt set, we generate images from four T2I models and rate them across four human annotation templates.
By considering model rankings across annotation templates for a given prompt, we can further determine how reliably a prompt measures alignment.


Similarly, the choice of task may impact our results. 
Auto-eval metrics are typically evaluated using correlation with human judgement. 
However, in practice, we would want to use metrics for three tasks: (1) \paircompare, ranking T2I models based on {\em significant} relationships; (2) \pairscore, choosing whether a given output is better than another and (3) \pointscore, an estimation of a samples' overall alignment. Evaluating metrics on one task is not enough: we may {\em think} we are choosing the best metric for all three but we show that conclusions for one task {\em do not necessarily generalise}. An overview of our contributions follows:

\begin{itemize}
    \item {\em Gecko}: An evaluation suite for T2I alignment which includes a comprehensive set of 2K prompts, 4 human templates to evaluate 4 T2I models to give $\sim$100K human annotations (\Table{tab:humanevalcomparisons}). We get predictions from a wide range of auto-eval metrics and evaluate under 3 realistic settings (\paircompare, \pairscore, \pointscore).
    \item Using our suite, we demonstrate limitations of looking at a single slice of data as currently done in the literature: different metrics and models show different results depending on the prompt slice or template. 
    \item Based on our analyses, we introduce an interpretable state-of-the-art QA/VQA metric. It gets the most number of model comparisons right, and performs on average 40.5\%/22\% better than interpretable baselines on our dataset in terms of \pairscore~and \pointscore~respectively, and 10.5\% better on TIFA160 \citep{hu2023tifa}.
\end{itemize}

\begin{table}[t]
\scriptsize
  \resizebox{\textwidth}{!}{  
\centering
\begin{tabular}{lP{1cm}P{1cm}P{1cm}P{1cm}|P{1cm}cc|P{1.5cm}cc}
         & Likert & Word Level & DSG(H)  & SxS & \# prompts annotated  & $\frac{\texttt{\# anns}}{\texttt{\# img}}$ & \# anns  & \# Skills (All Categories) & \# Sub-Skills & \\ \hline
    DSG1K \citep{cho2023davidsonian} & \textcolor{red}{\xmark} &  \textcolor{red}{\xmark} & \textcolor{OliveGreen} {\bf \checkmark} & \textcolor{red}{\xmark} & 1.06K &  3 & 9.6K & 11(13) &\textcolor{red}{\xmark}  \\
    DrawBench \citep{saharia2022photorealistic} & \textcolor{red}{\xmark} & \textcolor{red}{\xmark} & \textcolor{red}{\xmark} & \textcolor{OliveGreen} {\bf \checkmark} & 200 & {\bf 25} & 25K & 7(11) & \textcolor{red}{\xmark} \\
    PartiP. \citep{yu2022scaling} & \textcolor{red}{\xmark} &  \textcolor{red}{\xmark} & \textcolor{red}{\xmark} & \textcolor{OliveGreen} {\bf \checkmark} & \underline{1.6K} & 5 & 16K & 9(23) & \textcolor{red}{\xmark} \\
    TIFA160 \citep{hu2023tifa} & \textcolor{OliveGreen} {\bf \checkmark} & \textcolor{red}{\xmark} & \textcolor{red}{\xmark} & \textcolor{red}{\xmark} & 160  & 2 & 1.6K & 8(12) & \textcolor{red}{\xmark} \\
    PaintSkills \citep{cho2023dall} & \textcolor{OliveGreen} {\bf \checkmark} &  \textcolor{red}{\xmark} & \textcolor{red}{\xmark} & \textcolor{red}{\xmark} & 150 & 5 & 2.25K & 3(3) & \textcolor{red}{\xmark} \\
    W-T2I \citep{zhu2023contrastive} & \textcolor{OliveGreen} {\bf \checkmark} &  \textcolor{red}{\xmark} & \textcolor{red}{\xmark} & \textcolor{red}{\xmark} & 200 & 3 & 2.4K & 15(20) & \textcolor{OliveGreen}{\bf 9} \\
    HEIM \citep{lee2024holistic} & \textcolor{OliveGreen} {\bf \checkmark} &  \textcolor{red}{\xmark}  & \textcolor{red}{\xmark} & \textcolor{red}{\xmark} & 708  & $\sim$5.4& $\sim${\bf 150K} & 6(6) & \textcolor{red}{\xmark} \\ \hline
    Gecko2K & \textcolor{OliveGreen} {\bf \checkmark} & \textcolor{OliveGreen} {\bf \checkmark} & \textcolor{OliveGreen} {\bf \checkmark} & \textcolor{OliveGreen} {\bf \checkmark} & {\bf 2K} & $\sim$\underline{13.5} & $\sim$\underline{108K} & 12(12) & \textcolor{OliveGreen}{\bf 36} \\
    \quad Gecko(R) & \textcolor{OliveGreen} {\bf \checkmark} & \textcolor{OliveGreen} {\bf \checkmark} & \textcolor{OliveGreen} {\bf \checkmark} & \textcolor{OliveGreen} {\bf \checkmark} & 1K &  $\sim$\underline{13.5} & $\sim$54K & 11(11) & \textcolor{red}{\xmark} \\
    \quad Gecko(S) & \textcolor{OliveGreen} {\bf \checkmark} & \textcolor{OliveGreen} {\bf \checkmark} & \textcolor{OliveGreen} {\bf \checkmark} & \textcolor{OliveGreen} {\bf \checkmark} & 1K &  $\sim$\underline{13.5} & $\sim$54K & 12(12) & \textcolor{OliveGreen}{\bf 36}
     \\ \hline
\end{tabular}
}
\caption{{\bf Comparison of annotated alignment datasets.} We report the amount of human annotation and skill division for each dataset.  We can see that many datasets include only a handful of annotated prompts or a small number of annotations (anns) per image or overall. No dataset besides Gecko2K collects ratings across multiple different human annotation templates. We also include the number of skills and sub-skills in each dataset. Again, \model~includes the most number of sub-skills, allowing for a fine-grained evaluation of metrics and models. When datasets do not include skills, we map their categories into skills/sub-skills as appropriate.}
\label{tab:humanevalcomparisons}
\vspace{-2mm}
\end{table}

\section{Related Work}

\noindent {\bf Benchmarking alignment in T2I models.}
Many benchmarks have been proposed to holistically evaluate model capabilities within T2I alignment.
Early benchmarks are small scale and created alongside model development to perform side-by-side model comparisons~\citep{saharia2022photorealistic,yu2022scaling,betker2023improving}.
Later work (\eg, TIFA~\citep{hu2023tifa}, DSG1K~\citep{cho2023davidsonian} and HEIM~\citep{lee2024holistic}) focuses on creating holistic benchmarks by drawing from existing datasets (\eg, MSCOCO~\citep{lin2014microsoft}, Localized Narratives \citep{pont2020connecting} and CountBench \citep{paiss2023teaching}) to evaluate a range of capabilities including counting, spatial relationships, and robustness. Other datasets focus on a specific challenge such as compositionality \citep{huang2024t2i}, contrastive reasoning \citep{zhu2023contrastive}, text rendering \citep{tuo2023anytext}, reasoning \citep{cho2023dall}, spatial reasoning \citep{gokhale2022benchmarking}, or specifically image ordering given an increasing number of errors \citep{saxon2024evaluates}.
The Gecko2K benchmark is similar in spirit to TIFA and DSG1K in that it evaluates a set of skills. However, in addition to drawing from previous datasets---which may be biased or poorly representative of the challenges of a particular skill---we collate prompts across sub-skills for each skill to obtain a discriminative prompt set.
Moreover, we gather human annotations across multiple templates and many prompts (see \Table{tab:humanevalcomparisons}).

\noindent {\bf Automatic metrics measuring T2I alignment.}
Inspired by work in image captioning, a widely used auto-eval metric is CLIPScore \citep{hessel2021clipscore}.
However, such metrics poorly capture finer-grained aspects of images~\citep{bugliarello2023measuring,yuksekgonul2022and}. 
Motivated by work in NLP on evaluation  using entailment or QA metrics \citep{maynez2020faithfulness,kryscinski2019evaluating,honovich2021q}, similar metrics~\citep{yarom2024you} have been devised for T2I alignment. 
However, such a metric may not generalise to new settings and is not interpretable---one cannot diagnose why an alignment score is given.
Visual question answering (VQA) methods such as TIFA~\citep{hu2023tifa}, VQ\textsuperscript{2} \cite{yarom2024you} and DSG~\citep{cho2023davidsonian} do not require task-specific finetuning and give an interpretable explanation for their score. These metrics create QA pairs which are then scored with a VLM given an image and aggregated into a single score.
However, the performance of such methods is conditional on the behaviour of the underlying LLMs used for question generation, and VLMs used for answering questions.

\section{{\em Gecko2K:} The Gecko Benchmark}
\label{sec:benchmark}

We curate a fine-grained skill-based benchmark, Gecko2K, with good coverage by curating two sets of prompts: one created systematically based on a set of skills and subskills (Gecko(S)) and one generated by combining existing datasets but tagging them and resampling to ensure good coverage over those tags (Gecko(R)).
We generate Gecko(R) by extending the DSG1K~\citep{cho2023davidsonian} benchmark creation approach to use automatic tagging and improve the distribution of skills and linguistic properties (see \Appendix{app:benchmark_details} for details on the automatic tagging). 
However, due to the automatic tagging and nature of the underlying datasets, Gecko(R) is limited in the skills/sub-skills it covers.
To generate our systematic set, we propose a hierarchical method combined with LLM generation in order to ensure a systematic distribution across skills (\eg, {\em counting}) and subskills (\eg, {\em simple modifier}: `1 cat' vs {\em additive}: `1 cat and 3 dogs').
This notion of sub-skills ensures we are capturing a wide distribution of prompts and not just one easy slice (\eg generating counts of 1-4 objects).

\subsection{Gecko(R): Resampling Davidsonian Scene Graph Benchmark}
The recent DSG1K benchmark \citep{cho2023davidsonian} curates a list of prompts from existing image-text datasets\footnote{TIFA\citep{hu2023tifa}, Stanford Par.\citep{krause2017hierarchical}, Localized Narr.\citep{pont2020connecting}, CountBench \citep{paiss2023teaching}, VRD \citep{lu2016visual}, DiffusionDB \citep{wang2022diffusiondb}, MJ \citep{turc2023midjourney}, PoseScript \citep{delmas2022posescript}, Whoops \citep{bitton2023breaking}, DrawText-Creative \citep{liu2022character}.} but does not control for the coverage or complexity of a given skill. The authors randomly sample 100 prompts and limit the prompt length to 200 characters. 
The resulting dataset is imbalanced in terms of the distribution of skills.
Also, as T2I models take in longer and longer prompts, the dataset will not test models on that capability.
We take a principled approach in creating Gecko(R) by resampling from the base datasets in DSG1K for better coverage and lifting the length limit. After this process, there are 175 prompts longer than 200 characters and a maximum length of 570 characters.
Also, this new dataset has better coverage over a variety of skills than the original DSG1K dataset (see \Figure{fig:dsg1k-resampled}).
\label{sec: Gecko-R}

While resampling improves the distribution of skills, Gecko(R) has the following shortcomings. Due to the limitations of automatic tagging, it does not include all skills we wish to explore  (\eg, language). It also does not include sub-skills: \eg, text rendering prompts do not focus on numerical text, or longer text (see \Figure{fig:text-rendering-visualisation}). Finally, automatic tagging can be error prone.

\subsection{Gecko(S): A Controlled and Diagnostic Prompt Set}
\label{sec: Gecko-S}
The aim of Gecko(S) is to generate prompts in a controllable manner for skills that are not well represented in previous work. We divide skills into sub-skills to diversify the difficulty and content of prompts. We take inspiration from psychology literature where possible (\eg, colour perception) and known limitations of current T2I models.

\begin{wraptable}{t}{0.5\textwidth}    
    \centering
    \vspace{-6mm}
    \tiny
\begin{tabular}{p{1.4cm}p{4cm}} \toprule
    Sub-skill & Example  \\ \midrule
    Numbers / symbols & equation of \textcolor{blue}{"3+4 = 7"} etched into a rock \\
    Length  &  a neon sign with the words \textcolor{blue}{"the future is already here..."} reflected on a rainy street. (n=29) \\
    Gibberish & graffiti made with bright pink paint on the concrete, saying \textcolor{blue}{"fluff floop floof!"} \\
    Typography &  \textcolor{blue}{"i love you"} written in serif font in grass \\ \bottomrule
\end{tabular}
\caption{{\bf Subcategories and corresponding motivations for the text rendering skill.}}
\label{tab:text_rendering_overview}
    \vspace{-1mm}
\end{wraptable}

\noindent {\bf Curating a controlled set of prompts with an LLM.} To generate a set of prompts semi-automatically, we use an LLM. We first decide on the sub-skills we wish to test for. For example, for text rendering, we may want to test for (1) English vs Gibberish to evaluate the model's ability to generate uncommon words, and (2) the length of the text to be generated. We then create a template which conditions the generation on these properties. Note that as we can generate as much data as desired, we can define a distribution over the properties and control the number of examples generated for each sub-skill.
Finally, we run the LLM and manually validate that the prompts are reasonable, fluent, and match the conditioning variables (\eg, the prompt has the right length and is Gibberish / English). A sample template is given in \Appendix{app:templates}.

\noindent {\bf Gecko(S) make up.} Using this approach and also some manual curation, we focus on twelve skills falling into five categories (\Figure{fig:dataset_overview}): (1) \textsc{named entities}; (2) \textsc{text rendering}; (3) \textsc{language/linguistic complexity}; (4) \textsc{relational}: \textsc{action}, \textsc{spatial}, \textsc{scale}; (5) \textsc{attributes}: \textsc{color}, \textsc{count}, \textsc{surfaces (texture/material)}, \textsc{shape}, \textsc{style}. For sub-skills, we give a full breakdown for all skills in \Appendix{app:geckosbreakdown}. In \Table{tab:text_rendering_overview} we give examples of the sub-skills and corresponding prompts for \textsc{text rendering}.
 Using this approach, we get better coverage over the given sub-skills than other datasets (including Gecko(R)) as shown in \Figure{fig:text-rendering-visualisation}.

\section{Comparing Annotation Templates for Modelling}
\label{sec:humanjudgement}

%
We examine  how the choice of human annotation template impacts results when comparing four models: SD1.5 \citep{rombach2022high}, SDXL \citep{podell2023sdxl}, Muse\footnote{Muse is based on the original model, but trained on different data sources.} \citep{chang2023muse}, and Imagen Vermeer \citep{vasconcelos2024greedygrowingenableshighresolution}.
We consider \emph{absolute comparison} templates (\ie, Likert, Word Level from \cite{liang2023rich}, and DSG(H) from \cite{cho2023davidsonian}) which evaluate models individually, and a template for \emph{relative comparison} of two models (side-by-side or SxS).
A high-level visualisation of each template is in  \Figure{fig:overview} and details in \Appendix{app:human_eval_templates}.
We further introduce a principled method to determine significant \paircompare s based on human judgements. 


\subsection{Data Quality}

We validate the reliability of each template and examine if the choice of the template impacts the quality of the collected data. Given the collected human ratings across the three templates, we compute  {inter-annotator agreement} (IAA) for each generative model by measuring Krippendorff’s $\alpha$, $\mathcal{K}_\alpha \in [-1,1]$ \citep{hayes2007answering}, where a value of 1 indicates perfect agreement and 0 chance \citep{zapf2016measuring}. 
Results reported in Table \ref{tab:h_eval_krippendorff_avg_std} show that agreements are all high, with $\alpha>0.5$, except for the Likert---SD1.5 pair for Gecko(R); we conjecture this is due to the lower quality images of SD1.5. Overall we find that fine-grained templates (WL and DSG(H)) are more reliable (\eg, have higher IAA) and WL achieves the highest IAA for the diverse Gecko(R).
We also measure IAA for the SxS template in \Table{tab:sxs_krippendorffs}. We see lower IAA for the SxS template (though still far above chance) compared to the fine-grained ones. The IAA is $<0.5$ for 6 out of 12 model comparisons. It seems, given the same number of annotators, the SxS template is less reliable.

\input{./tables/human_eval_model_comparison_without_rel_prompts}

\noindent {\bf Reliable prompts.}
\label{sec:experiments_reliable}
Upon manual investigation, we find that differences in human ratings across templates can arise when prompts are difficult to judge with respect to alignment (and not due to the choice of the template): for example, when a prompt contains domain specific knowledge such as \textit{``A bottle of Irn-Bru is sitting on a shelf''} or subjective notions such as \textit{``a futuristic sculpture''}. To understand how this impacts our results, we consider a subset of the prompts that achieve high IAA across templates and models. 
For each model and absolute template, we select the prompts for which inter-rater \emph{disagreement}\footnote{Defined as the variance across image, word, and question ratings for Likert, WL and DSG(H) respectively.} is $<50$\% of the maximum \emph{disagreement} observed across all prompts for that model--template pair. The intersection of these prompts across models and templates gives our \emph{reliable prompts}. We additionally remove instances where all Likert ratings are \emph{Unsure} to get 531 and 725 \emph{reliable prompts} for Gecko(R) and Gecko(S), respectively. 
We first validate that using reliable prompts increases IAA on the SxS template (which was not used in the selection process) and find that it increases the average $\mathcal{K}_\alpha$ from 0.45 to 0.47 on Gecko(R), and 0.49 to 0.54 on Gecko(S) (see \Appendix{app:human_eval_add_results} for details).
In the next sections, we demonstrate how this subset of prompts increases agreement among templates, but at the expense of removing some potentially meaningful prompts.

\subsection{Absolute Annotation Templates: Comparing T2I Models}\label{sec:human_eval_comparing_models}
\textbf{Average ratings.} For the absolute annotation templates, previous work compares T2I models by comparing the average ratings across examples. We report these values in \Table{tab:h_eval_krippendorff_avg_std} and find that the chosen prompt set impacts which model is best (\eg~SDXL for Gecko(R) and Muse for Gecko(S)). Moreover, the model with the lowest rating depends on both the prompt set and the template: given Gecko(R), Imagen is worse if using Likert, but SD1.5 is worse if using DSG(H). This highlights the importance of examining models in various conditions. 
When using T2I models in practice, we need to make conclusions about model ordering with high confidence. We argue that this evaluation is not enough: it does not measure if the difference between models is significant, which is particularly important as models start to saturate. As a result, we introduce the \paircompare{}~task.

\textbf{\Paircompare{}.} 
%
We verify the significance of outcomes by performing the Wilcoxon signed-rank test with $p<0.001$. Where results indicate the null-hypothesis is rejected (\ie, the distribution of ratings is significantly different), we can say that one model is better than another. To determine which model is best, we compare the mean values of their ratings. 
In \Figure{fig:h_eval_wilcoxon_no_rel_prompts} we visualise the outcomes for all model pairs across all templates. We see that Muse is not worse than any of the contenders across all templates and prompt sets, except for 2 out of the 12 comparisons involving Muse for Gecko(R); we determine it is the best overall model. In contrast with the results presented in \Table{tab:h_eval_krippendorff_avg_std}, where SDXL is identified as the best model for Gecko(R) across all the templates, we observe that the significance results reveal that Muse and SDXL actually have similar performance, showcasing the importance of determining significance before drawing conclusions. 

\textbf{Reliable prompts.} Constraining Gecko(R) using the reliable subset decreases the number of conflicts between the different templates, but at the potential expense of comparing models on fewer, potentially easier, prompts. Considering the two prompt sets, we observe that when using the synthetic prompts, Gecko(S)-rel, all templates agree in \Figure{fig:h_eval_wilcoxon} in Appendix \ref{app:human_eval_add_results}. We hypothesise this is because the skills (\eg, color or shape), while hard to generate, are easy to evaluate within generation. For Gecko(R)-rel, we see disagreements between templates, where surprisingly, DSG(H) often result in a different relation than the two other templates. We also consider the full prompt-set, Gecko2K-rel, as it better captures the overall use cases of T2I models: we find that there is always a majority agreement, and the two fine-grained templates (WL and DSG(H)) always agree.

\textbf{Results by skill.} 
We explore how human judgements vary by skill and template; average ratings for each absolute template are shown in \Figure{fig:results_by_category_mean_likert}-\Figure{fig:results_by_category_mean_dsg_color} in the appendix.
A lower average ratings per skill across templates indicates how `challenging' a given skill is: we can see that `lang compositional', `lang complexity', `count' and `text' are consistently difficult across templates.





\subsection{Relative Annotation Template: Comparing T2I Models}
\label{sec:experiments_sxs}

\noindent{\bf \Paircompare{}.} 
For the SxS template, a model is considered better if it is chosen as preferred more often than the competitor and the {\em Unsure} rating. To assess statistical significance, we perform a similar procedure as for the absolute annotation templates using binary scores for the ratings: 0 when there was a tie, +1 when a model was preferred by the majority of raters, and -1 otherwise.

We also compare the SxS template with the considered absolute annotation templates by computing the accuracy obtained by each absolute template when predicting the preferred model given by SxS on Gecko2K-rel. Results presented in \Table{tab:sxs_accuracy} in App.\ref{app:human_eval_add_results} show that all absolute annotation templates predict SxS judgements with similar average accuracy of around 70\%, with DSG being the overall best. This shows that, although the results of pairwise model comparisons are the same in many cases for Gecko2K-rel as shown in \Figure{fig:h_eval_wilcoxon}, absolute and side-by-side annotations do not necessarily correspond to the same model ordering at the datapoint level.

\begin{mdframed}[style=mystyle]
{\bf Takeaway 1:} Fine-grained templates (\ie~ones that require multiple annotations per example), WL and DSG(H), yield the highest inter-annotator agreement. 
{\bf Takeaway 2:} All three absolute annotation templates achieve similar, but not perfect, accuracy when predicting relative comparison annotations for each datapoint.
{\bf Takeaway 3:} To compare models reliably, we need to measure the {\em significant model ordering}. Model ordering depends on the human template and prompt set, but some prompt sets lead to consistent agreement across templates (\eg, are {\em discriminative}) such as our skill-based Gecko(S) or the larger, reliable set Gecko2k-Rel.
\end{mdframed}

\setlength{\tabcolsep}{6pt}
\begin{table}[t]
\centering
\renewcommand{\arraystretch}{1.2}
\resizebox{\textwidth}{!}{
\begin{tabular}{c|cccccc|cccccc}
\hline
\multirow{3}{*}{\textbf{Gen. model}} & \multicolumn{6}{c|}{\bf Inter annotator agreement}                                               & \multicolumn{6}{c}{\bf Scores}                                                                                                     \\ \cline{2-13} 
                            & \multicolumn{3}{c|}{\bf Gecko(R)}                                     & \multicolumn{3}{c|}{\bf Gecko(S)} & \multicolumn{3}{c|}{\bf Gecko(R)}                                            & \multicolumn{3}{c}{\bf Gecko(S)}                        \\ \cline{2-13} 
                            & WL               & Likert        & \multicolumn{1}{c|}{DSG(H)}        & WL  & Likert    & DSG(H)     & WL          & Likert              & \multicolumn{1}{c|}{DSG(H)}         & WL          & Likert              & DSG(H)         \\ \hline
Imagen                      & \textbf{0.81}    & 0.64          & \multicolumn{1}{c|}{0.68}          & 0.72    & 0.57  & \textbf{0.75} & 0.74\tiny{$\pm$0.30}    & 0.60\tiny{$\pm$0.22}    & \multicolumn{1}{c|}{0.84\tiny{$\pm$0.18}}    & 0.80\tiny{$\pm$0.24}    & 0.59\tiny{$\pm$0.20}    & 0.78\tiny{$\pm$0.23}    \\
Muse                        & \textbf{0.82}    & 0.78          & \multicolumn{1}{c|}{0.72}          & 0.69    & 0.58  & \textbf{0.72} & 0.84\tiny{$\pm$0.24}    & 0.61\tiny{$\pm$0.25}    & \multicolumn{1}{c|}{0.83\tiny{$\pm$0.22}}    & \textbf{0.88}\tiny{$\pm$0.18} & \textbf{0.63}\tiny{$\pm$0.21} & \textbf{0.84}\tiny{$\pm$0.21} \\
SDXL                        & 0.75             & \textbf{0.76} & \multicolumn{1}{c|}{0.57}          & 0.67    & 0.56  & \textbf{0.70} & \textbf{0.87}\tiny{$\pm$0.19} & \textbf{0.68}\tiny{$\pm$0.22} & \multicolumn{1}{c|}{\textbf{0.86}\tiny{$\pm$0.16}} & 0.80\tiny{$\pm$0.23}    & 0.60\tiny{$\pm$0.21}    & 0.79\tiny{$\pm$0.22}    \\
SD1.5                       & \textbf{0.66}    & 0.36          & \multicolumn{1}{c|}{\textbf{0.66}} & 0.69    & 0.59  & \textbf{0.74} & 0.86\tiny{$\pm$0.16}   & 0.67\tiny{$\pm$0.22}    &  \multicolumn{1}{c|}{0.76\tiny{$\pm$0.23}}   & 0.61\tiny{$\pm$0.33}   & 0.49\tiny{$\pm$0.21}    & 0.68\tiny{$\pm$0.27}    \\ \hline
\end{tabular}}
\vspace{1mm}
\caption{\textbf{Inter-annotator agreement and ratings for all models and templates.} We measure inter-annotator agreement for each human evaluation template with Krippendorff's $\alpha$. Higher values indicate better agreement. We also show the mean and std.~deviation for the annotated judgements of all templates after mapping the ratings to the $[0,1]$ interval, with 1 indicating perfect alignment. }
\label{tab:h_eval_krippendorff_avg_std}
\end{table}

\section{The Gecko Metric}

An auto-eval metric is more useful if it is (1) interpretable---it reports where a model fails in addition to its overall goodness, (2) reference-free--does not require a reference distribution, and (3) modular---can easily leverage better pretrained models for improved performance.
As a result, we focus on improving recent work using a two-stage QA/VQA metric \citep{hu2023tifa,cho2023davidsonian,yarom2024you} that matches this criteria (as opposed to metrics such as VNLI \citep{yarom2024you} and CLIP \citep{radford2021learning}).
However, the QA/VQA pipelines are impacted by the shortcomings of the pretrained models used. In particular, we identify two main limitations of these pipelines and address them: the QA generation is not always {\em grounded} in the prompt as the generated questions might not necessarily cover {\em all} key parts of the prompt and also there might be {\em hallucinated} questions that are not related to the prompt. Moreover, at the VQA stage, the highest scoring answer might still be low probably but is treated as the ``right'' answer---we model this {\em uncertainty} in the VQA responses. Finally, we simplify the previously proposed methods by removing complexities (such as scene graph generation in DSG) and show that our simplified and improved setup is significantly better across the board.



A standard QA setup (\eg, \citet{hu2023tifa}) consists of three  steps: (1) QA generation: prompting an LLM to generate a set of binary question-answer pairs $\{Q_i, A_i\}_{i=1}^N$ on a given T2I text description $T$.  (2) VQA assessment: employing a VQA model to predict answer $\{A'_i\}_{i=1}^N$ for the generated questions given the generated image $I$. (3) Scoring: computing the alignment score by assessing the VQA accuracy using \Equation{eq:vqa_acc}:
\begin{equation}
\vspace{-0.1in}
    \mathit{Alignment}(T, I) = \frac{1}{N}\sum_{i=1}^N \mathbbm{1}[A'_i = A_i].
\label{eq:vqa_acc}
\end{equation}


\noindent {\bf Groundedness: increasing coverage.}
\label{paragrah: coverage}
To ensure the coverage of questions over the key elements in a text sentence $T$, we split the QA generation into two steps. We first prompt the LLM to index the visually groundable words in the sentence. For example, the sentence ``{\em A red colored dog.}'' is transformed into ``{\em A \{1\}[red colored] \{2\}[dog]}.'' Subsequently, using the text with annotated keywords $\{W'_i\}_{i=1}^N$ as input, we prompt the LLM again to generate a QA pair $\{q_i, a_i\}$ for each word labelled $\{w'_i\}$ in an iterative manner (see App. \ref{app:gecko_metric_details} for the prompting details). This two-step process ensures a more comprehensive and controllable QA generation process, particularly for complex or detailed text descriptions where the prompted LLM often selectively generates questions for specific segments of the text while overlooking others.
\label{paragrah:coverage}

\begin{table}[t]
\renewcommand{\arraystretch}{1.1}
\setlength{\tabcolsep}{3pt}

\centering
\resizebox{0.7\textwidth}{!}{%
\begin{tabular}{c|c|cccc|cccc}
\hline
                                  &                                       & \multicolumn{4}{c|}{\textbf{Gecko(R)}}                                                                                                     & \multicolumn{4}{c}{\textbf{Gecko(S)}}                                                                                                                                                                                                                                           \\ \cline{3-10} 
                                  &                                       & WL                          & Likert                      & \multicolumn{1}{c|}{DSG(H)}                      & SxS                         & WL                          & Likert                               & \multicolumn{1}{c|}{DSG(H)}                               & SxS      \\ \cline{3-10} 
\multirow{-3}{*}{\textbf{Metrics}} & \multirow{-3}{*}{\textbf{Zero-shot}} & \multicolumn{3}{c|}{SpearmanR}                                                                               & Acc                         & \multicolumn{3}{c|}{SpearmanR}                                                                                                 & Acc              \\ \hline
{\em Interpretable (QA/VQA)} & & & & \multicolumn{1}{c|}{} & & & & \multicolumn{1}{c|}{} \\
TIFA$_{\texttt{PALM-2/PALI}}$                                & {\color{ForestGreen}\cmark}                                 & 0.26                        & 0.34                        & \multicolumn{1}{c|}{0.28}                        & 41.7                        & 0.39                        & 0.32                                 & \multicolumn{1}{c|}{0.39}                                 & 53.2                             \\
DSG$_{\texttt{PALM-2/PALI}}$                              & {\color{ForestGreen}\cmark}                                 & 0.35                        & 0.47                        & \multicolumn{1}{c|}{0.42}                        & 49.6                        & 0.45                        & 0.45                                 & \multicolumn{1}{c|}{{0.45}}                        & 58.1                                   \\ 
Gecko$_{\texttt{PALM-2/PALI}}$                              & {\color{ForestGreen}\cmark}                                 & \underline{0.41}               & \underline{0.55}               & \multicolumn{1}{c|}{\underline{0.46}}               & \underline{62.1}                        & \underline{0.47}               & \underline{0.52}                        & \multicolumn{1}{c|}{\underline{0.45}}                        & \underline{74.6}                        \\ \hline
Gecko$_{\texttt{Gemini Flash}}$                      & {\color{ForestGreen}\cmark}                                 & \underline{\textbf{0.43}}               & \underline{\textbf{0.58}}               & \multicolumn{1}{c|}{\underline{\textbf{0.48}}}               & \underline{72.2}                        & \underline{\textbf{0.54}}               & \underline{\textbf{0.59}}                        & \multicolumn{1}{c|}{\underline{\textbf{0.56}}}                        & \underline{\textbf{78.8}}       \\ \hline
{\em Uninterpretable (single score)} & & & & \multicolumn{1}{c|}{} & & & & \multicolumn{1}{c|}{} \\

CLIP                               & {\color{ForestGreen}\cmark}                                 & 0.14                        & 0.16                        & \multicolumn{1}{c|}{0.13}                        & 54.4                        & 0.25                        & 0.18                                 & \multicolumn{1}{c|}{0.26}                                 & 67.2                      \\
PyramidCLIP                        & {\color{ForestGreen}\cmark}                                 & 0.26                        & 0.27                        & \multicolumn{1}{c|}{0.26}                        & {64.3}               & 0.22                        & 0.25                                 & \multicolumn{1}{c|}{0.23}                                 & 70.7         \\
VQAScore$_\texttt{Gemini Flash}$                        & {\color{ForestGreen}\cmark}                                 & \underline{0.42}                        & \underline{0.54}                       & \multicolumn{1}{c|}{\underline{0.45}}                        & \underline{\textbf{73.1}}               & \underline{0.51}                        & \underline{0.57}                                 & \multicolumn{1}{c|}{\underline{0.49}}                                 & \underline{76.5}                    \\
{\color[HTML]{9B9B9B} VNLI}        & {\color{red!60} \xmark}          & {\color[HTML]{9B9B9B} 0.37} & {\color[HTML]{9B9B9B} 0.49} & \multicolumn{1}{c|}{{\color[HTML]{9B9B9B} 0.42}} & {\color[HTML]{9B9B9B} 54.4} & {\color[HTML]{9B9B9B} 0.45} & {\color[HTML]{9B9B9B} {0.55}} & \multicolumn{1}{c|}{{\color[HTML]{9B9B9B} {0.45}}} & {\color[HTML]{9B9B9B} 72.7} \\ \hline
\end{tabular}%
}
\caption{\textbf{Correlation between auto-eval metrics and human ratings across annotation templates on Gecko2K.}  With the same backend, \model~outperforms all other QA/VQA metrics across all evaluations and \model~with GeminiFlash performs even better; it performs better or similar to the strongest single-score approach (VQAScore). \small{\textbf{Bold}: Top results. \underline{Underlined}: Top results by category. }}
\label{tab:corr}
\end{table}

\noindent {\bf Groundedness: removing hallucination.}
LLMs can hallucinate~\citep{bang2023multitask,guerreiro2023hallucinations}, leading to the generation of low-quality, unreliable QA pairs. We filter out hallucinated QA pairs by taking inspiration from previous work in NLP \citep{maynez2020faithfulness,kryscinski2019evaluating}: we employ a Natural Language Inference (NLI) model~\citep{honovich2022true} model for measuring the factual consistency between the text $T$ and QA pairs $\{ Q_i, A_i \}$. QA pairs with a consistency score lower than a threshold $r$ are removed, ensuring that the remaining QAs are about the prompt. 

\noindent {\bf Uncertainty: VQA score normalisation.}
Finally, we improve aggregation of scores from the VQA  model.
The reliance on binary judgement---strictly matching $A'_i$ and $A_i$ without considering the predicted probability of $A'_i$---overlooks the inherent uncertainty in the predictions; a VQA model can predict a very similar score for two answers. If we simply take the max, then we lose this notion of uncertainty reflected in the scores. As a result, we normalise the scores as follows, 
\begin{equation}
    \mathit{Alignment}(T, I) = \frac{1}{N}\sum_{i=1}^N \frac{s_{a}}{\sum_i{s_i}},
\label{eq:vqa_norm}
\end{equation}
where the negative log likelihood of answer $A'_i$ is $s_i$ and the correct answer is $A'_a$ with score $s_a$.
\section{Experiments on Auto-Eval Metrics}
\label{sec:experiments}

We evaluate metrics across multiple prompt sets and templates to determine how they fare on the three tasks: (1) Do they give a good numeric measurement of overall alignment -- {\bf \pointscore}; (2) Are they good indicators on side by side comparisons -- {\bf \pairscore}; (3) Can they predict {\bf \paircompare}. 
We demonstrate that the task {\em can} impact rankings but that our \model~metric consistently performs best for Gecko(S)/(R) across tasks and on TIFA160. 
\Appendix{app:intuitive_explanation_tasks} gives a thorough description of each task and intuitive examples for how they differ.
%

 
\subsection{Experimental Setup}
\label{sec:experimentalsetup}
\noindent {\bf Metrics.} We benchmark two types of metrics. First, metrics that give a {\em single score}, including  
contrastive models (CLIP~\citep{radford2021learning}, PyramidCLIP~\citep{gao2022pyramidclip} and 16 variants in \Section{app:all_correlations}); (2) VNLI~\citep{yarom2024you}; and (3) VQAScore \citep{lin2024evaluating}. 
Second, interpretable QA/VQA based methods: TIFA~\citep{hu2023tifa}, DSG~\citep{cho2023davidsonian} and our metric \model.

\noindent {\bf Back-end models.}
For CLIP, we use a ViT-B/32~\citep{dosovitskiy2020image} CLIP model and ViT-B/16~\citep{dosovitskiy2020image} PyramidCLIP model. 
For VQAScore, we use a GeminiFlash \citep{reid2024gemini} backend. 
For all the VQA-based metrics, we use PaLM-2~\citep{anil2023palm} as the LLM and PaLI~\citep{chen2022pali} as the VQA models in all the metrics for fair comparison.
When evaluating the \model~metric, apart from using the LLM and VQA models above, we utilise a T5-11B model from \citet{honovich2022true} for NLI filtering and set the threshold $r$ at 0.005. This threshold was determined by examining QA pairs with NLI probability scores below 0.05. We observed that QAs with scores below 0.005 are typically hallucinations. We re-use the original prompts from TIFA for generating QAs, and add coverage notation to their selected texts as described in \Section{paragrah: coverage}.
We additionally explore how the performance of \pointscore~changes for the \model~metric if we swap out the QA/VQA models for a stronger Gemini Flash model.
Finally, some baseline models are trained with a maximum text input length $L$,\eg~$L_\text{CLIP}=77$ and $L_\text{VNLI}=82$. For these models, we only take the first $L$ tokens from the text as input.


\begin{figure}
\CenterFloatBoxes
\begin{floatrow}
\capbtabbox{%
  \begin{adjustbox}{max width=6cm} 
\begin{tabular}{l|ccc|ccc}
\hline
\multicolumn{1}{c|}{\multirow{3}{*}{\textbf{Metrics}}} & \multicolumn{3}{c|}{\textbf{Gecko(R)}}  & \multicolumn{3}{c}{\textbf{Gecko(S)}} \\ \cline{2-7} 
\multicolumn{1}{c|}{}   & \multicolumn{1}{c}{WL}     & \multicolumn{1}{c}{Likert}   & \multicolumn{1}{c|}{DSG(H)}     & \multicolumn{1}{c}{WL}        & \multicolumn{1}{c}{Likert}   & \multicolumn{1}{c}{DSG(H)}      \\ \cline{2-7} 
\multicolumn{1}{c|}{}                                  & \multicolumn{6}{c}{Pearson}      \\ \hline
TIFA baseline      & 0.21   & 0.32    &  0.25   & 0.39    & 0.32  & 0.39  \\ \hline
\quad + coverage    & 0.28    & 0.34        & 0.32          & 0.41          & 0.33            & 0.40      \\
\quad \quad + VQA score norm    & 0.32        & 0.42       & 0.37     & 0.43        & 0.37    & 0.41      \\
\quad \quad \quad+ NLI filtering                                & \textbf{0.38}  & \textbf{0.51}  & \textbf{0.42} & \textbf{0.46}  & \textbf{0.48} &  \textbf{0.46} \\ \hline
\end{tabular}%
\end{adjustbox}
}{%
\caption{\textbf{Validation of each component of the proposed Gecko metric on Gecko2K.} We evaluate the utility of the three proposed improvements by adding them to the TIFA baseline one by one. They all bring higher correlation with human judgement across the board on Gecko2K. }
\label{tab:gecko_ablation}
}
\ffigbox{%
  \includegraphics[width=0.48\textwidth, trim={0 0.2cm 0 0cm}, clip]{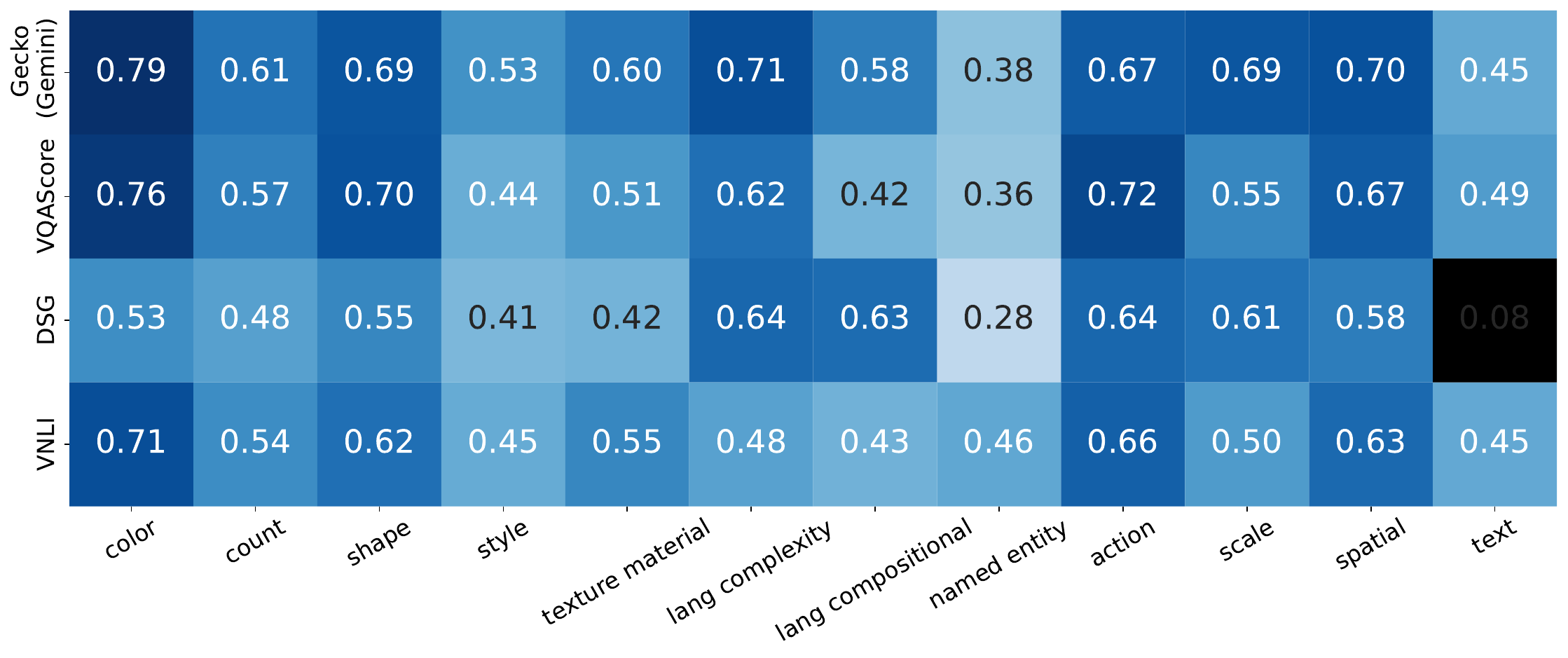}
}{%
  \caption{{\bf Per skill results of different metrics.} Likert correlation for each skill; square black indicates p-values $>0.05$. Full results in \Appendix{app:additional_experimental_results}.}%
    \label{fig:results_by_category_high_level}
}
\end{floatrow}
\end{figure}

\subsection{Comparing Auto-eval Metrics on \Pointscore}
\label{sec:resultsgeckobenchmark}

We first evaluate how well metrics measure  T2I alignment at an instance level. We compute the Pearson and Spearman Ranked correlation between the auto-eval scores and human scores on all the instances in a prompt set. The evaluations are done on the Gecko benchmark and TIFA160.
%

%

\noindent {\bf{Component validation on proposed Gecko metric.}}
\label{sec:ablationstudy}
We validate the utility of the three key improvements we proposed: coverage, linear normalisation, and NLI filtering. Starting from our baseline TIFA, we include the improvements one at a time. The results in \Table{tab:gecko_ablation} uniformly demonstrate a positive impact. NLI filtering brings the largest boost among the three, underscoring the limitation of the PaLM-2 LLM in reliably generating high-quality and accurate QA pairs.


\noindent {\bf Results on Gecko benchmark}
We next compare auto-eval metrics. We start with metrics (CLIP and its variants, TIFA, DSG, \model) that do not rely on fine-tuning.
%
%
As shown in \Table{tab:corr}, the \model~metric outperforms other QA/VQA metrics using the same backend by a wide margin.
Swapping out the backend of \model~with a stronger GeminiFlash model leads to large improvements across the board.
Contrastive models (\eg~CLIP variants) are worse than QA-based metrics, but VQAScore is a strong baseline. 
Finally, we compare \model~with VNLI, our supervised baseline, as it is fine-tuned for text--image alignment on a mixed dataset containing COCO (which is used in Gecko(R)), while other metrics are zero-shot.
It is worth noting that the correlation scores of different auto-eval metrics are generally higher on Gecko(S) than on Gecko(R). This validates that our skills-based benchmark has a more objective and balanced measure of alignment.
We observe similar conclusions on the Gecko2K Reliable Prompts (Gecko2K-Rel) subset; results are  in \Appendix{app_sec:corr_pearson}. 

\begin{figure}
\CenterFloatBoxes
\begin{floatrow}
\capbtabbox{%
  \begin{adjustbox}{max width=6cm} 
\begin{tabular}{c|cc|cc}
\hline
\textbf{Metrics}      & \textbf{QA}    & \textbf{VQA}   & \textbf{Spearman's $\rho$  } & \textbf{Kendall’s $\tau$} \\ \hline
ROUGE-L               & \multirow{4}{*}{N/A} & \multirow{4}{*}{N/A} & 0.33              & 0.25               \\
METEOR                &                      &                      & 0.34              & 0.27               \\
SPICE                 &                      &                      & 0.33              & 0.23               \\
CLIP                  &                      &                      & 0.33              & 0.23               \\ \hline
\multirow{3}{*}{TIFA} & GPT-3                & BLIP-2               & 0.56              & 0.44               \\
                      & GPT-3                & MPLUG                & 0.60              & 0.47               \\
                      & PALM                 & PaLI                 & 0.43              & 0.32               \\ \hline
DSG                   & PALM                 & PaLI                 & 0.57              & 0.46               \\ \hline
Gecko                 & PALM                 & PaLI                 & \textbf{0.64}     & \textbf{0.50}      \\ \hline
\end{tabular}%
\end{adjustbox}
}{%
  \caption{\textbf{Comparing different metrics by their correlation with human Likert ratings on TIFA160.} The Gecko metric outperforms the others by a significant margin.}%
\label{tab:tifa160}
}
\capbtabbox{%
  \begin{adjustbox}{max width=6cm} 
\begin{tabular}{c|cc|c}
\hline                          
\multicolumn{1}{c|}{\multirow{2}{*}{\textbf{Metrics}}}
 & \multicolumn{1}{c}{WL} & \multicolumn{1}{c|}{Likert} & SxS \\
\multicolumn{1}{c|}{}  & \multicolumn{2}{c|}{Pearson} &  Acc \\ 
\hline
VideoCLIP & 0.18 & 0.21 & 28.0 \\
VQAScore$_{\texttt{Gemini Flash}}$ & 0.30 & 0.33 & 52.0 \\
Gecko$_{\texttt{Gemini Flash}}$ & \textbf{0.43} & \textbf{0.45} & \textbf{55.8} \\ 
\hline
\end{tabular}
\end{adjustbox}
}{%
\caption{\textbf{Correlation between auto-eval metrics and human ratings for text-to-video evaluations.} Gecko outperforms other auto-eval metrics on VBench overall consistency prompts, demonstrating the generality of the approach to other modalities.}
\label{tab:corr_pearson_video}
}
\end{floatrow}
\vspace{-5mm}
\end{figure}

\noindent {\bf TIFA160 results.}
We compare the \model~metric with other metrics on TIFA160~\citep{hu2023tifa}, a set of 160 text--image pairs, each annotated with two Likert ratings.  
In \Table{tab:tifa160}, we list the results reported in \citet{hu2023tifa} and \citet{cho2023davidsonian}, and compare them with \model~as well as our re-implementation of TIFA / DSG.  \model~has the highest correlation, with an average correlation 0.07 higher than that of DSG, when using the same QA and VQA models. This shows that the power of our proposed metric is from the method itself, not from the advance of models used.

{\bf Skill-based evaluation with \model.} To better understand the differences between auto-eval metrics/annotation templates with respect to various skills, we visualise a breakdown of skills in Gecko(S) in \Figure{fig:results_by_category_high_level} and \Appendix{app:t2imodelpercategory}, \ref{app:metricpercategory}. The metrics have different strengths: \eg, we see that while \model, VQAScore, VNLI metrics are consistently good across skills, the \model~metric is better on more complex and compositional language, DSG is best on compositional prompts, and VNLI is better on named entities. As with the overall results, these per skill conclusions seem to hold across templates.

%

\begin{figure}[t]
\tiny
\renewcommand{\arraystretch}{0.8}
\centering
\begin{tabular}{c|cccc|cccc}
    {\bf Skill (subskill):} & \multicolumn{4}{c}{lang/complexity ({\bf negation})} & \multicolumn{4}{c}{Shape: ({\bf hierarchical})} \\
    {\bf Prompt:} & \multicolumn{4}{c}{A bridge with no cars on it.} & \multicolumn{4}{c}{The number 0 made of smaller circles} \\ \midrule
    & \includegraphics[width=0.075\textwidth]{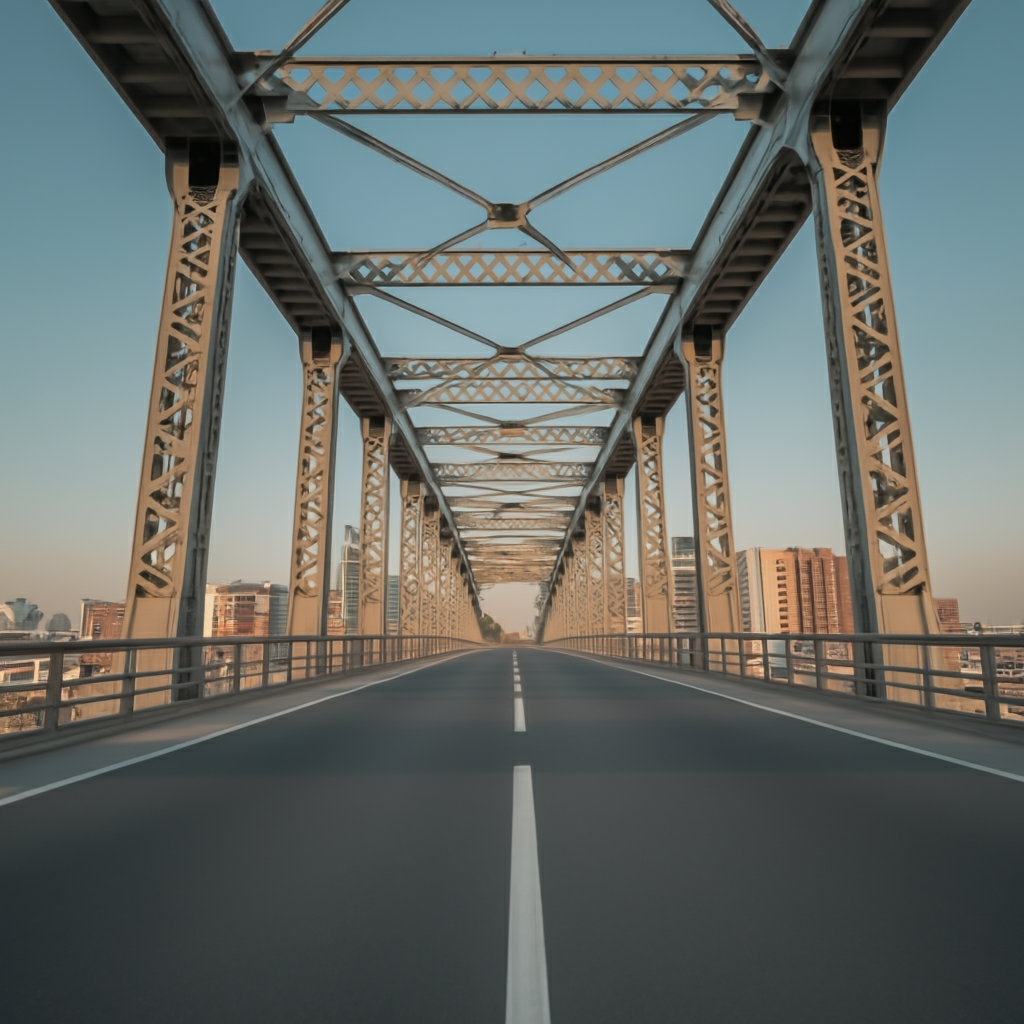} & \includegraphics[width=0.075\textwidth]{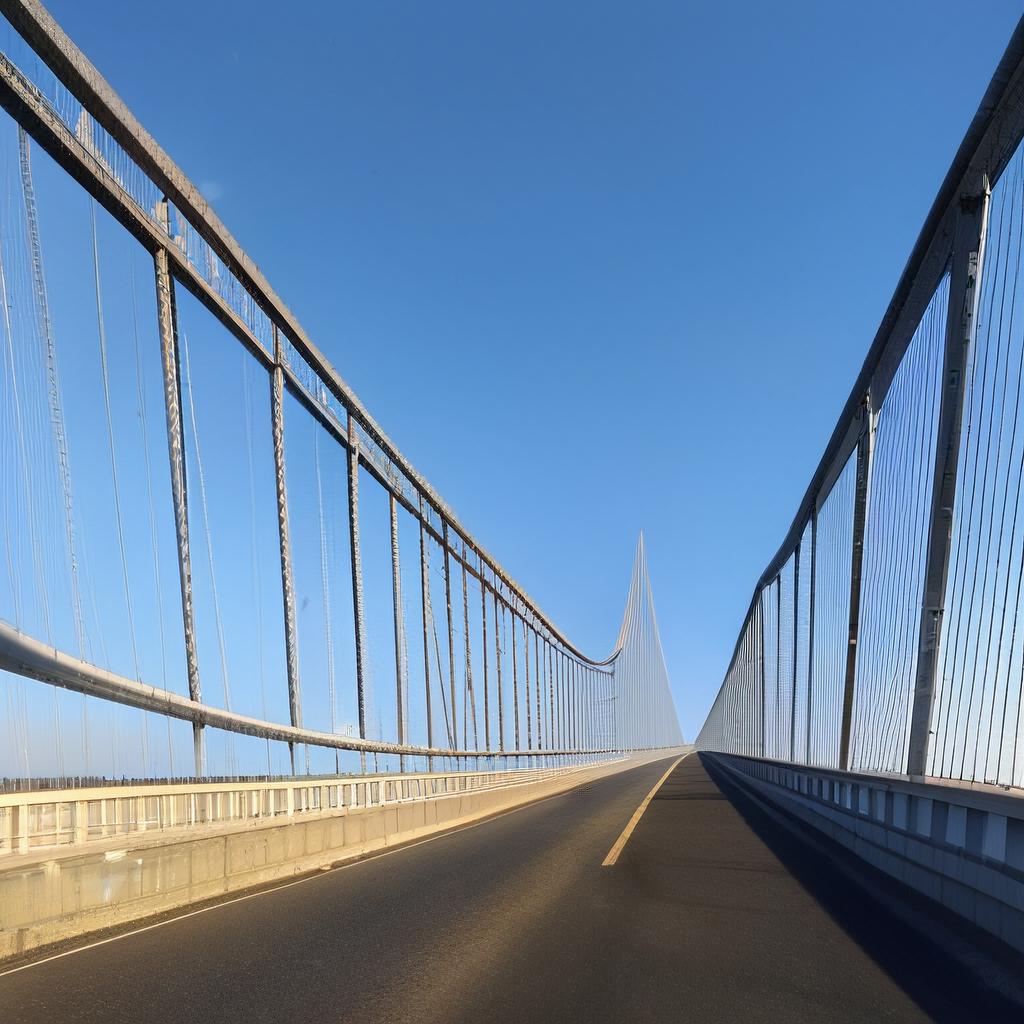}  & \includegraphics[width=0.075\textwidth]{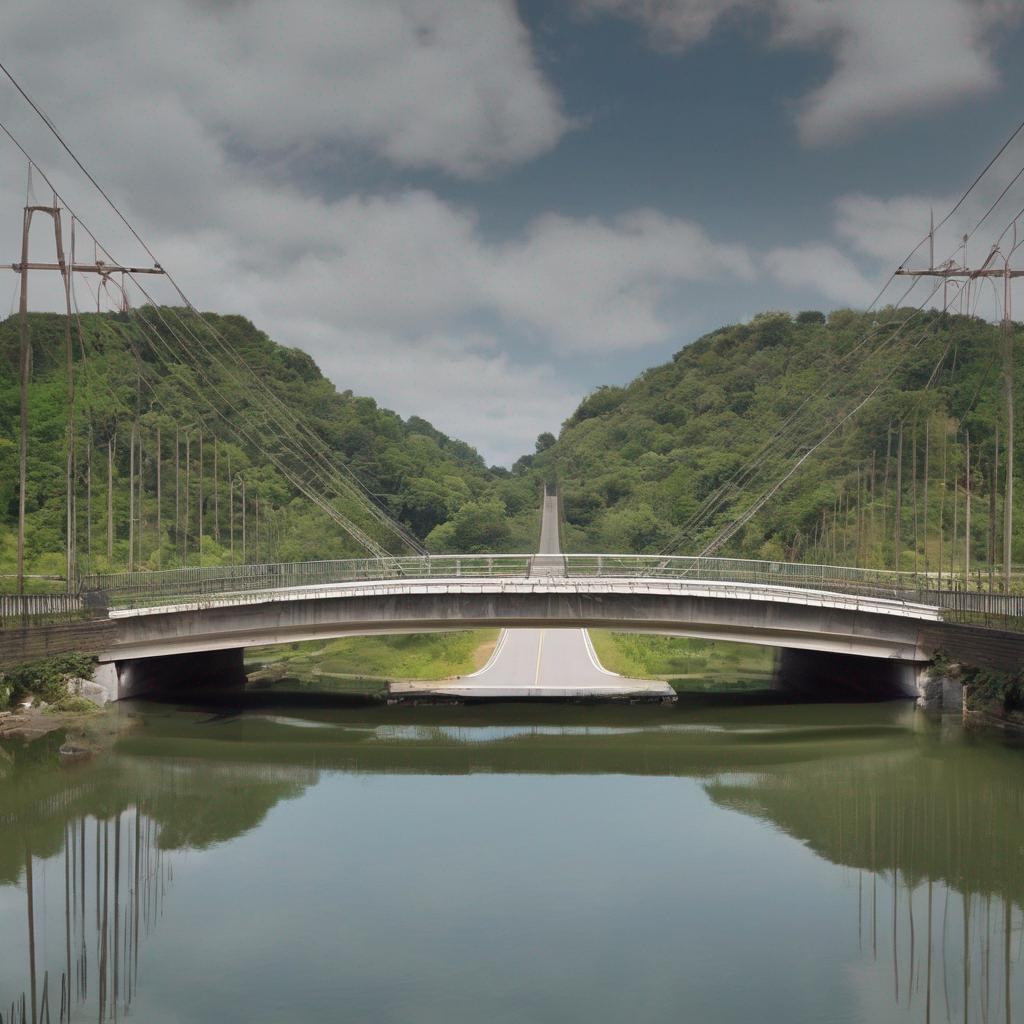} & \includegraphics[width=0.075\textwidth]{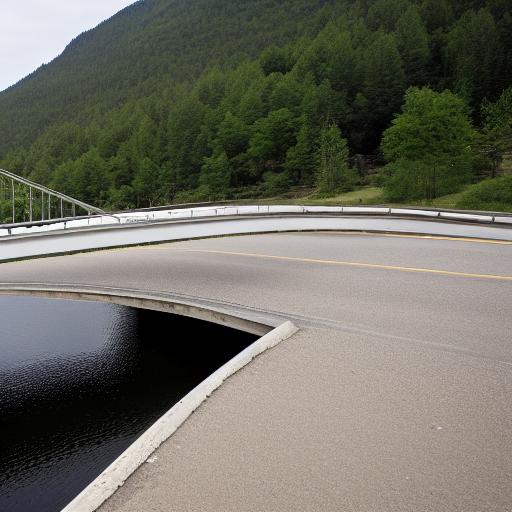} & \includegraphics[width=0.075\textwidth]{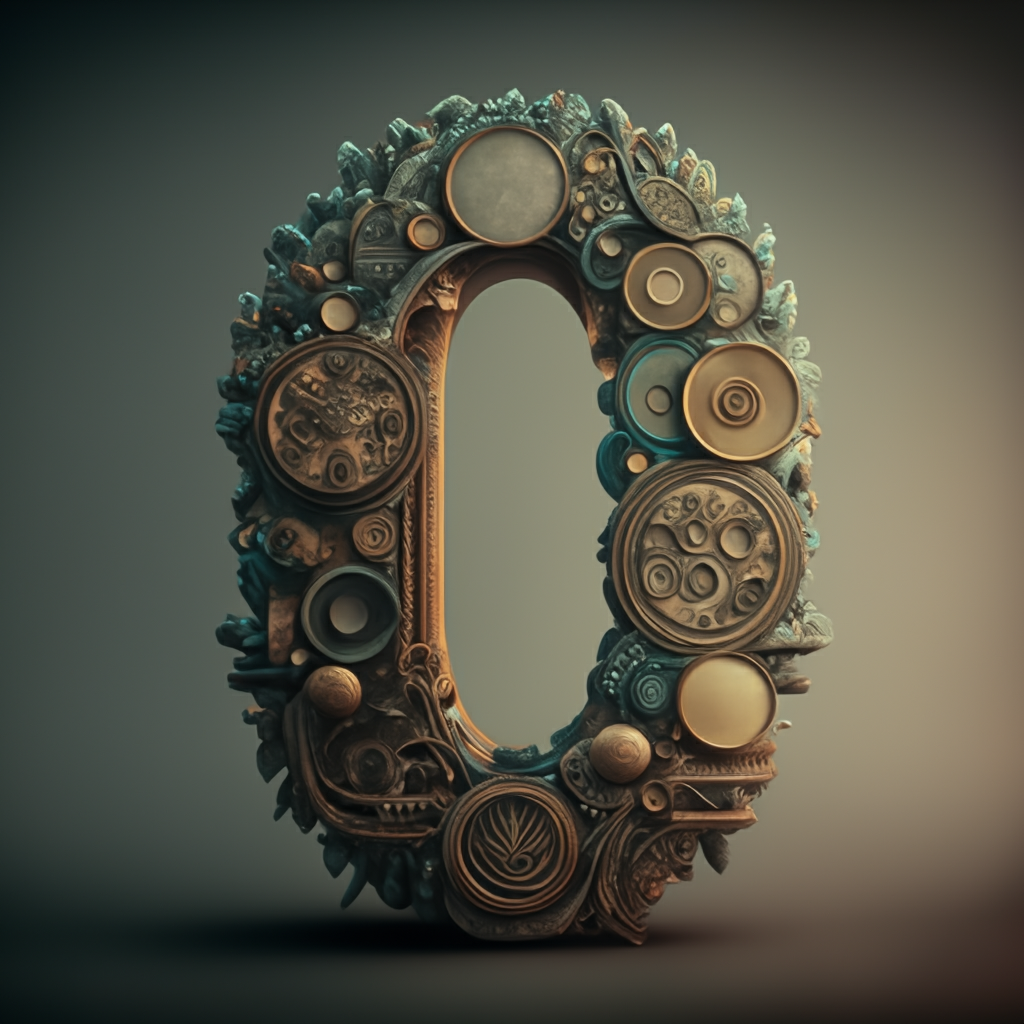} & \includegraphics[width=0.075\textwidth]{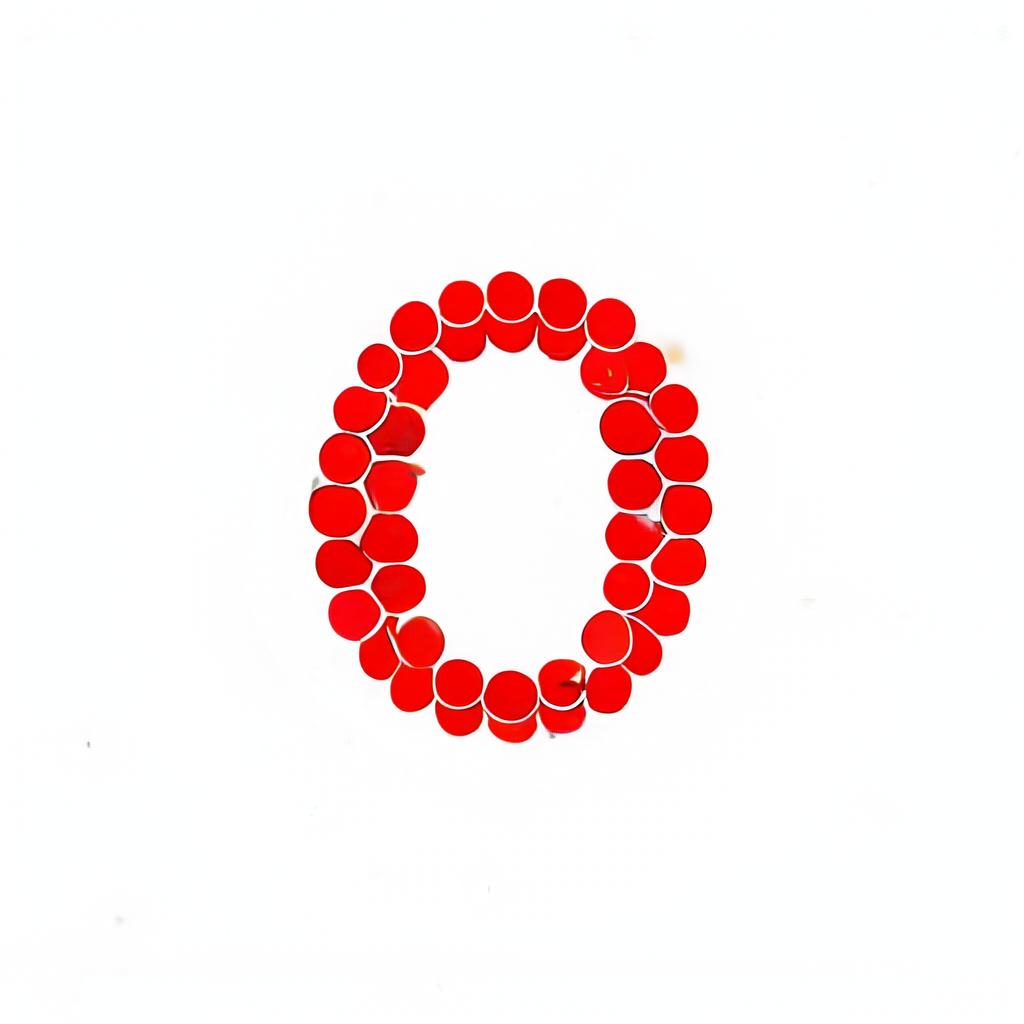}  & \includegraphics[width=0.075\textwidth]{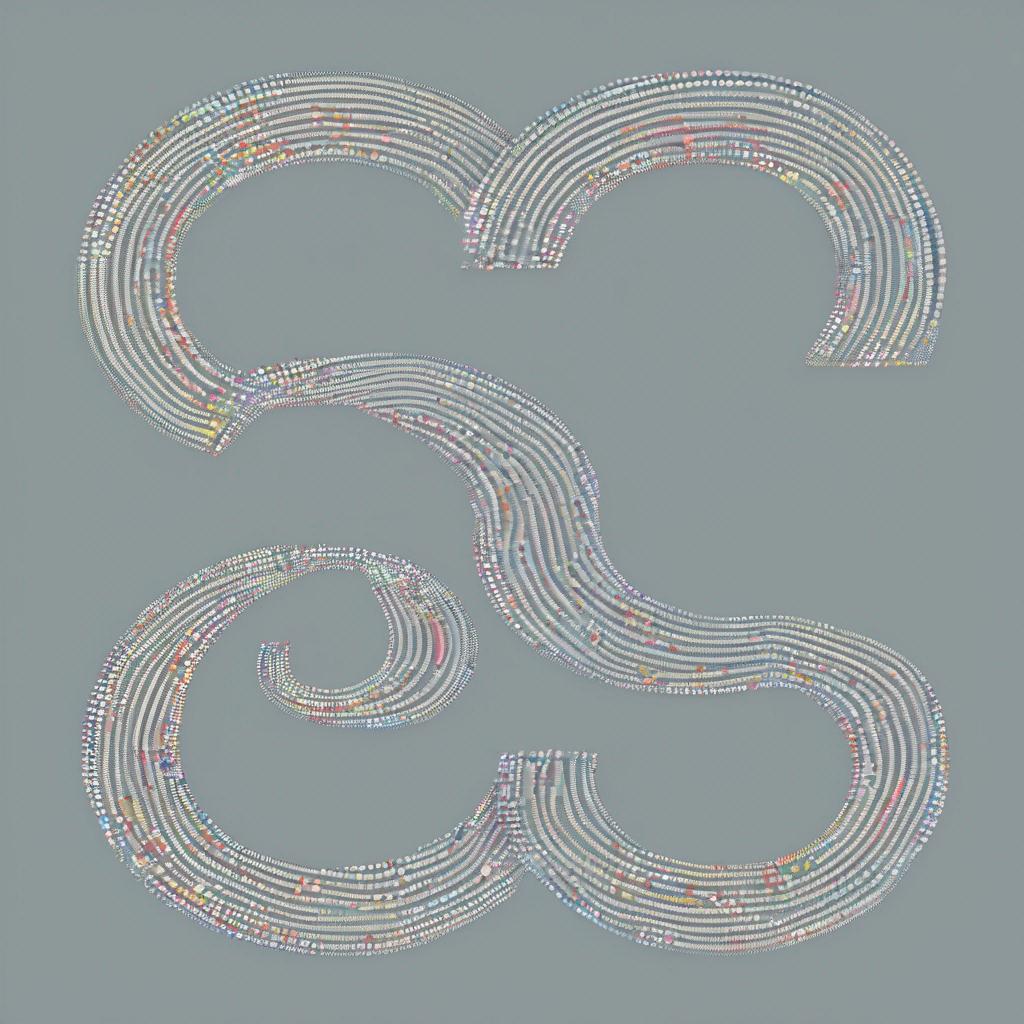} & \includegraphics[width=0.075\textwidth]{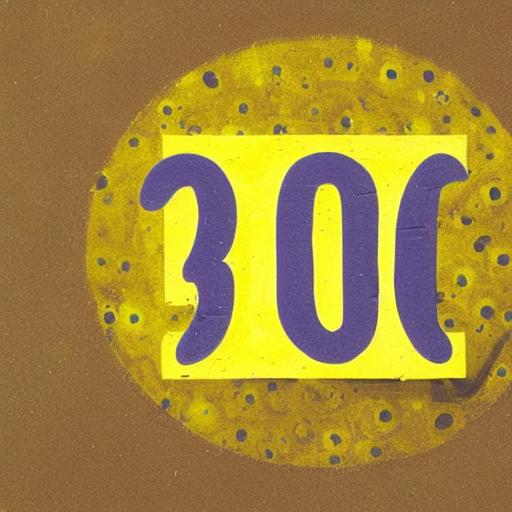} \\
    & {Imagen} & Muse & SDXL & SD1.5 & {Imagen} & Muse & SDXL & SD1.5 \\ \midrule
    WL: & 1. & 1. & 1. & 1. & 1. & 1. & 0. & 0.67 \\
    Likert: & 1. & 1. & 1. & 0.87 & 1. & 0.87 & 0.2 & 0.67 \\
    DSG(H): & 0.5 & 0.5 & 0.5 & 0.67 & 0.92 & 1. & 0. & 0.89 \\ \midrule
    Gecko: & 0.96 & 0.94 & 0.93 & 0.91 & 0.9 & 0.95 & 0.55 & 0.75 \\
    DSG: & 0.25 & 0.25 & 0.25 & 0.25 & 1. & 1. & 0. & 0.  \\
    VNLI: & 0.4 & 0.42 & 0.32 & 0.30 & 0.36 & 0.79 & 0.24 & 0.32 \\
    \bottomrule
\end{tabular}
\vspace{-1mm}
\caption{\textbf{Qualitative results.} Image generations of the four T2I models on prompts in Gecko(S), with the human annotation ratings and auto-eval scores.}
\label{fig:resultssynthetic}
\end{figure}

{\bf Qualitative examples.} We visualise examples in \Figure{fig:resultssynthetic}. For the negation example, the reason DSG(H) gives inconsistent results with WL/Likert here is that the question generation is confused by the negation (asking if there {\em are} cars as opposed to {\em no} cars). We can also see that VNLI and DSG mistakenly think none of the images are aligned. VNLI and DSG perform better on the shape prompt but VNLI scores Imagen incorrectly and DSG gives hard scores per question (0 or 1) and so it is sometimes not able to capture subtler differences in the human ratings.

\subsection{Comparing auto-eval metrics on \pairscore}
We measure how well an auto-eval metric is able to select between two generations given a prompt.
We compare metrics' predictions with the human choices we collected by computing  accuracy--the percentage of times the metric gets the comparison right. Results are in the {\em SxS} column in \Table{tab:corr}.
Although \model~was the clear winner on \pointscore, single-score metrics are generally very good at SxS comparison.  PyramidCLIP was worse than TIFA and DSG on \pointscore, but it has a much higher SxS accuracy, showing that different human annotation templates {\em do not} always give the same result, and single-score metrics can be a good estimator on the \pairscore~task. While VQAScore is better than the \model~metric on SxS comparison on Gecko(S), \model~is better on Gecko(R) and the \model~metric is the only interpretable metric that has better or comparable performance with single-score metrics on SxS comparisons.


\subsection{Comparing auto-eval metrics on \paircompare}
\label{sec:expmodelordering}

A good auto-eval metric should be able to give an overall \paircompare~for a set of prompts.
To decide on a ground-truth ordering, we use Gecko2K-rel as it is the largest subset that has highest agreement across templates. We take the majority vote relationship in \Figure{fig:h_eval_wilcoxon} as the ground truth.
We compare these results to the significant relationships found using the auto-eval metrics in \Appendix{app:model_ordering_eval} (we only use PaLM/PaLI-2 backends if there is a choice).
We find that CLIP performs poorly, confusing wins with losses.
All other auto-eval metrics perform well, never confusing a win with a loss but sometimes not finding significant relations when there is one or vice versa.
\model~correctly finds and predicts {\em all} significant relations, unlike the other metrics.

\subsection{Extending Gecko to Other Modalities}
\label{main:video_results}

To explore the generality of our approach on different modalities, we validate it on text-to-video generation. We choose a prompt set from VBench~\citep{huang2024vbench} and compare the following models: Lumiere~\citep{bar2024lumiere}, Phenaki~\citep{villegas2022phenaki} and WALT~\citep{gupta2023photorealistic}. For human evaluation, we  consider absolute (i.e., Likert, Word Level) and side-by-side templates. For automatic evaluation, we benchmark contrastive models (i.e., VideoCLIP;~\citealt{xu2021videoclip}) and VQA-based metrics. For VQA-based metrics, we extend the VQAScore and our fine-grained Gecko metric on videos using Gemini Flash, which can process long context multimodal inputs. We present the results in Table~\ref{tab:corr_pearson_video} and find that the Gecko metric agrees more closely with human judgement across all human templates than other metrics. See Appendix~\ref{app:video_results} for more details.

\begin{mdframed}[style=mystyle]
{\bf Takeaway:}  
Although \model~is the best metric on different human templates and modalities, we find that the ranking of different auto-eval metrics can change depending on whether they are evaluated on an instance-level template (\eg, Likert or DSG(H)), a comparative template (\eg~SxS) or for model ordering. It is important to evaluate metrics across a range of settings and in particular on one relative and one absolute template if under budget constraints.
\end{mdframed}

\section{Conclusions}
\vspace{-2mm}

We introduce the Gecko evaluation suite, a comprehensive set of prompts, human ratings across templates, and tasks to evaluate T2I models and alignment metrics.
We find that looking at a single slice of the data (\eg, one annotation template, or one evaluation task) can give misleading observations of the relative benefits of one model or metric. Instead, we show that we need to use a comprehensive prompt set (or manually evaluated ``reliable prompts'') to achieve consistent model orderings and thereby confidence in model rankings.
Given this evaluation suite, we demonstrate that our \model~metric performs consistently best across three tasks, measuring how metrics perform in scoring each image--text instance with respect to their alignment as well as ranking models. 
Our work highlights the importance of standardising the evaluation framework with respect to the prompt sets, the annotation templates, and metrics used. This is crucial when conducting research on models and metrics, and also to make informed decisions.


%
%

\clearpage
\section{Ethics Statement}
When gathering our dataset, we ensure that raters are compensated and provide consent as described in \Appendix{app:sec:humanevaldatacollectiondetails}. 
We also run safety filters over the generated images before giving them to the raters.
This work is a step towards better evaluation of text-to-image models which are known to hallucinate. 
It gives tools to others developers and practitioners to properly understand and evaluate T2I models in the future.

\section{Reproducibility Statement}
We give extensive details of our setup in the Appendix.
For human annotation, we visualise the templates used and give extensive detail on how these raw ratings are aggregated in \Appendix{app:human_eval_templates}.
For the dataset collation, we give the few shot prompts used to generate tags and templates: the few shot prompt for Gecko(R) is given in  \Listing{alg:auto_tagging_prompt}.
For Gecko(S), we give our full decomposition of skills in \Table{tab:breakdown_prompts} with examples and an explanation of how we generated prompts for each specific skill in \Appendix{app:geckosbreakdown} with sample few shot prompts.
For metrics, we give full details of the baselines in \Section{sec:experimentalsetup} and the additional CLIP baselines in \Section{app:all_correlations}.
For \model~metrics, we give the few shot prompt for generating coverage in \Listing{alg:few_shot_prompt_llm_coverage} and for generating the QAs in \Listing{alg:few_shot_prompt_llm_qa}.

\paragraph{Acknowledgements}
We thank Zi Wang, Miloš Stanojević, and Jason Baldridge for their feedback throughout the project. We are grateful to Andrew Zisserman for his feedback on the manuscript. We thank Aayush Upadhyay and the rest of the Podium team for their help in running models.
\clearpage
\bibliographystyle{iclr2025_conference}
\bibliography{egbib}

\begin{thebibliography}{76}
\providecommand{\natexlab}[1]{#1}
\providecommand{\url}[1]{\texttt{#1}}
\expandafter\ifx\csname urlstyle\endcsname\relax
  \providecommand{\doi}[1]{doi: #1}\else
  \providecommand{\doi}{doi: \begingroup \urlstyle{rm}\Url}\fi

\bibitem[Agrawal et~al.(2018)Agrawal, Batra, Parikh, and
  Kembhavi]{agrawal2018don}
Aishwarya Agrawal, Dhruv Batra, Devi Parikh, and Aniruddha Kembhavi.
\newblock Don't just assume; look and answer: Overcoming priors for visual
  question answering.
\newblock In \emph{Proceedings of the IEEE conference on computer vision and
  pattern recognition}, pp.\  4971--4980, 2018.

\bibitem[Anil et~al.(2023)Anil, Dai, Firat, Johnson, Lepikhin, Passos, Shakeri,
  Taropa, Bailey, Chen, et~al.]{anil2023palm}
Rohan Anil, Andrew~M Dai, Orhan Firat, Melvin Johnson, Dmitry Lepikhin,
  Alexandre Passos, Siamak Shakeri, Emanuel Taropa, Paige Bailey, Zhifeng Chen,
  et~al.
\newblock Palm 2 technical report.
\newblock \emph{arXiv preprint arXiv:2305.10403}, 2023.

\bibitem[Bang et~al.(2023)Bang, Cahyawijaya, Lee, Dai, Su, Wilie, Lovenia, Ji,
  Yu, Chung, et~al.]{bang2023multitask}
Yejin Bang, Samuel Cahyawijaya, Nayeon Lee, Wenliang Dai, Dan Su, Bryan Wilie,
  Holy Lovenia, Ziwei Ji, Tiezheng Yu, Willy Chung, et~al.
\newblock A multitask, multilingual, multimodal evaluation of chatgpt on
  reasoning, hallucination, and interactivity.
\newblock \emph{arXiv preprint arXiv:2302.04023}, 2023.

\bibitem[Bar-Tal et~al.(2024)Bar-Tal, Chefer, Tov, Herrmann, Paiss, Zada,
  Ephrat, Hur, Li, Michaeli, et~al.]{bar2024lumiere}
Omer Bar-Tal, Hila Chefer, Omer Tov, Charles Herrmann, Roni Paiss, Shiran Zada,
  Ariel Ephrat, Junhwa Hur, Yuanzhen Li, Tomer Michaeli, et~al.
\newblock Lumiere: A space-time diffusion model for video generation.
\newblock \emph{arXiv preprint arXiv:2401.12945}, 2024.

\bibitem[Betker et~al.(2023)Betker, Goh, Jing, Brooks, Wang, Li, Ouyang,
  Zhuang, Lee, Guo, et~al.]{betker2023improving}
James Betker, Gabriel Goh, Li~Jing, Tim Brooks, Jianfeng Wang, Linjie Li, Long
  Ouyang, Juntang Zhuang, Joyce Lee, Yufei Guo, et~al.
\newblock Improving image generation with better captions.
\newblock \emph{Computer Science. https://cdn. openai. com/papers/dall-e-3.
  pdf}, 2\penalty0 (3):\penalty0 8, 2023.

\bibitem[Bitton-Guetta et~al.(2023)Bitton-Guetta, Bitton, Hessel, Schmidt,
  Elovici, Stanovsky, and Schwartz]{bitton2023breaking}
Nitzan Bitton-Guetta, Yonatan Bitton, Jack Hessel, Ludwig Schmidt, Yuval
  Elovici, Gabriel Stanovsky, and Roy Schwartz.
\newblock Breaking common sense: Whoops! a vision-and-language benchmark of
  synthetic and compositional images.
\newblock In \emph{ICCV}, 2023.

\bibitem[Bugliarello et~al.(2023)Bugliarello, Sartran, Agrawal, Hendricks, and
  Nematzadeh]{bugliarello2023measuring}
Emanuele Bugliarello, Laurent Sartran, Aishwarya Agrawal, Lisa~Anne Hendricks,
  and Aida Nematzadeh.
\newblock Measuring progress in fine-grained vision-and-language understanding.
\newblock \emph{arXiv preprint arXiv:2305.07558}, 2023.

\bibitem[Caesar et~al.(2018)Caesar, Uijlings, and Ferrari]{caesar2018cvpr}
Holger Caesar, Jasper Uijlings, and Vittorio Ferrari.
\newblock Coco-stuff: Thing and stuff classes in context.
\newblock In \emph{Computer vision and pattern recognition (CVPR), 2018 IEEE
  conference on}. IEEE, 2018.

\bibitem[Chang et~al.(2023)Chang, Zhang, Barber, Maschinot, Lezama, Jiang,
  Yang, Murphy, Freeman, Rubinstein, Li, and Krishnan]{chang2023muse}
Huiwen Chang, Han Zhang, Jarred Barber, Aaron Maschinot, Jose Lezama, Lu~Jiang,
  Ming-Hsuan Yang, Kevin~Patrick Murphy, William~T. Freeman, Michael
  Rubinstein, Yuanzhen Li, and Dilip Krishnan.
\newblock Muse: Text-to-image generation via masked generative transformers.
\newblock In Andreas Krause, Emma Brunskill, Kyunghyun Cho, Barbara Engelhardt,
  Sivan Sabato, and Jonathan Scarlett (eds.), \emph{Proceedings of the 40th
  International Conference on Machine Learning}, volume 202 of
  \emph{Proceedings of Machine Learning Research}, pp.\  4055--4075. PMLR,
  23--29 Jul 2023.
\newblock URL \url{https://proceedings.mlr.press/v202/chang23b.html}.

\bibitem[Chen et~al.(2022)Chen, Wang, Changpinyo, Piergiovanni, Padlewski,
  Salz, Goodman, Grycner, Mustafa, Beyer, Kolesnikov, Puigcerver, Ding, Rong,
  Akbari, Mishra, Xue, Thapliyal, Bradbury, Kuo, Seyedhosseini, Jia, Ayan,
  Riquelme, Steiner, Angelova, Zhai, Houlsby, and Soricut]{chen2022pali}
Xi~Chen, Xiao Wang, Soravit Changpinyo, AJ~Piergiovanni, Piotr Padlewski,
  Daniel Salz, Sebastian Goodman, Adam Grycner, Basil Mustafa, Lucas Beyer,
  Alexander Kolesnikov, Joan Puigcerver, Nan Ding, Keran Rong, Hassan Akbari,
  Gaurav Mishra, Linting Xue, Ashish Thapliyal, James Bradbury, Weicheng Kuo,
  Mojtaba Seyedhosseini, Chao Jia, Burcu~Karagol Ayan, Carlos Riquelme, Andreas
  Steiner, Anelia Angelova, Xiaohua Zhai, Neil Houlsby, and Radu Soricut.
\newblock Pali: A jointly-scaled multilingual language-image model.
\newblock \emph{arXiv preprint arXiv:2209.06794}, 2022.

\bibitem[Chen et~al.(2023)Chen, Fern{\'a}ndez, and Pezzelle]{chen2023bla}
Xinyi Chen, Raquel Fern{\'a}ndez, and Sandro Pezzelle.
\newblock The bla benchmark: Investigating basic language abilities of
  pre-trained multimodal models.
\newblock In \emph{Proceedings of the 2023 Conference on Empirical Methods in
  Natural Language Processing}, pp.\  5817--5830, 2023.

\bibitem[Cho et~al.(2023{\natexlab{a}})Cho, Hu, Garg, Anderson, Krishna,
  Baldridge, Bansal, Pont-Tuset, and Wang]{cho2023davidsonian}
Jaemin Cho, Yushi Hu, Roopal Garg, Peter Anderson, Ranjay Krishna, Jason
  Baldridge, Mohit Bansal, Jordi Pont-Tuset, and Su~Wang.
\newblock Davidsonian scene graph: Improving reliability in fine-grained
  evaluation for text-image generation.
\newblock \emph{arXiv preprint arXiv:2310.18235}, 2023{\natexlab{a}}.

\bibitem[Cho et~al.(2023{\natexlab{b}})Cho, Zala, and Bansal]{cho2023dall}
Jaemin Cho, Abhay Zala, and Mohit Bansal.
\newblock Dall-eval: Probing the reasoning skills and social biases of
  text-to-image generation models.
\newblock In \emph{ICCV}, 2023{\natexlab{b}}.

\bibitem[Clark et~al.(2021)Clark, August, Serrano, Haduong, Gururangan, and
  Smith]{clark-etal-2021-thats}
Elizabeth Clark, Tal August, Sofia Serrano, Nikita Haduong, Suchin Gururangan,
  and Noah~A. Smith.
\newblock All that{'}s {`}human{'} is not gold: Evaluating human evaluation of
  generated text.
\newblock In Chengqing Zong, Fei Xia, Wenjie Li, and Roberto Navigli (eds.),
  \emph{Proceedings of the 59th Annual Meeting of the Association for
  Computational Linguistics and the 11th International Joint Conference on
  Natural Language Processing (Volume 1: Long Papers)}, 2021.

\bibitem[Delis et~al.(1986)Delis, Robertson, and Efron]{DELIS1986205}
Dean~C. Delis, Lynn~C. Robertson, and Robert Efron.
\newblock Hemispheric specialization of memory for visual hierarchical stimuli.
\newblock \emph{Neuropsychologia}, 24\penalty0 (2):\penalty0 205--214, 1986.
\newblock ISSN 0028-3932.

\bibitem[Delmas et~al.(2022)Delmas, Weinzaepfel, Lucas, Moreno-Noguer, and
  Rogez]{delmas2022posescript}
Ginger Delmas, Philippe Weinzaepfel, Thomas Lucas, Francesc Moreno-Noguer, and
  Gr{\'e}gory Rogez.
\newblock Posescript: 3d human poses from natural language.
\newblock In \emph{ECCV}, 2022.

\bibitem[Dosovitskiy et~al.(2020)Dosovitskiy, Beyer, Kolesnikov, Weissenborn,
  Zhai, Unterthiner, Dehghani, Minderer, Heigold, Gelly,
  et~al.]{dosovitskiy2020image}
Alexey Dosovitskiy, Lucas Beyer, Alexander Kolesnikov, Dirk Weissenborn,
  Xiaohua Zhai, Thomas Unterthiner, Mostafa Dehghani, Matthias Minderer, Georg
  Heigold, Sylvain Gelly, et~al.
\newblock An image is worth 16x16 words: Transformers for image recognition at
  scale.
\newblock \emph{arXiv preprint arXiv:2010.11929}, 2020.

\bibitem[Gao et~al.(2022)Gao, Liu, Xu, Zhang, Li, Ji, and
  Shen]{gao2022pyramidclip}
Yuting Gao, Jinfeng Liu, Zihan Xu, Jun Zhang, Ke~Li, Rongrong Ji, and Chunhua
  Shen.
\newblock Pyramidclip: Hierarchical feature alignment for vision-language model
  pretraining.
\newblock In \emph{NeurIPS}, 2022.

\bibitem[Geirhos et~al.(2020)Geirhos, Meding, and Wichmann]{geirhos2020beyond}
Robert Geirhos, Kristof Meding, and Felix~A Wichmann.
\newblock Beyond accuracy: quantifying trial-by-trial behaviour of cnns and
  humans by measuring error consistency.
\newblock \emph{Advances in Neural Information Processing Systems},
  33:\penalty0 13890--13902, 2020.

\bibitem[Geva et~al.(2021)Geva, Khashabi, Segal, Khot, Roth, and
  Berant]{Mor_Geva_2021}
Mor Geva, Daniel Khashabi, Elad Segal, Tushar Khot, Dan Roth, and Jonathan
  Berant.
\newblock {Did Aristotle Use a Laptop? A Question Answering Benchmark with
  Implicit Reasoning Strategies}.
\newblock \emph{TACL}, 9:\penalty0 346--361, 2021.

\bibitem[Gokhale et~al.(2022)Gokhale, Palangi, Nushi, Vineet, Horvitz, Kamar,
  Baral, and Yang]{gokhale2022benchmarking}
Tejas Gokhale, Hamid Palangi, Besmira Nushi, Vibhav Vineet, Eric Horvitz, Ece
  Kamar, Chitta Baral, and Yezhou Yang.
\newblock Benchmarking spatial relationships in text-to-image generation.
\newblock \emph{arXiv preprint arXiv:2212.10015}, 2022.

\bibitem[Guerreiro et~al.(2023)Guerreiro, Alves, Waldendorf, Haddow, Birch,
  Colombo, and Martins]{guerreiro2023hallucinations}
Nuno~M Guerreiro, Duarte~M Alves, Jonas Waldendorf, Barry Haddow, Alexandra
  Birch, Pierre Colombo, and Andr{\'e}~FT Martins.
\newblock Hallucinations in large multilingual translation models.
\newblock \emph{TACL}, 11:\penalty0 1500--1517, 2023.

\bibitem[Gupta et~al.(2019)Gupta, Dollar, and Girshick]{gupta2019lvis}
Agrim Gupta, Piotr Dollar, and Ross Girshick.
\newblock {LVIS}: A dataset for large vocabulary instance segmentation.
\newblock In \emph{Proceedings of the {IEEE} Conference on Computer Vision and
  Pattern Recognition}, 2019.

\bibitem[Gupta et~al.(2023)Gupta, Yu, Sohn, Gu, Hahn, Fei-Fei, Essa, Jiang, and
  Lezama]{gupta2023photorealistic}
Agrim Gupta, Lijun Yu, Kihyuk Sohn, Xiuye Gu, Meera Hahn, Li~Fei-Fei, Irfan
  Essa, Lu~Jiang, and Jos{\'e} Lezama.
\newblock Photorealistic video generation with diffusion models.
\newblock \emph{arXiv preprint arXiv:2312.06662}, 2023.

\bibitem[Hayes \& Krippendorff(2007)Hayes and Krippendorff]{hayes2007answering}
Andrew~F Hayes and Klaus Krippendorff.
\newblock Answering the call for a standard reliability measure for coding
  data.
\newblock \emph{Communication methods and measures}, 1\penalty0 (1):\penalty0
  77--89, 2007.

\bibitem[Heo et~al.(2022)Heo, Kim, Park, and Back]{heo2022comparison}
Cindy~Yoonjoung Heo, Bona Kim, Kwangsoo Park, and Robin~M Back.
\newblock A comparison of best-worst scaling and likert scale methods on
  peer-to-peer accommodation attributes.
\newblock \emph{Journal of business research}, 148:\penalty0 368--377, 2022.

\bibitem[Hessel et~al.(2021)Hessel, Holtzman, Forbes, Le~Bras, and
  Choi]{hessel2021clipscore}
Jack Hessel, Ari Holtzman, Maxwell Forbes, Ronan Le~Bras, and Yejin Choi.
\newblock Clipscore: A reference-free evaluation metric for image captioning.
\newblock In \emph{EMNLP}, pp.\  7514--7528, 2021.

\bibitem[Honnibal \& Montani(2017)Honnibal and Montani]{spacy2}
Matthew Honnibal and Ines Montani.
\newblock Spacy 2: Natural language understanding with bloom embeddings,
  convolutional neural networks and incremental parsing.
\newblock \emph{spacy.io}, 2017.

\bibitem[Honovich et~al.(2021)Honovich, Choshen, Aharoni, Neeman, Szpektor, and
  Abend]{honovich2021q}
Or~Honovich, Leshem Choshen, Roee Aharoni, Ella Neeman, Idan Szpektor, and Omri
  Abend.
\newblock Q2: Evaluating factual consistency in knowledge-grounded dialogues
  via question generation and question answering.
\newblock \emph{arXiv preprint arXiv:2104.08202}, 2021.

\bibitem[Honovich et~al.(2022)Honovich, Aharoni, Herzig, Taitelbaum, Kukliansy,
  Cohen, Scialom, Szpektor, Hassidim, and Matias]{honovich2022true}
Or~Honovich, Roee Aharoni, Jonathan Herzig, Hagai Taitelbaum, Doron Kukliansy,
  Vered Cohen, Thomas Scialom, Idan Szpektor, Avinatan Hassidim, and Yossi
  Matias.
\newblock True: Re-evaluating factual consistency evaluation.
\newblock \emph{arXiv preprint arXiv:2204.04991}, 2022.

\bibitem[Hu et~al.(2023)Hu, Liu, Kasai, Wang, Ostendorf, Krishna, and
  Smith]{hu2023tifa}
Yushi Hu, Benlin Liu, Jungo Kasai, Yizhong Wang, Mari Ostendorf, Ranjay
  Krishna, and Noah~A Smith.
\newblock Tifa: Accurate and interpretable text-to-image faithfulness
  evaluation with question answering.
\newblock \emph{arXiv preprint arXiv:2303.11897}, 2023.

\bibitem[Huang et~al.(2024{\natexlab{a}})Huang, Sun, Xie, Li, and
  Liu]{huang2024t2i}
Kaiyi Huang, Kaiyue Sun, Enze Xie, Zhenguo Li, and Xihui Liu.
\newblock T2i-compbench: A comprehensive benchmark for open-world compositional
  text-to-image generation.
\newblock \emph{NeurIPS}, 36, 2024{\natexlab{a}}.

\bibitem[Huang et~al.(2024{\natexlab{b}})Huang, He, Yu, Zhang, Si, Jiang,
  Zhang, Wu, Jin, Chanpaisit, et~al.]{huang2024vbench}
Ziqi Huang, Yinan He, Jiashuo Yu, Fan Zhang, Chenyang Si, Yuming Jiang, Yuanhan
  Zhang, Tianxing Wu, Qingyang Jin, Nattapol Chanpaisit, et~al.
\newblock Vbench: Comprehensive benchmark suite for video generative models.
\newblock In \emph{Proceedings of the IEEE/CVF Conference on Computer Vision
  and Pattern Recognition}, pp.\  21807--21818, 2024{\natexlab{b}}.

\bibitem[Ilharco et~al.(2021)Ilharco, Wortsman, Wightman, Gordon, Carlini,
  Taori, Dave, Shankar, Namkoong, Miller, Hajishirzi, Farhadi, and
  Schmidt]{ilharco_gabriel_2021_5143773}
Gabriel Ilharco, Mitchell Wortsman, Ross Wightman, Cade Gordon, Nicholas
  Carlini, Rohan Taori, Achal Dave, Vaishaal Shankar, Hongseok Namkoong, John
  Miller, Hannaneh Hajishirzi, Ali Farhadi, and Ludwig Schmidt.
\newblock Openclip, July 2021.
\newblock URL \url{https://doi.org/10.5281/zenodo.5143773}.
\newblock If you use this software, please cite it as below.

\bibitem[Kostumov et~al.(2024)Kostumov, Nutfullin, Pilipenko, and
  Ilyushin]{kostumov2024uncertainty}
Vasily Kostumov, Bulat Nutfullin, Oleg Pilipenko, and Eugene Ilyushin.
\newblock Uncertainty-aware evaluation for vision-language models.
\newblock \emph{arXiv preprint arXiv:2402.14418}, 2024.

\bibitem[Krause et~al.(2017)Krause, Johnson, Krishna, and
  Fei-Fei]{krause2017hierarchical}
Jonathan Krause, Justin Johnson, Ranjay Krishna, and Li~Fei-Fei.
\newblock A hierarchical approach for generating descriptive image paragraphs.
\newblock In \emph{CVPR}, 2017.

\bibitem[Kry{\'s}ci{\'n}ski et~al.(2019)Kry{\'s}ci{\'n}ski, McCann, Xiong, and
  Socher]{kryscinski2019evaluating}
Wojciech Kry{\'s}ci{\'n}ski, Bryan McCann, Caiming Xiong, and Richard Socher.
\newblock Evaluating the factual consistency of abstractive text summarization.
\newblock \emph{arXiv preprint arXiv:1910.12840}, 2019.

\bibitem[Lee et~al.(2024)Lee, Yasunaga, Meng, Mai, Park, Gupta, Zhang,
  Narayanan, Teufel, Bellagente, et~al.]{lee2024holistic}
Tony Lee, Michihiro Yasunaga, Chenlin Meng, Yifan Mai, Joon~Sung Park, Agrim
  Gupta, Yunzhi Zhang, Deepak Narayanan, Hannah Teufel, Marco Bellagente,
  et~al.
\newblock Holistic evaluation of text-to-image models.
\newblock \emph{NeurIPS}, 36, 2024.

\bibitem[Li et~al.(2023)Li, Li, Savarese, and Hoi]{li2023blip2}
Junnan Li, Dongxu Li, Silvio Savarese, and Steven Hoi.
\newblock Blip-2: bootstrapping language-image pre-training with frozen image
  encoders and large language models.
\newblock In \emph{Proceedings of the 40th International Conference on Machine
  Learning}, ICML'23. JMLR.org, 2023.

\bibitem[Liang et~al.(2020)Liang, Zou, and Yu]{liang2020beyond}
Weixin Liang, James Zou, and Zhou Yu.
\newblock Beyond user self-reported likert scale ratings: A comparison model
  for automatic dialog evaluation.
\newblock In \emph{Proceedings of the 58th Annual Meeting of the Association
  for Computational Linguistics}, pp.\  1363--1374, 2020.

\bibitem[Liang et~al.(2023)Liang, He, Li, Li, Klimovskiy, Carolan, Sun,
  Pont-Tuset, Young, Yang, et~al.]{liang2023rich}
Youwei Liang, Junfeng He, Gang Li, Peizhao Li, Arseniy Klimovskiy, Nicholas
  Carolan, Jiao Sun, Jordi Pont-Tuset, Sarah Young, Feng Yang, et~al.
\newblock Rich human feedback for text-to-image generation.
\newblock \emph{arXiv preprint arXiv:2312.10240}, 2023.

\bibitem[Lin et~al.(2014)Lin, Maire, Belongie, Hays, Perona, Ramanan,
  Doll{\'a}r, and Zitnick]{lin2014microsoft}
Tsung-Yi Lin, Michael Maire, Serge Belongie, James Hays, Pietro Perona, Deva
  Ramanan, Piotr Doll{\'a}r, and C~Lawrence Zitnick.
\newblock Microsoft coco: Common objects in context.
\newblock In \emph{ECCV}, 2014.

\bibitem[Lin et~al.(2024)Lin, Pathak, Li, Li, Xia, Neubig, Zhang, and
  Ramanan]{lin2024evaluating}
Zhiqiu Lin, Deepak Pathak, Baiqi Li, Jiayao Li, Xide Xia, Graham Neubig,
  Pengchuan Zhang, and Deva Ramanan.
\newblock Evaluating text-to-visual generation with image-to-text generation.
\newblock \emph{arXiv preprint arXiv:2404.01291}, 2024.

\bibitem[Liu et~al.(2022)Liu, Garrette, Saharia, Chan, Roberts, Narang, Blok,
  Mical, Norouzi, and Constant]{liu2022character}
Rosanne Liu, Dan Garrette, Chitwan Saharia, William Chan, Adam Roberts, Sharan
  Narang, Irina Blok, RJ~Mical, Mohammad Norouzi, and Noah Constant.
\newblock Character-aware models improve visual text rendering.
\newblock \emph{arXiv preprint arXiv:2212.10562}, 2022.

\bibitem[Liu et~al.(2024)Liu, Cun, Liu, Wang, Zhang, Chen, Liu, Zeng, Chan, and
  Shan]{liu2024evalcrafter}
Yaofang Liu, Xiaodong Cun, Xuebo Liu, Xintao Wang, Yong Zhang, Haoxin Chen,
  Yang Liu, Tieyong Zeng, Raymond Chan, and Ying Shan.
\newblock Evalcrafter: Benchmarking and evaluating large video generation
  models.
\newblock In \emph{Proceedings of the IEEE/CVF Conference on Computer Vision
  and Pattern Recognition}, pp.\  22139--22149, 2024.

\bibitem[Lu et~al.(2016)Lu, Krishna, Bernstein, and Fei-Fei]{lu2016visual}
Cewu Lu, Ranjay Krishna, Michael Bernstein, and Li~Fei-Fei.
\newblock Visual relationship detection with language priors.
\newblock In \emph{ECCV}, 2016.

\bibitem[Maynez et~al.(2020)Maynez, Narayan, Bohnet, and
  McDonald]{maynez2020faithfulness}
Joshua Maynez, Shashi Narayan, Bernd Bohnet, and Ryan McDonald.
\newblock On faithfulness and factuality in abstractive summarization.
\newblock \emph{arXiv preprint arXiv:2005.00661}, 2020.

\bibitem[Ohta et~al.(2013)Ohta, Fukui, and Sakai]{Ohta2013}
Shinri Ohta, Naoki Fukui, and Kuniyoshi~L. Sakai.
\newblock Computational principles of syntax in the regions specialized for
  language: integrating theoretical linguistics and functional neuroimaging.
\newblock \emph{Frontiers in Behavioral Neuroscience}, 2013.

\bibitem[Paiss et~al.(2023)Paiss, Ephrat, Tov, Zada, Mosseri, Irani, and
  Dekel]{paiss2023teaching}
Roni Paiss, Ariel Ephrat, Omer Tov, Shiran Zada, Inbar Mosseri, Michal Irani,
  and Tali Dekel.
\newblock Teaching clip to count to ten.
\newblock \emph{arXiv preprint arXiv:2302.12066}, 2023.

\bibitem[Podell et~al.(2023)Podell, English, Lacey, Blattmann, Dockhorn,
  M{\"u}ller, Penna, and Rombach]{podell2023sdxl}
Dustin Podell, Zion English, Kyle Lacey, Andreas Blattmann, Tim Dockhorn, Jonas
  M{\"u}ller, Joe Penna, and Robin Rombach.
\newblock Sdxl: Improving latent diffusion models for high-resolution image
  synthesis.
\newblock \emph{arXiv preprint arXiv:2307.01952}, 2023.

\bibitem[Pont-Tuset et~al.(2020)Pont-Tuset, Uijlings, Changpinyo, Soricut, and
  Ferrari]{pont2020connecting}
Jordi Pont-Tuset, Jasper Uijlings, Soravit Changpinyo, Radu Soricut, and
  Vittorio Ferrari.
\newblock Connecting vision and language with localized narratives.
\newblock In \emph{ECCV}, 2020.

\bibitem[Radford et~al.(2021)Radford, Kim, Hallacy, Ramesh, Goh, Agarwal,
  Sastry, Askell, Mishkin, Clark, Krueger, and Sutskever]{radford2021learning}
Alec Radford, Jong~Wook Kim, Chris Hallacy, Aditya Ramesh, Gabriel Goh,
  Sandhini Agarwal, Girish Sastry, Amanda Askell, Pamela Mishkin, Jack Clark,
  Gretchen Krueger, and Ilya Sutskever.
\newblock Learning transferable visual models from natural language
  supervision.
\newblock In \emph{ICML}, 2021.

\bibitem[Reid et~al.(2024)Reid, Savinov, Teplyashin, Lepikhin, Lillicrap,
  Alayrac, Soricut, Lazaridou, Firat, Schrittwieser, et~al.]{reid2024gemini}
Machel Reid, Nikolay Savinov, Denis Teplyashin, Dmitry Lepikhin, Timothy
  Lillicrap, Jean-baptiste Alayrac, Radu Soricut, Angeliki Lazaridou, Orhan
  Firat, Julian Schrittwieser, et~al.
\newblock Gemini 1.5: Unlocking multimodal understanding across millions of
  tokens of context.
\newblock \emph{arXiv preprint arXiv:2403.05530}, 2024.

\bibitem[Rombach et~al.(2022)Rombach, Blattmann, Lorenz, Esser, and
  Ommer]{rombach2022high}
Robin Rombach, Andreas Blattmann, Dominik Lorenz, Patrick Esser, and Bj{\"o}rn
  Ommer.
\newblock High-resolution image synthesis with latent diffusion models.
\newblock In \emph{CVPR}, 2022.

\bibitem[Saharia et~al.(2022)Saharia, Chan, Saxena, Li, Whang, Denton,
  Ghasemipour, Gontijo~Lopes, Karagol~Ayan, Salimans,
  et~al.]{saharia2022photorealistic}
Chitwan Saharia, William Chan, Saurabh Saxena, Lala Li, Jay Whang, Emily~L
  Denton, Kamyar Ghasemipour, Raphael Gontijo~Lopes, Burcu Karagol~Ayan, Tim
  Salimans, et~al.
\newblock Photorealistic text-to-image diffusion models with deep language
  understanding.
\newblock \emph{NeurIPS}, 35:\penalty0 36479--36494, 2022.

\bibitem[Saxon et~al.(2024)Saxon, Jahara, Khoshnoodi, Lu, Sharma, and
  Wang]{saxon2024evaluates}
Michael Saxon, Fatima Jahara, Mahsa Khoshnoodi, Yujie Lu, Aditya Sharma, and
  William~Yang Wang.
\newblock Who evaluates the evaluations? objectively scoring text-to-image
  prompt coherence metrics with t2iscorescore (ts2).
\newblock \emph{arXiv preprint arXiv:2404.04251}, 2024.

\bibitem[Schuster et~al.(2015)Schuster, Krishna, Chang, Fei-Fei, and
  Manning]{schuster2015generating}
Sebastian Schuster, Ranjay Krishna, Angel Chang, Li~Fei-Fei, and Christopher~D.
  Manning.
\newblock Generating semantically precise scene graphs from textual
  descriptions for improved image retrieval.
\newblock In \emph{Workshop on Vision and Language (VL15)}, Lisbon, Portugal,
  September 2015. Association for Computational Linguistics.

\bibitem[Speer(2022)]{robyn_speer_2022_7199437}
Robyn Speer.
\newblock Python library: rspeer/wordfreq: v3.0, September 2022.
\newblock URL \url{https://doi.org/10.5281/zenodo.7199437}.

\bibitem[Stroop(1935)]{stroop1935studies}
J~Ridley Stroop.
\newblock Studies of interference in serial verbal reactions.
\newblock \emph{Journal of experimental psychology}, 18\penalty0 (6):\penalty0
  643, 1935.

\bibitem[Sun et~al.(2023)Sun, Fang, Wu, Wang, and Cao]{sun2023eva}
Quan Sun, Yuxin Fang, Ledell Wu, Xinlong Wang, and Yue Cao.
\newblock Eva-clip: Improved training techniques for clip at scale.
\newblock \emph{arXiv preprint arXiv:2303.15389}, 2023.

\bibitem[Tuo et~al.(2023)Tuo, Xiang, He, Geng, and Xie]{tuo2023anytext}
Yuxiang Tuo, Wangmeng Xiang, Jun-Yan He, Yifeng Geng, and Xuansong Xie.
\newblock Anytext: Multilingual visual text generation and editing.
\newblock \emph{arXiv preprint arXiv:2311.03054}, 2023.

\bibitem[Turc \& Nemade(2023)Turc and Nemade]{turc2023midjourney}
Iulia Turc and Gaurav Nemade.
\newblock Midjourney user prompts \& generated images (250k), 2023.

\bibitem[Vasconcelos et~al.(2024)Vasconcelos, Rashwan, Waters, Walker, Xu, Yan,
  Qian, Luo, Parekh, Bunner, Fei, Garg, Guo, Kajic, Li, Nandwani, Pont-Tuset,
  Onoe, Rosston, Wang, Zhou, Swersky, Fleet, Baldridge, and
  Wang]{vasconcelos2024greedygrowingenableshighresolution}
Cristina~N. Vasconcelos, Abdullah Rashwan, Austin Waters, Trevor Walker, Keyang
  Xu, Jimmy Yan, Rui Qian, Shixin Luo, Zarana Parekh, Andrew Bunner, Hongliang
  Fei, Roopal Garg, Mandy Guo, Ivana Kajic, Yeqing Li, Henna Nandwani, Jordi
  Pont-Tuset, Yasumasa Onoe, Sarah Rosston, Su~Wang, Wenlei Zhou, Kevin
  Swersky, David~J. Fleet, Jason~M. Baldridge, and Oliver Wang.
\newblock Greedy growing enables high-resolution pixel-based diffusion models,
  2024.
\newblock URL \url{https://arxiv.org/abs/2405.16759}.

\bibitem[Villegas et~al.(2022)Villegas, Babaeizadeh, Kindermans, Moraldo,
  Zhang, Saffar, Castro, Kunze, and Erhan]{villegas2022phenaki}
Ruben Villegas, Mohammad Babaeizadeh, Pieter-Jan Kindermans, Hernan Moraldo,
  Han Zhang, Mohammad~Taghi Saffar, Santiago Castro, Julius Kunze, and Dumitru
  Erhan.
\newblock Phenaki: Variable length video generation from open domain textual
  descriptions.
\newblock In \emph{International Conference on Learning Representations}, 2022.

\bibitem[Wang et~al.(2022)Wang, Montoya, Munechika, Yang, Hoover, and
  Chau]{wang2022diffusiondb}
Zijie~J Wang, Evan Montoya, David Munechika, Haoyang Yang, Benjamin Hoover, and
  Duen~Horng Chau.
\newblock Diffusiondb: A large-scale prompt gallery dataset for text-to-image
  generative models.
\newblock \emph{arXiv preprint arXiv:2210.14896}, 2022.

\bibitem[Weyand et~al.(2020)Weyand, Araujo, Cao, and Sim]{weyand2020google}
Tobias Weyand, Andre Araujo, Bingyi Cao, and Jack Sim.
\newblock Google landmarks dataset v2-a large-scale benchmark for
  instance-level recognition and retrieval.
\newblock In \emph{Proceedings of the IEEE/CVF conference on computer vision
  and pattern recognition}, pp.\  2575--2584, 2020.

\bibitem[Xu et~al.(2021)Xu, Ghosh, Huang, Okhonko, Aghajanyan, Metze,
  Zettlemoyer, and Feichtenhofer]{xu2021videoclip}
Hu~Xu, Gargi Ghosh, Po-Yao Huang, Dmytro Okhonko, Armen Aghajanyan, Florian
  Metze, Luke Zettlemoyer, and Christoph Feichtenhofer.
\newblock Videoclip: Contrastive pre-training for zero-shot video-text
  understanding.
\newblock In \emph{Proceedings of the 2021 Conference on Empirical Methods in
  Natural Language Processing}, pp.\  6787--6800, 2021.

\bibitem[Yarom et~al.(2024)Yarom, Bitton, Changpinyo, Aharoni, Herzig, Lang,
  Ofek, and Szpektor]{yarom2024you}
Michal Yarom, Yonatan Bitton, Soravit Changpinyo, Roee Aharoni, Jonathan
  Herzig, Oran Lang, Eran Ofek, and Idan Szpektor.
\newblock What you see is what you read? improving text-image alignment
  evaluation.
\newblock \emph{NeurIPS}, 36, 2024.

\bibitem[Yu et~al.(2022{\natexlab{a}})Yu, Wang, Vasudevan, Yeung,
  Seyedhosseini, and Wu]{yu2022coca}
Jiahui Yu, Zirui Wang, Vijay Vasudevan, Legg Yeung, Mojtaba Seyedhosseini, and
  Yonghui Wu.
\newblock Coca: Contrastive captioners are image-text foundation models.
\newblock \emph{Transactions on Machine Learning Research}, 2022{\natexlab{a}}.
\newblock ISSN 2835-8856.
\newblock URL \url{https://openreview.net/forum?id=Ee277P3AYC}.

\bibitem[Yu et~al.(2022{\natexlab{b}})Yu, Xu, Koh, Luong, Baid, Wang,
  Vasudevan, Ku, Yang, Ayan, et~al.]{yu2022scaling}
Jiahui Yu, Yuanzhong Xu, Jing~Yu Koh, Thang Luong, Gunjan Baid, Zirui Wang,
  Vijay Vasudevan, Alexander Ku, Yinfei Yang, Burcu~Karagol Ayan, et~al.
\newblock Scaling autoregressive models for content-rich text-to-image
  generation.
\newblock \emph{arXiv preprint arXiv:2206.10789}, 2\penalty0 (3):\penalty0 5,
  2022{\natexlab{b}}.

\bibitem[Yuksekgonul et~al.(2022)Yuksekgonul, Bianchi, Kalluri, Jurafsky, and
  Zou]{yuksekgonul2022and}
Mert Yuksekgonul, Federico Bianchi, Pratyusha Kalluri, Dan Jurafsky, and James
  Zou.
\newblock When and why vision-language models behave like bags-of-words, and
  what to do about it?
\newblock In \emph{ICLR}, 2022.

\bibitem[Zapf et~al.(2016)Zapf, Castell, Morawietz, and
  Karch]{zapf2016measuring}
Antonia Zapf, Stefanie Castell, Lars Morawietz, and Andr{\'e} Karch.
\newblock Measuring inter-rater reliability for nominal data--which
  coefficients and confidence intervals are appropriate?
\newblock \emph{BMC medical research methodology}, 16:\penalty0 1--10, 2016.

\bibitem[Zeng et~al.(2022)Zeng, Zhang, and Li]{zeng2022xvlm}
Yan Zeng, Xinsong Zhang, and Hang Li.
\newblock Multi-grained vision language pre-training: Aligning texts with
  visual concepts.
\newblock In Kamalika Chaudhuri, Stefanie Jegelka, Le~Song, Csaba Szepesvari,
  Gang Niu, and Sivan Sabato (eds.), \emph{Proceedings of the 39th
  International Conference on Machine Learning}, volume 162 of
  \emph{Proceedings of Machine Learning Research}, pp.\  25994--26009. PMLR,
  17--23 Jul 2022.
\newblock URL \url{https://proceedings.mlr.press/v162/zeng22c.html}.

\bibitem[Zhai et~al.(2023)Zhai, Mustafa, Kolesnikov, and
  Beyer]{zhai2023sigmoid}
Xiaohua Zhai, Basil Mustafa, Alexander Kolesnikov, and Lucas Beyer.
\newblock Sigmoid loss for language image pre-training.
\newblock \emph{arXiv preprint arXiv:2303.15343}, 2023.

\bibitem[Zhou et~al.(2017)Zhou, Lapedriza, Khosla, Oliva, and
  Torralba]{zhou2017places}
Bolei Zhou, Agata Lapedriza, Aditya Khosla, Aude Oliva, and Antonio Torralba.
\newblock Places: A 10 million image database for scene recognition.
\newblock \emph{IEEE Transactions on Pattern Analysis and Machine
  Intelligence}, 2017.

\bibitem[Zhu et~al.(2023)Zhu, Sun, Wang, Liu, Li, Xiao, and
  Huang]{zhu2023contrastive}
Xiangru Zhu, Penglei Sun, Chengyu Wang, Jingping Liu, Zhixu Li, Yanghua Xiao,
  and Jun Huang.
\newblock A contrastive compositional benchmark for text-to-image synthesis: A
  study with unified text-to-image fidelity metrics.
\newblock \emph{arXiv preprint arXiv:2312.02338}, 2023.

\end{thebibliography}
\clearpage
\appendix
\lstdefinestyle{mystyle}{
    backgroundcolor=\color{backcolour},   
    commentstyle=\color{codegreen},
    keywordstyle=\color{magenta},
    numberstyle=\tiny\color{codegray},
    stringstyle=\color{codepurple},
    basicstyle=\ttfamily\tiny,
    breakatwhitespace=false,         
    breaklines=true,                 
    captionpos=b,                    
    keepspaces=true,                 
    numbers=left,                    
    numbersep=5pt,                  
    showspaces=false,                
    showstringspaces=false,
    showtabs=false,                  
    tabsize=2
}

\section*{Appendices}

\section{Overview}

In the Appendix, we give additional information on the benchmark, human annotation and corresponding results for T2I models, and experimental results for the auto-eval metrics.

\noindent \paragraph{Gecko Benchmark: } For the benchmark, we give further information on how we automatically tag Gecko(R) and semi-automatically generate prompts in Gecko(S) in \Appendix{app:automatictagging} and \ref{app:templates} respectively. We then give more detail about the skill breakdown in Gecko(R) in \Appendix{app:geckor_distribution}. We define and give examples for the sub-skills in Gecko(S) in \Appendix{app:geckosbreakdown}.

\noindent \paragraph{Gecko Metric: } We give further details on the Gecko metric in \Appendix{app:gecko_metric_details}.

\noindent \paragraph{Human Annotation: } For the human annotation, we give additional details of our setup including screenshots of the annotation templates used and qualitative limitations of each setup in \Appendix{app:human_eval_templates}. We further discuss more experimental results comparing inter-annotator agreement and the raw predictions under each template in \Appendix{app:human_eval_add_results}. Finally, we visualise the most and least reliable prompts in \Appendix{app:human_eval_rel_prompts_examples}, giving an intuition for the properties of the prompt that lead to more or less agreement across templates.

\noindent \paragraph{Additional results on T2I models:} We give further results on using the annotated data to (1) compare T2I models by skill in \Appendix{app:t2imodelpercategory}. We also compare how well prompts in TIFA160 are able to discriminate models under our human annotation setup in \Appendix{app:tifa160_dataset_experiments} and find that they are less discriminative.

\noindent \paragraph{Additional results for auto-eval metrics:} We give an intuitive explanation of each task as well as how they can lead to different metric orderings in \Appendix{app:intuitive_explanation_tasks}. We then give additional results for the auto-eval metrics on Gecko2K and Gecko2K-rel, including more correlation results in \Appendix{app_sec:corr_pearson} but we find that conclusions are the same irrespective of how we compute correlation or using the reliable subset or full set.
We give the raw results for the model-ordering evaluation in \Appendix{app:model_ordering_eval} and results for different CLIP variants in \Appendix{app:all_correlations}. Finally, we explore results per skill for different auto-eval metrics in \Appendix{app:metricpercategory}, give additional visualisations in \Appendix{app:additionalvisualisations} and demonstrate that we can use Gecko to evaluate the per-word accuracy of the metric (this is not possible with other auto-eval metrics) in \Appendix{app:results_per_word}.

\section{Gecko2K: More details}
\label{app:benchmark_details}
\begin{figure}[t]
  \centering
  \begin{floatrow}
  \ffigbox{%
    \includegraphics[width=0.5\textwidth]{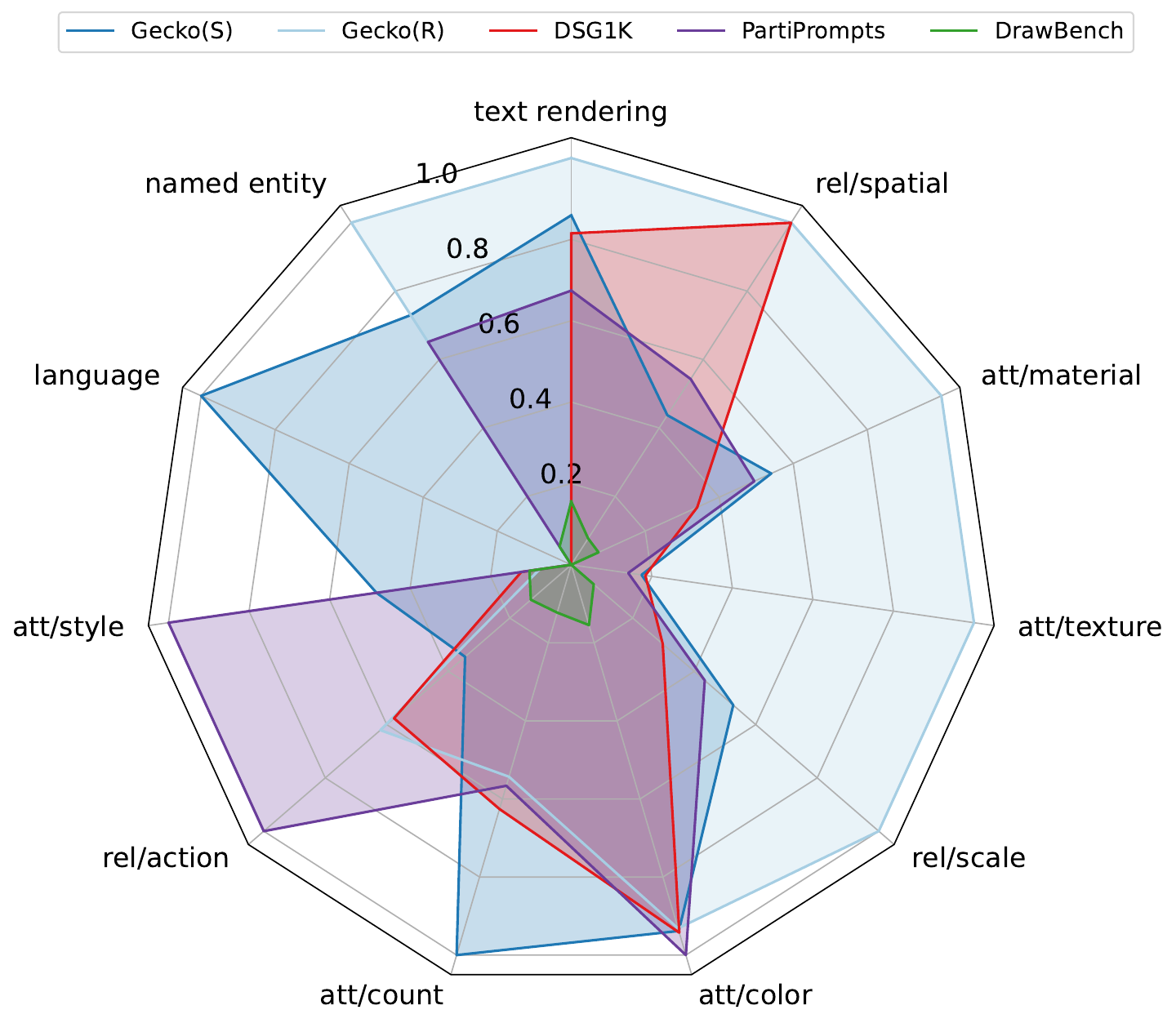}%
    }{
    \caption{\textbf{Distribution of skills.} We visualise the distribution of prompts across different skills for Gecko(S)/(R), DSG1K~\citep{cho2023davidsonian},  PartiPrompts~\citep{yu2022scaling} and DrawBench~\citep{saharia2022photorealistic}. We use automatic tagging and, for each skill, normalise by the maximum number of prompts in that skill over all datasets.  For most skills, Gecko2K has the most number of prompts within that skill.}
  \label{fig:dsg1k-resampled}
  }
\ffigbox{
    \includegraphics[width=0.5\textwidth]{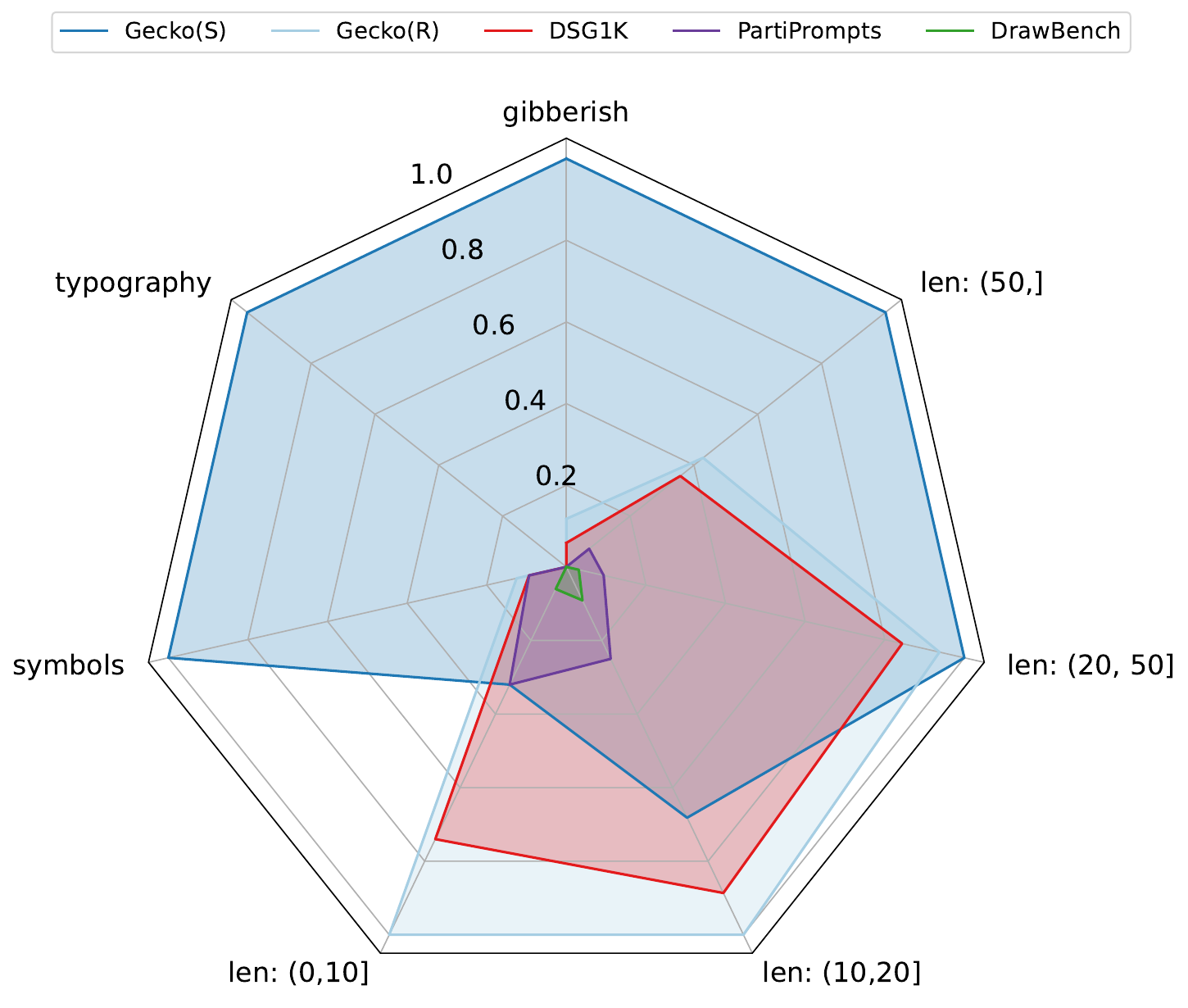}
    }{
    \caption{\textbf{{\em \textsc{text rendering}} skill.} We visualise the distribution of prompts across seven sub-skills explained in \Table{tab:text_rendering_overview} (`len: ...' corresponds to bucketing different lengths of the text to be rendered). We normalise by the maximum number of prompts in the sub-skill (note that we only count unique texts to be rendered). The Gecko(S) dataset fills in much more of the distribution here than other datasets.
    }
 \label{fig:text-rendering-visualisation}
 }
 \end{floatrow}
\end{figure}

As described in \Section{sec: Gecko-R}, we use automatic tagging in order to tag prompts with different skills in Gecko(R). However, this has a few issues: (1) it can be error prone; (2) we are limited by the tagging mechanism in the skills that we tag; (3) we do not tag sub--skills.
As a result, we devise a semi-automatic approach to build Gecko(S) by few-shot prompting an LLM, as discussed in \Section{sec: Gecko-S} and curate a dataset with a number of skills and sub-skills for each skill.
This dataset covers more skills and sub-skills than other datasets, as shown in \Figure{fig:dsg1k-resampled}, \ref{fig:text-rendering-visualisation}.

\begin{figure*}[t]
    \centering
    \includegraphics[width=\linewidth]{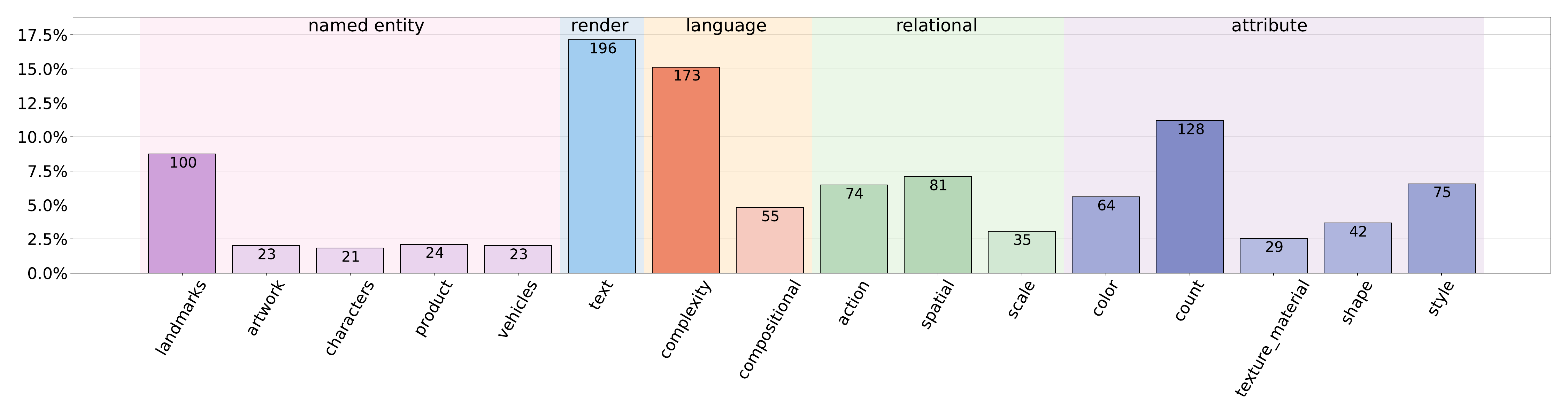}
    \caption{{\bf Overview of Gecko(S).} The set of skills (coloured by the corresponding category) covered by the synthetic prompts. Note that we gather prompts by breaking each skill into sub-skills.}
    \label{fig:dataset_overview}
\end{figure*}

\subsection{Automatic tagging for Gecko(R)}
\label{app:automatictagging}
As mentioned in \Section{sec: Gecko-R}, to obtain a better control for the skill coverage and prompt length, we resampled from the 10 datasets used in DSG1k~\citep{cho2023davidsonian}.
To identify the categories covered in the prompts, we adopted an automatic tagging method similar to that used in DSG1K. This method utilizes a Language Model (LLM) to tag words in the text prompt, as shown in \Listing{alg:auto_tagging_prompt}. The only difference is that we also included named entities and landmarks to be the original categories, such as {\em whole}, {\em part}, {\em state}, {\em color} etc.

\lstset{style=mystyle}

\begin{lstlisting}[language=Python,caption={The prompt used to automatically tag skills given text prompts from the base datasets in DSG1K in order to generate a more balanced Gecko(R).},label={alg:auto_tagging_prompt}]

""" 
id: synthetic_v1_1
input: a man is holding an iPhone.
output: 1 | entity - whole (man)
        2 | entity - named entity (iPhone)
        3 | action - hold (man, iPhone)
     
id: diffusiondb_79
input: an hd painting by Vincent van Gogh. a bunch of zombified mallgoths hanging out at a hot topic store in the mall.
output: 1 | global - style (hd painting)
        2 | global - style (Vincent van Gogh)
        3 | entity - whole (mallgoths)
        4 | attribute - state (mallgoths, zombified)
        5 | other - count (mallgoths, ==bunch)
        6 | entity - whole (hot topic store)
        7 | entity - whole (mall)
        8 | relation - spatial (mallgoths, hot topic store, at)
        9 | relation - spatial (hot topic store, mall, in)
...

id: {image_id}
input: {text_input}
output: {LLM_output}
"""

\end{lstlisting}

\subsection{Prompt distribution in Gecko(R)}
\label{app:geckor_distribution}
\begin{figure}[t]
  \centering

    \begin{subfigure}[t]{0.49\linewidth}
        \includegraphics[width=\linewidth]{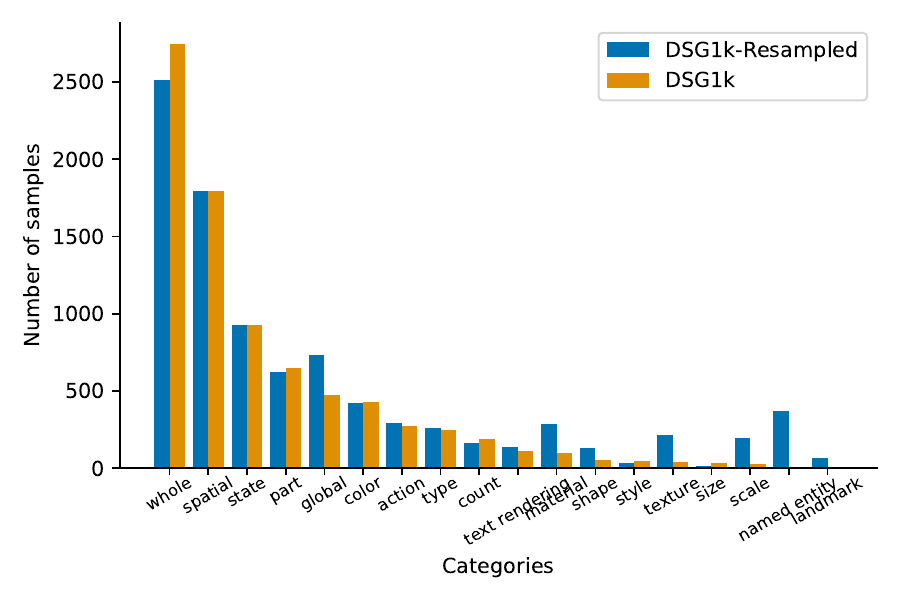}
        \caption{The skill distribution (which is tagged at the word level).}
    \end{subfigure}%
    \begin{subfigure}[t]{0.49\linewidth}
        \includegraphics[width=\linewidth]{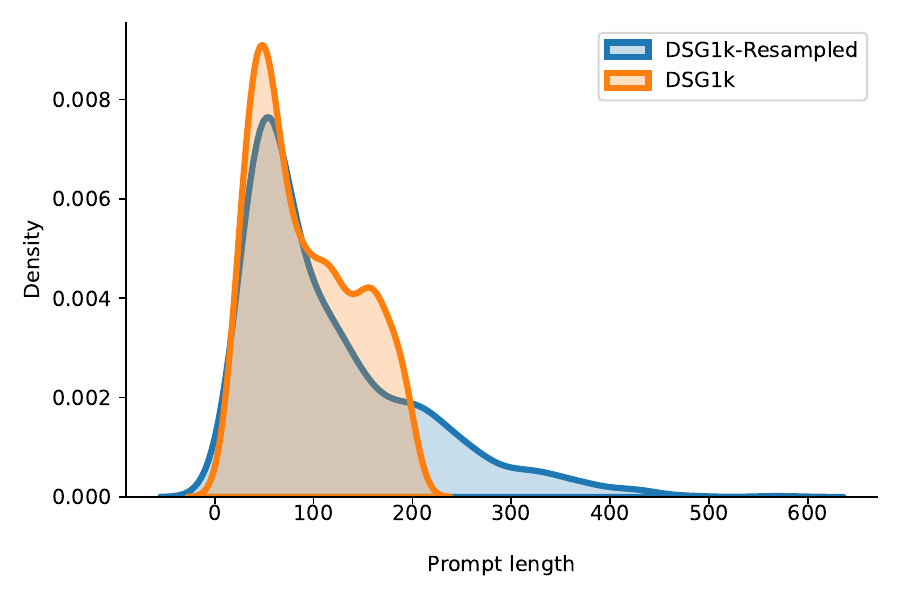}
        \caption{The prompt length distribution.}
    \end{subfigure}%
  \caption{\textbf{Prompt distribution in DSG1K-Resampled(Gecko-R) and DSG1K.}}
  \label{fig:gecko-r-distribution}
\end{figure}

We resample 1000 prompts from the base datasets used in DSG1K and ensure a more uniform distribution over skills and prompt length.
To sample more prompts featuring words from under-represented skills (\eg \textsc{text rendering}, \textsc{shape}, \textsc{named identity} and \textsc{landmarks}), we use automatic tagging in \Appendix{app:automatictagging} to categorize the words in all prompts as pertaining to a given skill. We then resample, assigning higher weights to the under-represented skills. The resulting skill distribution is shown in \Figure{fig:gecko-r-distribution}. Although the resampling increases the proportion of under-represented skills, the overall distribution remains unbalanced. This underscores the necessity of acquiring a synthetic subset with a more controlled and balanced prompt distribution.
To sample long prompts, we eliminate the constraint set in DSG1K~\citep{cho2023davidsonian}, which mandates that the sampled prompts should be shorter than 200 characters. This adjustment results in a more diverse prompt length distribution as shown in \Figure{fig:gecko-r-distribution}.

\subsection{Templates to few-shot an LLM for Gecko(S)}
\label{app:templates}
As discussed in \Section{sec: Gecko-S}, we semi-automatically create prompts for Gecko(S) by few-shot prompting an LLM. We give an example for the \textsc{text rendering} skill in \Listing{alg:few_shot_prompt_llm}.
In short, we define a set of properties based on the sub-skills we want included in our dataset. 
In this case, we define {\em text length} and {\em language} (we use {\em English} and {\em Gibberish} but we note this could be easily extended to more languages). 
We then create examples that have those properties to create our few-shot prompt. 
We can query the LLM as many times as we like to create a distribution of prompts across different text lengths and languages.
We do a similar setup for each of the skills and sub-skills we define below.

\lstset{style=mystyle}

\begin{lstlisting}[language=Python,caption={Sample LLM template.},label={alg:few_shot_prompt_llm}]
""" 
Generate captions for the given text of varying length. Be creative and imagine new settings.

Text length: 20
Language: English
Text: "look at that shadow!"
Caption: shadow of a stone, taken from the point of view of an ant, with the caption "look at that shadow!"

...

Text length: {text_length}
Language: {language}
Text: {LLM_output}
Caption: {LLM_output}
"""

\end{lstlisting}

\subsection{Breakdown by skill/sub-skill in Gecko(S)}
\label{app:geckosbreakdown}

An overview of Gecko(S) is given in \Figure{fig:dataset_overview} and comparisons to other datasets for skills and a given subskill in \Figure{fig:dsg1k-resampled},\ref{fig:text-rendering-visualisation}. In this section we give more information on the skills and sub-skills within Gecko(S).
We provide a detailed breakdown of each prompt sub-skill, including examples and justifications.
Skills and sub-skills are listed in Table~\ref{tab:breakdown_prompts}.
We aim to cover semantic skills, some of which have already been covered in previous work (\eg \emph{shapes, colors} or \emph{counts}), while further subdividing each skill to capture its different aspects and difficulty levels.
By varying the difficulty of the prompts within a challenge we ensure we are testing the models and metrics at different difficulty levels and can find where models and metrics begin to break.

\begin{table}
\caption{Breakdown by skill and sub-skill including examples of prompts.}
\scriptsize
\begin{tabularx}{\linewidth}{ccp{7.3cm}}
\toprule
\textbf{Skill} &
  \textbf{Sub-skill} &
  \textbf{Examples} \\ \midrule
\begin{tabular}[c]{@{}c@{}}Spatial \\ Relationships\\ rel/spatial\end{tabular} &
  Simple &
  \begin{tabular}[c]{@{}l@{}}A cat above a dog.\\ The lemon is in the middle of the apples.\\ A bus is behind a truck going down the highway.\end{tabular} \\ \cmidrule(l){2-3} 
 &
  Composed &
  \begin{tabular}[c]{@{}l@{}}The cat is near the banana. The banana is below the horse. The horse is on\\ the truck.\end{tabular} \\ \midrule
\begin{tabular}[c]{@{}c@{}}Action\\ rel/action\end{tabular} &
  1, 2, or 3 entities &
  \begin{tabular}[c]{@{}l@{}}A bear is running through a field.\\ A basketball is passed to a team member.\end{tabular} \\ \cmidrule(l){2-3} 
 &
  Reverse actions &
  \begin{tabular}[c]{@{}l@{}}A ladybug is riding on the back of a flying unicorn.\\ A koala climbs a tree, an eagle stands on the branch and a penguin flies\\ overhead.\end{tabular} \\ \midrule
\begin{tabular}[c]{@{}c@{}}Scale\\ rel/scale\end{tabular} &
  Single object &
  \begin{tabular}[c]{@{}l@{}}A small ship in a bottle.\\ A giant couch in a field.\end{tabular} \\ \cmidrule(l){2-3} 
 &
  Comparative &
  \begin{tabular}[c]{@{}l@{}}A garlic is next to an onion and a tomato on a cutting board. The onion is\\ larger than the garlic. The tomato is smaller than the garlic.\end{tabular} \\ \cmidrule(l){2-3} 
 &
  Same size &
  \begin{tabular}[c]{@{}l@{}}The table is the same size as the cake.\\ The mouse is the same size as the dragon.\end{tabular} \\ \midrule
\begin{tabular}[c]{@{}c@{}}Counting\\ att/count\end{tabular} &
  Simple modifier &
  \begin{tabular}[c]{@{}l@{}}2 cats. \\ Four lemurs.\end{tabular} \\ \cmidrule(l){2-3} 
 &
  Additive &
  \begin{tabular}[c]{@{}l@{}}5 burgers and one bonsai.\\ 1 baobab, 2 cats and 3 dogs.\end{tabular} \\ \cmidrule(l){2-3} 
 &
  \begin{tabular}[c]{@{}c@{}}Quantifiers and\\ negations\end{tabular} &
  \begin{tabular}[c]{@{}l@{}}Some shirts and some pizzas. There are more shirts than pizzas.\\ An image with fewer dogs than cats.\\ An image with no flowers in the vase.\end{tabular} \\ \midrule
\begin{tabular}[c]{@{}c@{}}Shape\\ att/shape\end{tabular} &
  Basic shapes &
  \begin{tabular}[c]{@{}l@{}}A line.\\ An octagon.\end{tabular} \\ \cmidrule(l){2-3} 
 &
  Composed shapes &
  \begin{tabular}[c]{@{}l@{}}A star-shaped cookie.\\ Strawberries arranged in a heart shape.\end{tabular} \\ \cmidrule(l){2-3} 
 &
  \begin{tabular}[c]{@{}c@{}}Hierarchical\\ shapes\end{tabular} &
  \begin{tabular}[c]{@{}l@{}}A square made of smaller letters g.\\ A smiley face made of strawberries.\end{tabular} \\ \midrule
\begin{tabular}[c]{@{}c@{}}Text\\ Rendering\\ render/text\end{tabular} &
  Rendering &
  \begin{tabular}[c]{@{}l@{}}A creature with a clock shaped head with the words\\ "flimflam, bishbash, gorp" written on it. The creature has a small body\\ and two legs, and is pointing at the ground. There are two clocks on the\\ ground, one showing the time of 7:24 and the other showing 3:30 p.m.\end{tabular} \\ \cmidrule(l){2-3} 
 &
  Numerical &
  \begin{tabular}[c]{@{}l@{}}equation of "3+4 = 7" etched into a rock.\\ "Lorem ipsum dolor sit amet, \$\%\&*(), 12345 + adipiscing elit" written\\ on a chalkboard with a piece of chalk next to it.\end{tabular} \\ \cmidrule(l){2-3} 
 &
  Font &
  \begin{tabular}[c]{@{}l@{}}"congratulations" written in fancy decorative cursive font on an antique\\ 1920's typewriter. \\ "happiness" written in decorative font with a happy face next to it.\end{tabular} \\ \midrule
\begin{tabular}[c]{@{}c@{}}Color\\ att/color\end{tabular} &
  Simple colors &
  \begin{tabular}[c]{@{}l@{}}A pink salad.\\ A grey vase.\end{tabular} \\ \cmidrule(l){2-3} 
 &
  Composed expressions &
  \begin{tabular}[c]{@{}l@{}}A pink unicorn, a white airplane, and a green potato.\\A yellow couch and a green cookie.\end{tabular} \\ \cmidrule(l){2-3} 
 &
  Colors and abstract shapes &
  \begin{tabular}[c]{@{}l@{}}A red rectangle on a green background. \\A pink circle on a green background.\end{tabular} \\ \cmidrule(l){2-3} 
 &
  Descriptive color terms &
  \begin{tabular}[c]{@{}l@{}}A pastel coloured train passing through the station.\\A rainbow-colored bicycle leaning against a wall.\end{tabular} \\ \cmidrule(l){2-3} 
 &
  Stroop &
  \begin{tabular}[c]{@{}l@{}}Text saying "yellow" in blue letters.\\ Text saying "black" in green letters.\end{tabular} \\ \bottomrule
\end{tabularx}

\label{tab:breakdown_prompts}
\end{table}
\begin{table}[]
\scriptsize
\begin{tabularx}{\linewidth}{ccp{7.3cm}}
\toprule
\textbf{Skill} &
  \textbf{Sub-skill} &
  \textbf{Examples} \\ \midrule
\begin{tabular}[c]{@{}c@{}}Surface\\ Characteristics\\ att/texture+material\end{tabular} &
  Texture only &
  \begin{tabular}[c]{@{}l@{}}A fluffy floor in the bathroom.There is a silky fabric on a bumpy\\ couch in the room.\end{tabular} \\ \cmidrule(l){2-3} 
 &
  Material only &
  There is a metal lime in the bowl.A paper snake on the table. \\ \cmidrule(l){2-3} 
 &
  Combined &
  \begin{tabular}[c]{@{}l@{}}There is a soft floor made of wax and and a shiny silver table in the room.\\ A glossy diamond road.\end{tabular} \\ \midrule
\begin{tabular}[c]{@{}c@{}}Style\\ att/style\end{tabular} &
  \begin{tabular}[c]{@{}c@{}}style\end{tabular} &
  \begin{tabular}[c]{@{}l@{}}A brightly colored canal in Venice, by Canaletto\\ A cartoon of a cat by Goya.\end{tabular} \\ \cmidrule(l){2-3} 
 &
  Visual medium &
  \begin{tabular}[c]{@{}l@{}}A sketch of a drawing of a flower in a pot.\\ A glass vase in the style of el greco and the impressionists.\end{tabular} \\ \midrule
\begin{tabular}[c]{@{}c@{}}Language\\ Complexity\\ lang/complexity\end{tabular} &
  Negation &
  \begin{tabular}[c]{@{}l@{}}A pencil sharpener without any pencils in it.\\ A belt buckle with no belt.\\ A leaflet with no text on it.\end{tabular} \\ \cmidrule(l){2-3} 
 &
  Long prompt &
  \begin{tabular}[c]{@{}l@{}}An outdoors top-down view of purple sidewalk chalk on a concrete sidewalk\\ reading, "Fear/ of/ Chores". The left side of the concrete slab has green algae\\ growth that fades to the right. Small bits of smashed acorns are scattered\\ across the slab.\end{tabular} \\ \cmidrule(l){2-3} 
 &
  True paraphrase &
  \begin{tabular}[c]{@{}l@{}}The giraffe feeds the cat.\\ The bystanders watch the dog.\end{tabular} \\ \cmidrule(l){2-3} 
\multicolumn{1}{l}{} &
  False paraphrase &
  \begin{tabular}[c]{@{}l@{}}The snake observes the kangaroo.\\ The horse looks at the deer.\end{tabular} \\ \midrule
\begin{tabular}[c]{@{}c@{}}Compositional\\ Language\\ lang/compositional\end{tabular} &
  \begin{tabular}[c]{@{}c@{}}Vary number \\ of\\ entities \\ \& attributes\end{tabular} &
  \begin{tabular}[c]{@{}l@{}}An orange metal train.\\ A plastic couch, a cyan blueberry, and 2 plates.\\ 3 wooden pencils and 1 plastic fly.\\ A brown plastic bus, a yellow plastic vase, and a yellow wooden salad.\end{tabular} \\ \midrule
\begin{tabular}[c]{@{}c@{}}Named\\ Entities\end{tabular} &
  \begin{tabular}[c]{@{}c@{}}Landmarks\\ ne/landmarks\end{tabular} &
  \begin{tabular}[c]{@{}l@{}}Ulvetanna Peak, Queen Maud Land, Antarctica during sunrise.\\ Ashikaga Flower Park with beautiful wisteria in shades of violet and white.\\ Burj Al Arab Jumeirah hotel with fireworks in the night and glowing water\\ around.\end{tabular} \\ \cmidrule(l){2-3} 
 &
  \begin{tabular}[c]{@{}c@{}}Animal Characters\\ ne/characters\end{tabular} &
  \begin{tabular}[c]{@{}l@{}}Grumpy Cat is sitting on a couch with a catnip toy.\\ A cartoon of Laika playing in the snow.\\ the lion cub named Simba is catching a ball.\end{tabular} \\ \cmidrule(l){2-3} 
 &
  \begin{tabular}[c]{@{}c@{}}Vehicles\\ ne/vehicles\end{tabular} &
  \begin{tabular}[c]{@{}l@{}}A BMW M3 is on the road.\\ A Ferrari is driving through roads in an Italian landscape.\\ A Opel Ampera is upside down.\end{tabular} \\ \cmidrule(l){2-3} 
 &
  \begin{tabular}[c]{@{}c@{}}Products\\ ne/product\end{tabular} &
  \begin{tabular}[c]{@{}l@{}}A bottle of Irn-Bru is sitting on a shelf.\\ A Gucci bag with a red and white striped pattern.\\ A Samsung Galaxy S III is on a car seat.\end{tabular} \\ \cmidrule(l){2-3} 
 &
  \begin{tabular}[c]{@{}c@{}}Artwork\\ ne/artwork\end{tabular} &
  \begin{tabular}[c]{@{}l@{}}Charles IV of Spain and His Family is being painted on an easel.\\ A painting of The Milkmaid hangs on the wall of a living room.\\ A painting of Bacchus and Ariadne hanging in a stone building.\end{tabular} \\ \bottomrule

\end{tabularx}
\end{table}

Each skill (such as \textsc{shape}, \textsc{color}, or \textsc{numerical}) is divided into sub-skills, so that prompts within that sub-skill can be distinguished based on difficulty or, if applicable, some other criteria that is unique to that sub-skill (\ie prompts inspired by literature in psychology).
We create a larger number of examples and subsample to create our final 1K set of prompts to be labelled.

\subsubsection{Spatial Relationships} This skill captures a variety of spatial relationships (such as \emph{above, on, under, far from, etc.}) between two to three objects. In the most simple case, we measure a model's ability to understand common relationships between two objects. The difficulty is increased by combining simpler entities and requiring the ability to reason about implicit relationships.
We use an LLM as described in \Section{app:templates} to create these prompts, subsample and manually verify prompts are reasonable.

\subsubsection{Action} This skill examines whether the model can bind the right action to the right object, including unusual cases where we flip the subject and the object (\ie \emph{Reverse actions}). An example of a reverse setup is that we swap the entities in `A penguin is diving while a dolphin swims' and create `A penguin is swimming while a dolphin is diving'.  We vary the difficulty by increasing the number of entities.
We use an LLM as described in \Section{app:templates} to create these prompts, subsample and manually verify prompts are reasonable.

\subsubsection{Scale} We measure whether the model can reason about scale cues referring to commonly used descriptors such as \emph{small, big} or \emph{massive}. To reduce ambiguity, we typically refer to two objects, so that they can be compared in size.
We test the ability to implicitly reason about scales by having \emph{Comparative} prompts that contain several statements about objects, their relations and sizes.
We use an LLM as described in \Section{app:templates} to create these prompts, subsample and manually verify prompts are reasonable.

\subsubsection{Counting} The simplest sub-skill \emph{Simple modifier} contains a number (digits such as ``2'', ``3'' or numerals such as ``two'', ``three'') and an entity.
When selecting a vocabulary of words, we aimed to include words that occur less frequently in ordinary language (for example, ``lemur'' and ``seahorse'' occur less frequently than ``dog'' and ``cat'')~\citep{robyn_speer_2022_7199437}. We focus on numbers 1---10, or 1---5 in more complex cases. Complexity is introduced by combining simple prompts containing just one attribute into compositional prompts containing several attributes. For example, simple prompts ``1 cat'' and ``2 dogs'' are combined into a single prompt ``1 cat and 2 dogs'' in the sub-skill \emph{Additive}.
We also test approximate understanding of quantities based on linguistic concepts of \emph{many} and \emph{few} in the \emph{Quantifiers and negations} sub-skill.

\subsubsection{Shape} We test for basic and composed shapes, where composed shapes include objects arranged in a certain shape. \emph{Hierarchial shapes} are of the following type: ``The letter H made up of smaller letters S''. This kind of challenge is used to study spatial cognition and the trade-off between global and local attention in literature on higher-order cognition~\citep{DELIS1986205}.
We use an LLM as described in \Section{app:templates} to create these prompts, subsample and manually verify prompts are reasonable.

\subsubsection{Text Rendering} We investigate a model's ability to generate text (both semantically meaningful and meaningless), including text of different lengths. We further test for the ability to generate symbols and numbers, as well as different types of fonts.
We use an LLM as described in \Section{app:templates} to create these prompts, subsample and manually verify prompts are reasonable.

\subsubsection{Colour}
The simplest prompts in this skill include basic colours bound to objects.
As before, we introduce complexity by combining several simpler prompts (either two or three objects bound with a colour attribute).
To include diversity of possible colour attributes, we also test descriptive colour terms such as ``pastel'' or ``rainbow-coloured''. Finally, the sub-skill \emph{stroop} contains prompts of the type ``Text saying `blue' in green letters'' similar to the incongruent condition in the Stroop task~\citep{stroop1935studies} used to study interference between different cognitive processes.

\subsubsection{Surface Characteristics} Surface characteristics include texture and material. We first test for each sub-skill individually, and then combined. Generally, some prompts in this skill can be difficult to visualise as they might include descriptions that are typically of tactile nature (``abrasive'' or ``soft'').

\subsubsection{Style} We divide prompts into two sub-skills: one depicting a style of an artist, and another to capture different visual mediums (such as \emph{photo, stained glass} or \emph{ceramics}). 
We use an LLM as described in \Section{app:templates} to create these prompts, subsample and manually verify prompts are reasonable.

\subsubsection{Named Entity}

This skill evaluates a model's knowledge of the world through named entities, focusing on specific entity types such as \emph{landmarks}, \emph{artwork}, \emph{animal characters}, \emph{products}, and \emph{vehicles}, which are free of personally identifiable information (PII).

For the landmark class, we choose landmarks from the Google Landmarks V2 dataset and ensure we cover different continents and choose landmarks with high popularity \citep{weyand2020google}.
Given this set of landmarks, we use an LLM as described in \Section{app:templates} to create these prompts, subsample and manually verify prompts are reasonable.

For the other classes, to curate diverse named entities, we first gather candidates from Wikidata using SPARQL queries.
A simple query (e.g., \texttt{instance of (P31)} is \texttt{painting (Q3305213)}) might yield an excessively large number of candidates.
Therefore, we impose conditions to narrow down the query responses.
See our criteria below.
\begin{description}[align=left, font=\it]
\vspace{-5pt}
\setlength\itemsep{0.05cm}
\item [Artwork]: Created before the 20th century; any media; any movement
\item [Animal Characters]: Anthropomorphic/fictional animals; real animals with names
\item [Products]: Electric devices; food/beverage; beauty/health
\item [Vehicle]: Automobiles; aircraft
\vspace{-5pt}
\end{description}
Once we have a candidate set for each entity class, we focus on selecting reasonably popular entities that are widely recognised and appropriate to present to models.
We assess popularity using the number of incoming links to and contributors on their English Wikipedia pages as proxies \citep{Mor_Geva_2021}.
Finally, we manually curate the final set of named entities, selecting them based on their ranked popularity scores.

\subsubsection{Language complexity}
We evaluate models on prompts with ``tricky'' language structure / wording. For this skill, we include 4 sub-skills: \emph{negation}, \emph{long prompt}, and \emph{true / false phrases}. We sampled 19 prompts for \emph{negation} from LVIS~\citep{gupta2019lvis}, COCO stuff~\citep{caesar2018cvpr}, and MIT Places~\citep{zhou2017places}; 58 \emph{true paraphrases} and 58 \emph{false paraphrases} from BLA~\citep{chen2023bla}; and finally crowdsourced 38 for \emph{long prompt} with the help from English major raters. It should be noted that while we do not cover the entire spectrum of complex language (e.g. passives, coordination, complex relative clausals, etc.), the subcategories included cover the most prominent pain points of image generation models per our experimentation.

We also include a language complexity metric which can be run over all prompts.
Here we treat language complexity from two perspectives -- semantic and syntactic.
\begin{itemize}
\item \emph{Semantic complexity}. The quantity of semantic elements included in a prompt.
\item \emph{Syntactic complexity}. The level of complexity of the syntactic structure of a prompt.
\end{itemize}
Concretely, we define \emph{semantic complexity} as the number of entities extracted from a prompt. Taking the visual relevance of the task into account, we apply \emph{Stanford Scene Graph Parser}~\citep{schuster2015generating} for entity extraction and count the number of unique entities as the proxy for semantic complexity.
For \emph{syntactic complexity}, we implement a modified variant of \citet{Ohta2013} to look for the deepest central branch in the dependency tree of a prompt (pseudo-code below)~\footnote{As an implementation note, we implemented \citet{schuster2015generating} and \citet{Ohta2013} with SpaCy 2~\citep{spacy2} as the workhorse parsing backend.} to gauge the complexity of its syntactic structure.

\begin{lstlisting}[language=Python]
def central_depth(node) -> Tuple[int, int]:
  return (max(central_depth(child)[1]+1 in node.children if child.position < node.position),
          max(central_depth(child)[0]+1 in node.children if child.position > node.position))
\end{lstlisting}

\section{Gecko Metric: More details}
\label{app:gecko_metric_details}

\subsection{LLM prompting for generating coverage}
\lstset{style=mystyle}

\begin{lstlisting}[language=Python,caption={Sample LLM template for generating word coverage.},label={alg:few_shot_prompt_llm_coverage}]
""" 
Given a image description, label the visually groundable words in the description, and a score indicating how visually groundable it is.
Classify each word into a type (entity, activity, attribute, counting, color, material, spatial, location, shape, style, other).

Description:
Portrait of a gecko wearing a train conductor's hat and holding a flag that has a yin-yang symbol on it. Woodcut.
The visual-groundable words and their scores are labelled below:
{1}[Portrait, style, 0.8] of {2}[a, count, 1.0] {3}[gecko, entity, 1.0] {4}[wearing, activity, 1.0] {5}[a, count, 1.0] {6}[train conductor's hat, entity, 1.0] and {7}[holding, entity, 1.0] {8}[a, count, 1.0] {9}[flag, entity, 1.0] that has {10}[a yin-yang symbol, entity, 1.0] on it. {11}[Woodcut, material, 1.0].


Description:
square blue apples on a tree with circular yellow leaves
The visual-groundable words and their scores are labelled below:
{1}[square, shape,1.0] {2}[blue, color, 1.0] {3}[apples, entities, 1.0] {4}[on, spatial, 1.0] {5}[a, count, 1.0] {6}[tree, entity, 1.0] with {7}[circular, shape, 1.0] {8}[yellow, color, 1.0] {9}[leaves, entity, 1.0]

Description:
A small dog running on a beach happily on a sunny day
The visual-groundable words and their scores are labelled below:
{1}[A, count, 1.0] {2}[small, attribute, 0.3] {3}[dog, entity, 1.0] {4}[running, activity, 1.0] {5}[on, spatial, 1.0] {6}[a, count, 0.0] {7}[beach, entity, 1.0] happily on {8}[a, count, 0.0] {9}[sunny, attribute, 0.5] day.

Description:
acrylic drawing, illustration, multiple mushrooms and pink jello, naive, flat, sketchy, purple background.
The visual-groundable words and their scores are labelled below:
{1}[crylic drawing, style, 1.0], {2}[illustration, style, 0.2], {3}[multiple, count, 0.5] {4}[mushrooms, eneity, 1.0] and {5}[pink, color, 1.0] {6}[jell, entity, 1.0], {7}[naive, attribute, 0.1], {8}[flat, attribute, 0.8] {9}[sketchy, style, 0.8] {10}[purple, color, 1.0] {11}[background, entity, 1.0].

Description:
a girl with many braids, riding away on her bike through the city, children's book cover illustration, detailed background, vibrant colors
The visual-groundable words and their scores are labelled below:
{1}[a, count, 1.0] {2}[girl, entity, 1.0], with {3}[many, counts, 0.8] {4}[braids, entity, 1.0], {5}[riding away, activity, 1.0] on her {6}[bike, entity, 1.0] through the {7}[city, place, 0.8], {8}[children's book cover illustration, style, 0.8], {9}}[detailed background, entity, 1.0] {10}[vibrant colors, colors, 1.0].

Description:
"""

\end{lstlisting}

\subsection{LLM prompting for generating QAs}

\begin{lstlisting}[language=Python,caption={Sample LLM template for generating QAs.},label={alg:few_shot_prompt_llm_qa}]
""" 
Given a image description, generate one or two multiple-choice questions that verifies if the image description is correct.
Classify each concept into a type (object, human, animal, food, activity, attribute, counting, color, material, spatial, location, shape, other), and then generate a question for each type.

Description:
A man posing for a selfie in a jacket and bow tie.
The visual-groundable words and their scores are labelled below:
A {1}[Man, human] {2}[posing, activity] for a {3}[selfie, object] in a {4}[jacket, object] and a {5}[bow tie, object].
Generated questions and answers are below:
About {1}:
Q: is there a man in the image?
Choices: yes, no
A: yes
About {2}:
Q: is the man posing for the selfie?
Choices: yes, no
A: yes
About {3}:
Q: is the man taking a selfie?
Choices: yes, no
A: yes
About {4}:
Q: is the man wearing a jacket?
Choices: yes, no
A: yes
About {5}:
Q: is the man wearing a bow tie?
Choices: yes, no
A: yes


Description:
A horse and several cows feed on hay.
The visual-groundable words and their scores are labelled below:
A {1}[horse, animal] and {2}[several, count] {3}[cows, animal] {4}[feed, activity] on a {5}[hay, object].
Generated questions and answers are below:
About {1}:
Q: is there a horse?
Choices: yes, no
A: yes
About {2}:
Q: are there several cows?
Choices: yes, no
A: yes
About {3}:
Q: are there cows?
Choices: yes, no
A: yes
About {4}:
Q: are the horse and cows feeding on hay?
Choices: yes, no
A: yes
About {5}:
Q: is there hay?
Choices: yes, no
A: yes

Description:
...

Description:
"""

\end{lstlisting}

\subsection{Discussion}
Here we discuss the potential limitations of the models we rely on and how we mitigate those issues, as well as give quantitative results around how impactful those potential issues are in practice. Note that we treat these models as black boxes (we do not consider fine-tuning or further calibration) which gives the benefit, as shown in \Table{tab:corr}, that as models improve (e.g.~by swapping \texttt{PALM-2/PALI} for \texttt{Gemini Flash}), so too will our metric, with no further effort.

\paragraph{Potential issue 1: Hallucination of the QA model.} The QA model could hallucinate, generating erroneous questions that are not grounded in the prompt, leading to worse performance. There are two factors that help mitigate this: (1) the use of the NLI model and (2) the ability of the LLM used to not hallucinate in the first place. We quantify the accuracy of the NLI model. To do so, we randomly chose 1.8K question/answer pairs from Gecko(R)/(S) and annotated whether they are hallucinations or not. We find that the NLI model is $\sim$93\% accurate on the \texttt{PALM-2} setup. The utility of the NLI model is further validated in \Table{tab:gecko_ablation}, which shows that adding NLI filtering improves results. We also evaluate how often the NLI model removes questions for the older \texttt{PALM-2} model as opposed to \texttt{Gemini Flash}. We find that the NLI model removes 13\% of questions for \texttt{PALM-2} but only 2\% for \texttt{Gemini Flash}, indicating that a better model will hallucinate less. This result validates the finding in \Table{tab:corr} that as models improve, so too does our metric. In all, we find that hallucination can be mitigated effectively through the use of the NLI model and that with better LLMs, the impact of this issue will be diminished. 

\paragraph{Potential issue 2: Bias of the VQA model.} The VQA model could be biased or give poor scores. We note several factors that indicate the scores are useful and that these models do not suffer from severe bias. First, \cite{cho2023davidsonian} have validated that \texttt{PALI} and even weaker models achieve high accuracy on such VQA style questions. We also use the largest model -- \texttt{PaLI-17B}; prior work has found that versions with the largest language component are better calibrated \citep{kostumov2024uncertainty}. Second, we break down the VQA score into multiple questions and so we are more robust to incorrect scores arising from a single question. Third, we check for a strong `yes' bias, which has been found in prior work evaluating VQA models \citep{agrawal2018don} and is relevant, as many of our QAs are yes/no questions due to the few-shot prompt. We evaluate how \texttt{Gemini Flash} responds to `blind' questions: given no image and a question, will the VQA model always output a given answer. We find that we obtain  20\% `yes' answers and 80\% `no', indicating that the model does not have a strong `yes' bias. We also note that if a model is biased, it is equally biased for any input, which means that the relative comparison is valid.

Finally, our comprehensive results demonstrate that this potential bias is {\em not} a problem.  First, we ablate the utility of the scoring component in \Table{tab:gecko_ablation} and find that it improves results. Second, we find that our metric performs well, obtaining 72/79\% agreement with human preference and on average 0.53 correlation (see \Table{tab:corr}). If the VQA model were terribly biased, it would {\em ignore} visual input, leading to chance performance on the \pairscore~task. Thus the high performance on our comprehensive benchmark, as well as our ablations, demonstrate the utility of leveraging the scores from the VQA models.

\clearpage

\section{Human annotation: more details and experiments}

\subsection{Annotation templates}\label{app:human_eval_templates}

\noindent {\bf Likert scale.} We follow the template of \citet{cho2023davidsonian} and collect human judgements using a 5-point Likert scale by asking the annotators ``How consistent is the image with the prompt?'' where \emph{consistency} is defined as how well the image matches the text description. Annotators are asked to choose a rating from the given scale, where 1 represents \emph{inconsistent} and 5 \emph{consistent}, or a sixth \textit{Unsure} option for cases where the text prompt is not clear.  Choosing this template enables us to compare our results with previous work, but does not provide fine-grained, word-level alignment information. Moreover, while Likert provides a simple and fast way to collect data, challenges such as defining each rating especially when used without textual description (\eg, what 2 refers to in terms of image--text consistency), can lead to subjective and biased scores \citep{heo2022comparison,liang2020beyond}.

\begin{figure}[h]
    \centering
    \includegraphics[width=0.7\textwidth]{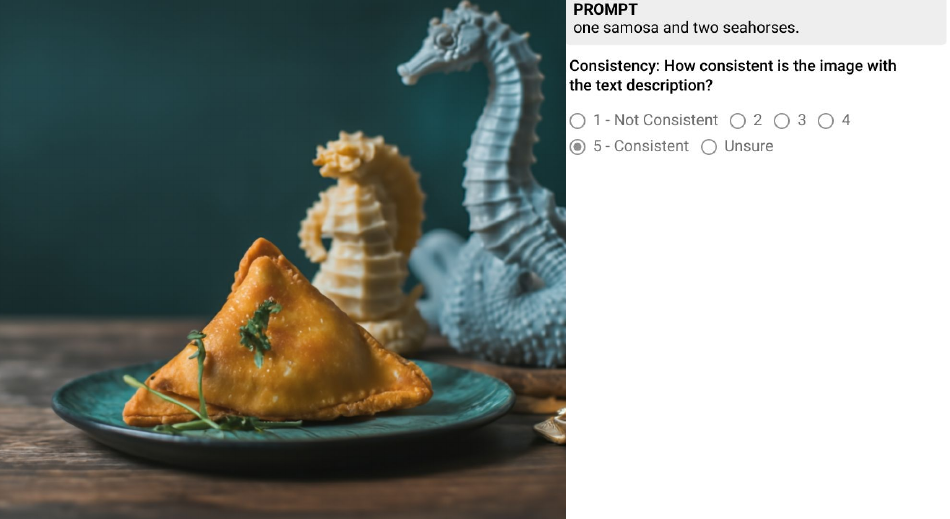}
    \caption{{\bf Likert annotation template user interface.} Example depicting the interface shown to the annotators when performing evaluation tasks with the Likert template. Raters are given the prompt and image and asked to rate on a 5-point scale how consistent the image is with respect to the prompt. An \textit{Unsure} option is also given the annotators.}
    \label{fig:human_eval_screenshot_likert}
\end{figure}

\noindent {\bf Word-level alignment (WL).} To collect word-level alignment annotations, we use the template of \citet{liang2023rich} and define an overall image--text alignment score using the word-level information.
Given a text--image pair, raters are asked to annotate each word in the prompt as \textit{Aligned}, \textit{Unsure}, or \textit{Not aligned}. Note that for each text--image pair under the evaluation, the number of effective annotations a rater must perform is equal to the number of words in the text prompt. Although potentially more time consuming than the Likert template, we find that annotators spend $\sim$30s more to rate a prompt--image pair with WL than Likert.

We compute a score for each prompt--image pair per rater by aggregating the annotations given to each word. A final score is then obtained by averaging the scores of 3 raters. 

\begin{figure}[h]
    \centering
    \includegraphics[width=0.7\textwidth]{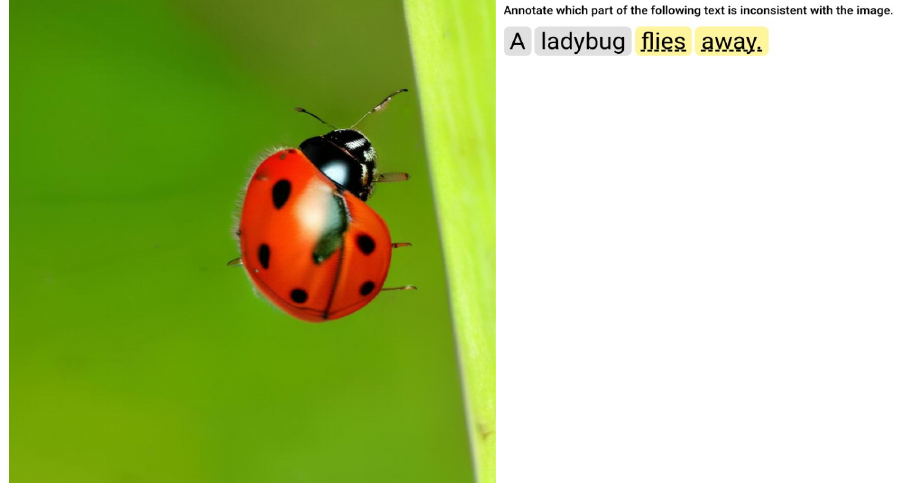}
    \caption{{\bf Word-level annotation template user interface.} Example depicting the interface shown to the annotators when performing evaluation tasks with the WL template. Raters are asked to click on words they find are not aligned with image, and double click on the words where they are unsure.}
    \label{fig:human_eval_screenshot_wl}
\end{figure}

\noindent {\bf DSG(H).} We also use the annotation template of \citet{cho2023davidsonian} that asks the raters to answer a series of questions for a given image, where the questions are generated automatically for the given text prompt as discussed in \citet{cho2023davidsonian}.

In addition, raters can mark a question as \textit{Invalid} in case a question contradicts another one.
The total number of \textit{Invalid} ratings per evaluated generative model is given in  \Appendix{app:human_eval_templates}. Annotators could also rate a question as \emph{Unsure}, in cases where they do not know the answer or find the question subjective or not answerable based on the given information.

For a given prompt, the number of annotations a rater must complete is given by the number of questions. We calculate an overall score for an image--prompt pair by aggregating the answers across all questions, and then averaging this number across raters to obtain the final score.  

\begin{figure}[h]
    \centering
    \includegraphics[width=0.7\textwidth]{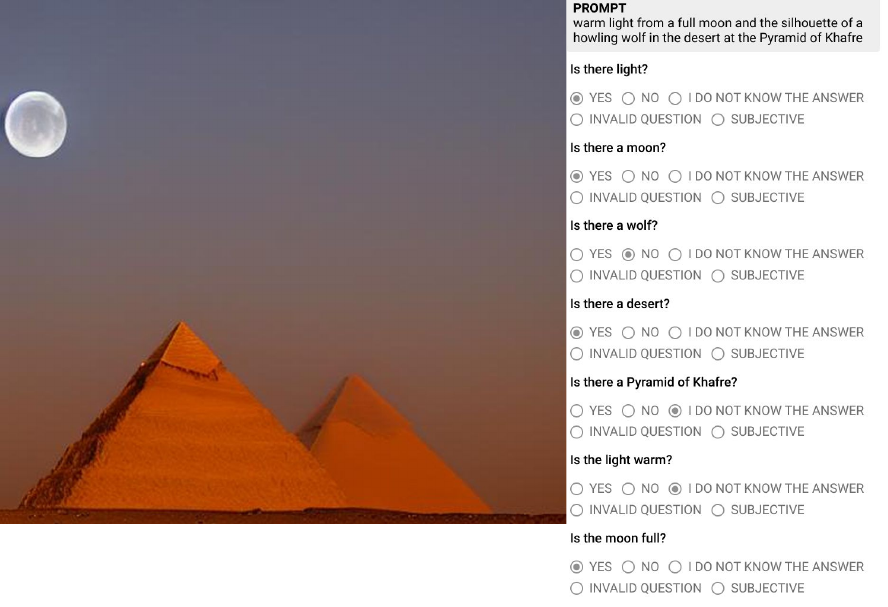}
    \caption{{\bf DSG(H) annotation template user interface.} Example depicting the interface shown to annotators when performing evaluation tasks with the DSG(H) template. Raters are given the image, prompt, and respective automatically generated questions. There are 5 options for answering each question. In our analysis, both \textit{I do not know the answer} and \textit{Subjective} answers are considered as \textit{Unsure}.}
    \label{fig:human_eval_screenshot_dsg}
\end{figure}

\noindent {\bf Side-by-side (SxS).} We consider a template in which pairs of images are directly compared. The annotators see two images from two models side-by-side and are asked to choose the image that \textit{is better aligned} with the prompt or select \textit{Unsure}. We obtain a score for each comparison by computing the majority voting across all 3 ratings. In case there is a tie, we assign \emph{Unsure} to the final score of an image--prompt pair.

\begin{figure}[h]
    \centering
    \includegraphics[width=\textwidth]{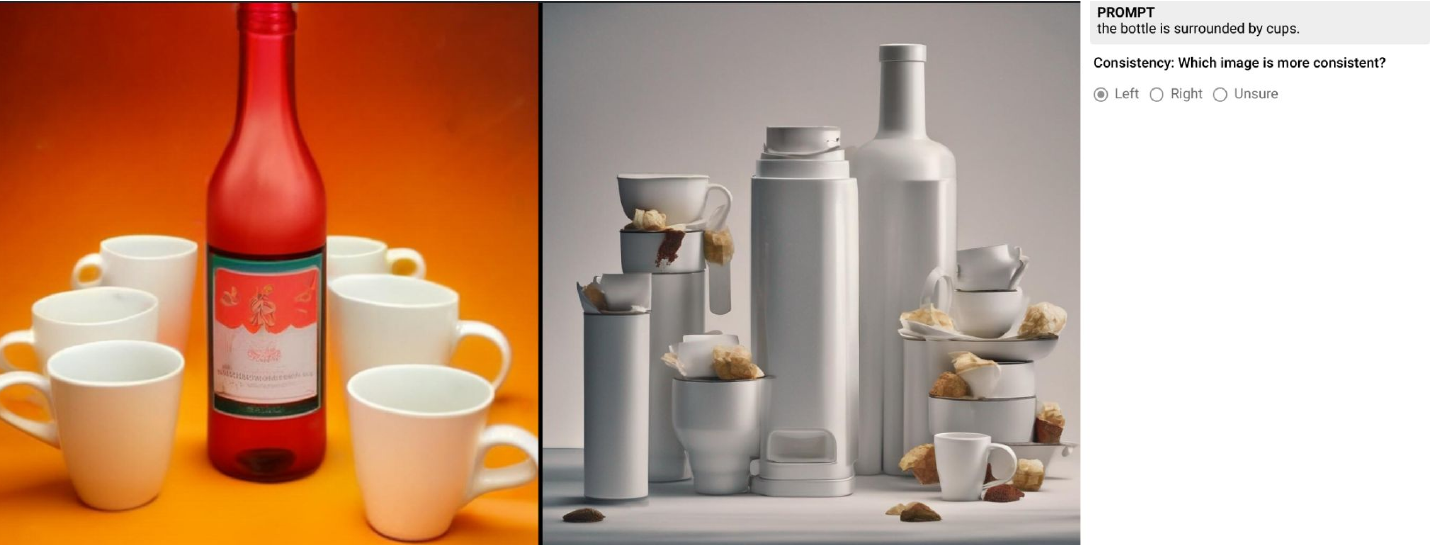}
    \caption{{\bf Side-by-side comparison annotation template user interface.} Example depicting the interface shown to the annotators when performing evaluation tasks with the side-by-side template. Raters are given a pair of images from different models, the prompt used to generate them and asked to pick which one is more consistent with the prompt. An \textit{Unsure} option is also given.}
    \label{fig:human_eval_screenshot_sxs}
\end{figure}

\subsubsection{Data Collection Details.}
\label{app:sec:humanevaldatacollectiondetails}
We recruited participants (N = 40) through a crowd-sourcing pool. The full details of our study design, including compensation rates, were reviewed by our institution's independent ethical review committee. All participants provided informed consent prior to completing tasks and were reimbursed for their time. Considering all four templates, both Gecko subsets, and the four evaluated generative models, approximately 108K answers were collected, totalling 2675 hours of evaluation.

\subsubsection{Percentage of unsure ratings for each annotation template/model}
One of the innovations of our human evaluation setup is to allow for annotators to reflect uncertainty in their ratings. In \Table{tab:app_human_eval_n_unsure} we show the percentage of \textit{Unsure} ratings for each absolute comparison annotation template. Overall, we find that evaluations with Gecko(R) yield a higher percentage of \textit{Unsure} ratings in comparison to Gecko(S). 
\begin{table}[h]
\begin{tabular}{c|cc|cc|cc}
\hline
\multirow{2}{*}{Gen. model} & \multicolumn{2}{c|}{\textbf{WL}} & \multicolumn{2}{c|}{\textbf{Likert}} & \multicolumn{2}{c}{\textbf{DSG(H)}} \\ \cline{2-7} 
                            & Gecko(R)        & Gecko(S)       & Gecko(R)          & Gecko(S)         & Gecko(R)         & Gecko(S)         \\ \hline
Imagen                      & 43.52           & 18.46          & 20.25             & 2.07             & 30.90            & 29.55            \\
Muse                        & 41.09           & 23.91          & 18.10             & 2.42             & 33.47            & 32.65            \\
SDXL                        & 20.09           & 20.94          & 4.24              & 2.05             & 31.77            & 29.64            \\
SD1.5                       & 13.35           & 26.48          & 10.04             & 4.09             & 37.02            & 33.08            \\ \hline
\end{tabular}
\caption{{\bf Percentage of \textit{Unsure} ratings.} Overall, evaluation with Gecko(S) yields fewer \textit{Unsure} ratings across all models and templates.}
\label{tab:app_human_eval_n_unsure}
\end{table}

\subsection{Additional experimental results} \label{app:human_eval_add_results}
\subsubsection{Pairwise model comparisons with reliable prompts}
\input{tables/human_eval_model_comparison_full}

\subsubsection{Correlation across templates and models.} We show the correlation between templates and models for both Gecko(R) and Gecko(S) in \Table{tab:h_eval_corr}. 

\begin{table}[h]
\resizebox{\textwidth}{!}{%
\begin{tabular}{c|cccccc|cccccc}
\hline
\multirow{3}{*}{Gen. models}    & \multicolumn{6}{c|}{\textbf{Gecko(R)}}                                                                                          & \multicolumn{6}{c}{\textbf{Gecko(S)}}                                                                                          \\ \cline{2-13} 
                                & \multicolumn{2}{c|}{Likert vs WL}     & \multicolumn{2}{c|}{Likert vs DSG(H)}   & \multicolumn{2}{c|}{WL vs DSG(H)} & \multicolumn{2}{c|}{Likert vs DSG(H)}   & \multicolumn{2}{c|}{Likert vs WL}     & \multicolumn{2}{c}{WL vs DSG(H)} \\ \cline{2-13} 
                                & Pearson & \multicolumn{1}{c|}{Spearman} & Pearson & \multicolumn{1}{c|}{Spearman} & Pearson          & Spearman         & Pearson & \multicolumn{1}{c|}{Spearman} & Pearson & \multicolumn{1}{c|}{Spearman} & Pearson         & Spearman         \\ \hline
\multicolumn{1}{c|}{SD1.5}     & 0.56   & \multicolumn{1}{c|}{0.64}    & 0.56   & \multicolumn{1}{c|}{0.60}    & 0.57            & 0.65            & 0.60   & \multicolumn{1}{c|}{0.61}    & 0.60   & \multicolumn{1}{c|}{0.62}    & 0.74           & 0.76            \\
\multicolumn{1}{c|}{SDXL}      & 0.60   & \multicolumn{1}{c|}{0.63}    & 0.52   & \multicolumn{1}{c|}{0.57}    & 0.56            & 0.61            & 0.60   & \multicolumn{1}{c|}{0.50}    & 0.56   & \multicolumn{1}{c|}{0.62}    & 0.78           & 0.79            \\
\multicolumn{1}{c|}{Muse}   & 0.67   & \multicolumn{1}{c|}{0.62}    & 0.61   & \multicolumn{1}{c|}{0.63}    & 0.61            & 0.66            & 0.51   & \multicolumn{1}{c|}{0.52}    & 0.51   & \multicolumn{1}{c|}{0.53}    & 0.77           & 0.75            \\
\multicolumn{1}{c|}{Imagen} & 0.65   & \multicolumn{1}{c|}{0.67}    & 0.59   & \multicolumn{1}{c|}{0.62}    & 0.68            & 0.71            & 0.63   & \multicolumn{1}{c|}{0.57}    & 0.59   & \multicolumn{1}{c|}{0.62}    & 0.81           & 0.80            \\ \hline
\end{tabular}}
\caption{{\bf Correlation between all absolute comparison templates.} We compute Pearson and Spearman correlation coefficients for all pairs of templates for both Gecko(R) and Gecko(S). We find significant results with $p<0.001$ for all cases and that scores of all metrics are at least moderately correlated, with the finer-grained templates, WL and DSG(H), being more correlated with each other in comparison to Likert.}
\label{tab:h_eval_corr}
\end{table}

\subsubsection{Distribution of scores per prompt-image pairs across annotation templates.} 
\begin{figure}[h]
  \centering
  
    \begin{subfigure}[t]{0.325\linewidth}
        \includegraphics[width=\linewidth]{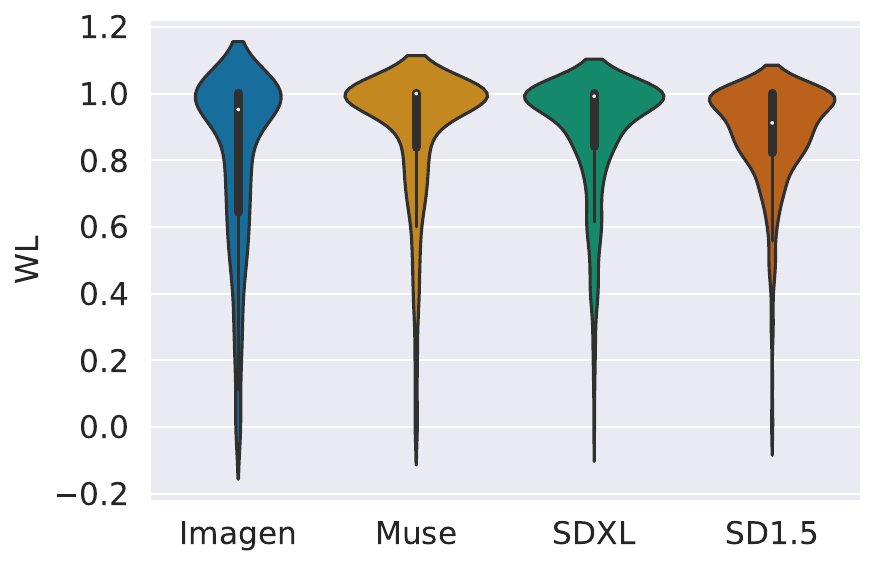}
        \caption{WL}
    \end{subfigure}%
    \begin{subfigure}[t]{0.325\linewidth}
        \includegraphics[width=\linewidth]{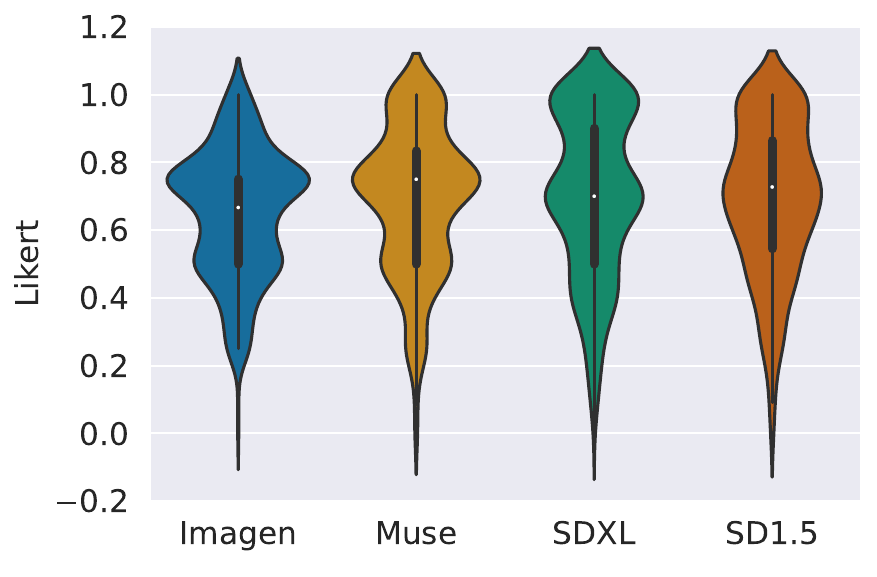}
        \caption{Likert}
    \end{subfigure}%
    \begin{subfigure}[t]{0.325\linewidth}
        \includegraphics[width=\linewidth]{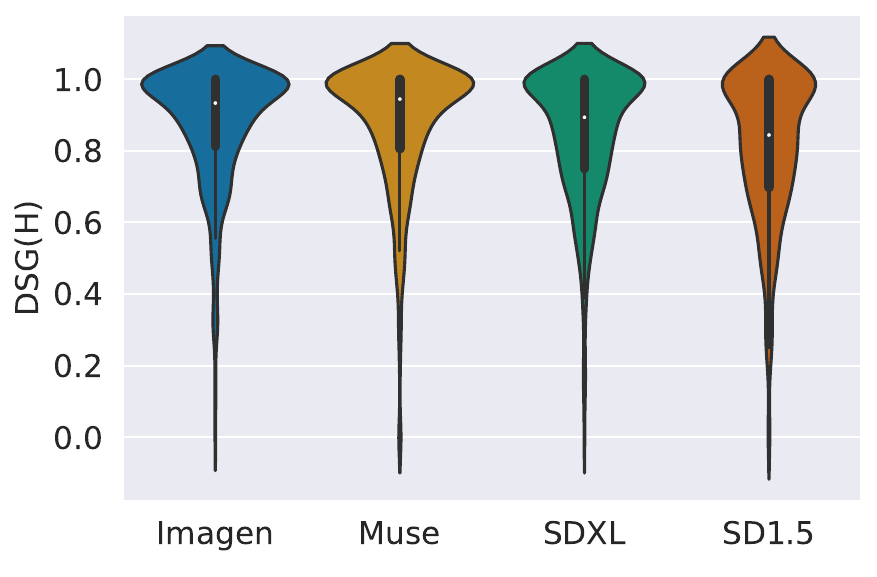}
        \caption{DSG(H)}
    \end{subfigure}%
  \caption{\textbf{Distribution of scores for Gecko(R).} We show violin plots for scores obtained with all absolute comparison templates. }
  \label{fig:human_eval_dsg1k_wo_unsure}
\end{figure}

\begin{figure}[h]
  \centering

    \begin{subfigure}[t]{0.325\linewidth}
        \includegraphics[width=\linewidth]{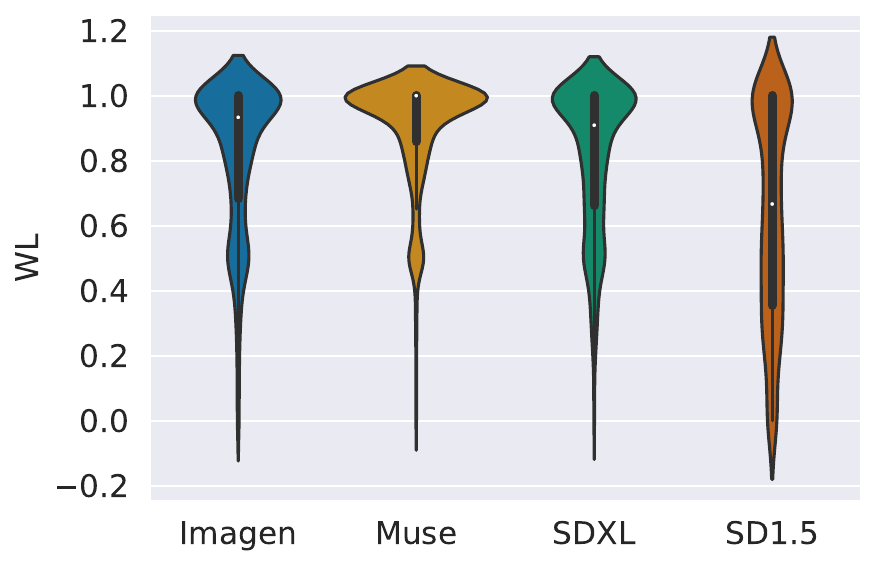}
        \caption{WL}
    \end{subfigure}%
    \begin{subfigure}[t]{0.325\linewidth}
        \includegraphics[width=\linewidth]{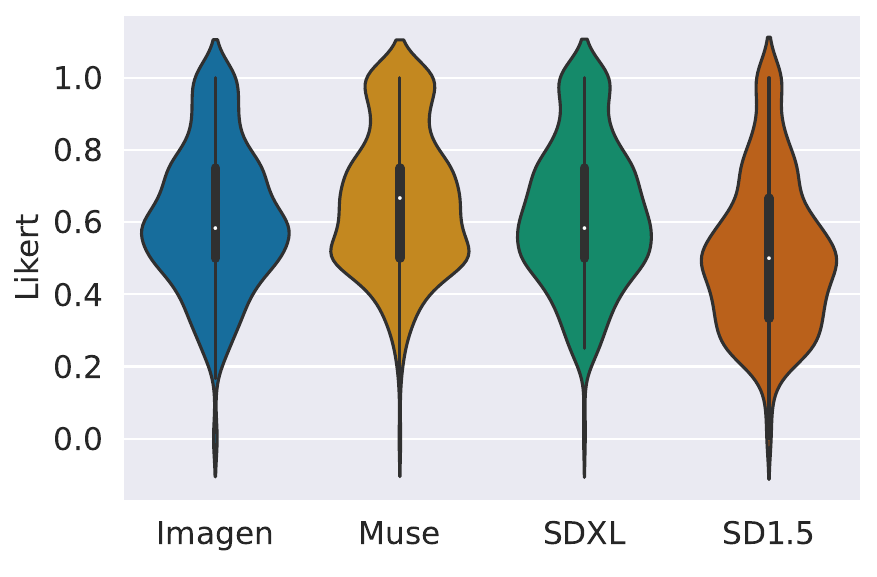}
        \caption{Likert}
    \end{subfigure}%
    \begin{subfigure}[t]{0.325\linewidth}
        \includegraphics[width=\linewidth]{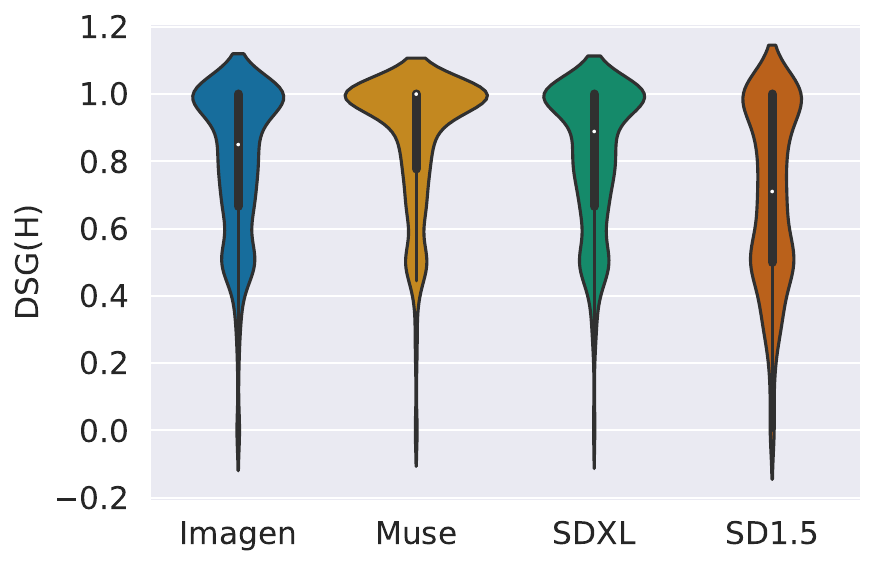}
        \caption{DSG(H)}
    \end{subfigure}%
  \caption{\textbf{Distribution of scores for Gecko(S).} We show violin plots for scores obtained with all absolute comparison templates.}
  \label{fig:human_eval_synth_wo_unsure}
\end{figure}

We plot the distribution of scores per each evaluated prompt-image pair for all the absolute comparison templates. The violin plots in \Figure{fig:human_eval_dsg1k_wo_unsure}-\ref{fig:human_eval_synth_wo_unsure} show the distributions for Gecko(R) and Gecko(S), respectively. It is possible to notice that scores obtained for Muse with WL and DSG(H) are more concentrated in values closer to 1 for both templates, corroborating findings from \Section{sec:human_eval_comparing_models} where results showed Muse was the overall best model.

\subsubsection{Side-by-side template.} In Table \ref{tab:sxs_krippendorffs} we show inter-annotator agreement results for the side-by-side annotation template for Gecko(R) and Gecko(S), along with the respective difference in agreement when using only the reliable prompts for both Gecko2K subsets. In Table \ref{tab:sxs_accuracy} we present the results of the comparison between the side-by-side template and the absolute comparison ones. 

\begin{table}[h]
\centering
\begin{tabular}{c|ccc|ccc}
\hline
                   & Gecko(R) & Gecko(R)-rel & $\Delta$  & Gecko(S) & Gecko(S)-rel & $\Delta$ \\ \hline
Imagen vs Muse & 0.438    & 0.465        & 0.027  & 0.485    & 0.625        & 0.140  \\
Imagen vs SDXL    & 0.440    & 0.425        & -0.015 & 0.521    & 0.608        & 0.087 \\
Imagen vs SD1.5   & 0.402    & 0.431        & 0.029  & 0.581    & 0.652        & 0.071 \\
Muse vs SDXL      & 0.471    & 0.489        & 0.018  & 0.570    & 0.638        & 0.068 \\
Muse vs SD1.5     & 0.539    & 0.592        & 0.053  & 0.600    & 0.617        & 0.017 \\
SDXL vs SD1.5    & 0.389    & 0.438        & 0.049  & 0.522    & 0.562        & 0.040  \\ \hline
\end{tabular}
\caption{{\bf Side-by-side template: inter-annotator agreement.} We compute Krippendorff's $\alpha$ for Gecko(R) and Gecko(S) and the difference ($\Delta$) in $\alpha$ when using only reliable prompts for both subsets of Gecko2K. In both cases, using the reliable subsets increases the overall inter-annotator agreement.}
\label{tab:sxs_krippendorffs}
\end{table}


\begin{table}[h]
\centering
\begin{tabular}{c|ccc}
\hline
                   & WL     & Likert          & DSG(H) \\ \hline
Imagen vs Muse & 0.736  & \textbf{0.755}  & 0.728      \\
Imagen vs SDXL    & 0.690  & 0.705           & \textbf{0.732}  \\
Imagen vs SD1.5   & 0.672  & 0.632           & \textbf{0.708}  \\
Muse vs SDXL      & 0.704  & \textbf{0.746}  & 0.703           \\ 
Muse vs SD1.5     & 0.759  & 0.740           & \textbf{0.749}  \\
SDXL vs SD1.5    & 0.693  & \textbf{0.700}           & 0.689   \\ \hline
Average          & 0.709  & 0.713            & \textbf{0.718} \\ \hline
\end{tabular}
\caption{{\bf Comparing side-by-side and absolute templates on Gecko2K-rel.} We compare the side-by-side template with the absolute comparison ones by computing the accuracy obtained by WL, Likert, and DSG(H) scores when using them to compare pairs of images on Gecko2K-rel. In this case, the ground-truth is assumed to be the results obtained with the side-by-side template.}
\label{tab:sxs_accuracy}
\end{table}


\subsection{Reliable prompts: Examples of image-prompt pairs with high human (dis)agreement}\label{app:human_eval_rel_prompts_examples}
In this section we show a representative list of prompts and corresponding images where human annotators were most likely to either agree or disagree in their ratings.
The annotators agreed in ratings if they gave similar scores across for an image-text pair, meaning that the resulting mean variance was zero or close to zero.
We refer to such prompts as ``high agreement'' prompts.
In contrast, if annotators gave different ratings for a text-image pair, this would result in higher mean variance and we call such prompts ``high disagreement'' prompts.

\begin{figure}[h]
  \centering
    \includegraphics[width=0.8\textwidth]{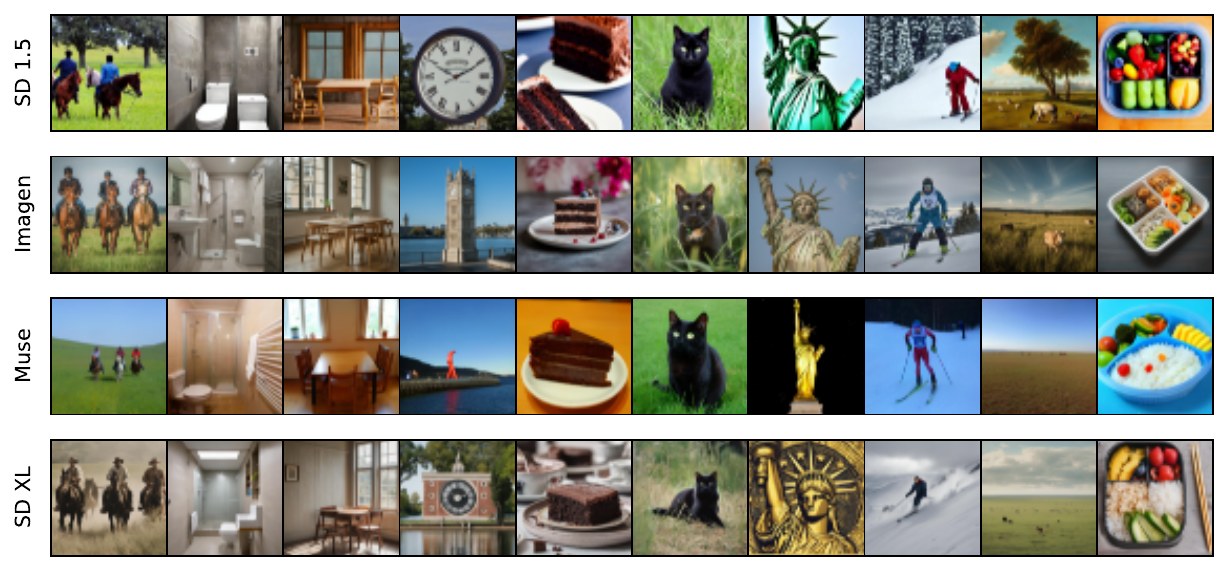}
  \caption{Images generated from Gecko(R) prompts with high level of agreement among raters. Prompts are listed in  Table~\ref{tab:prompts_agreement_r}.}
  \label{fig:prompts_agreement_r}
\end{figure}

\begin{figure}[h]
  \centering
  \includegraphics[width=0.8\textwidth]{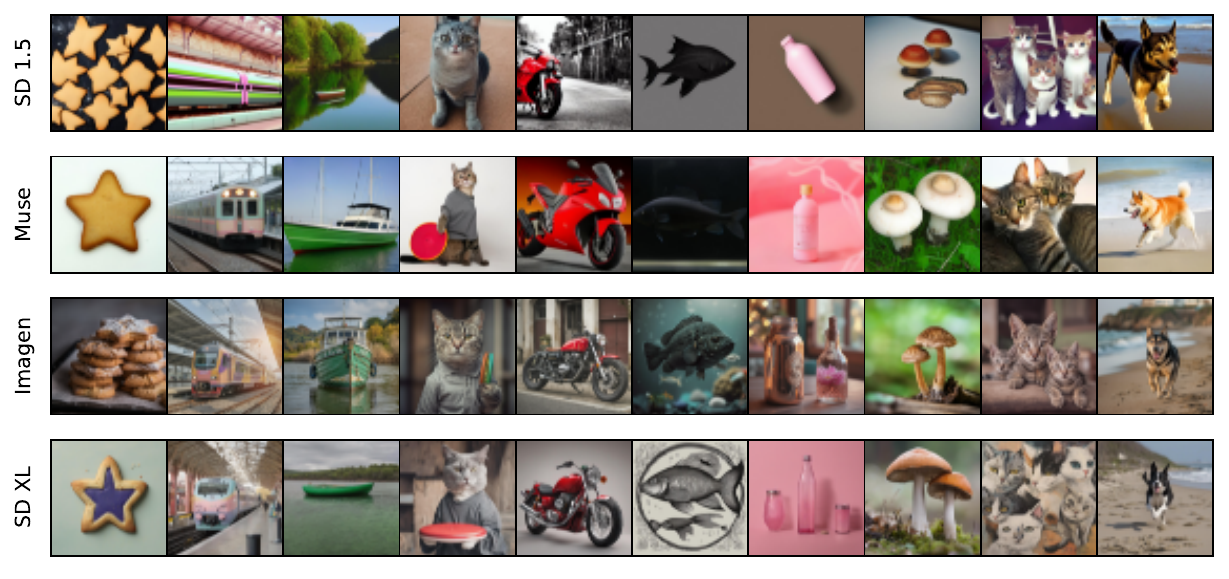}
  \caption{Images generated from Gecko(S) prompts with high level of agreement among raters. Prompts are listed in  Table~\ref{tab:prompts_agreement_s}.}
  \label{fig:prompts_agreement_s}
\end{figure}

\begin{figure}[h]
  \centering
  \includegraphics[width=0.8\textwidth]{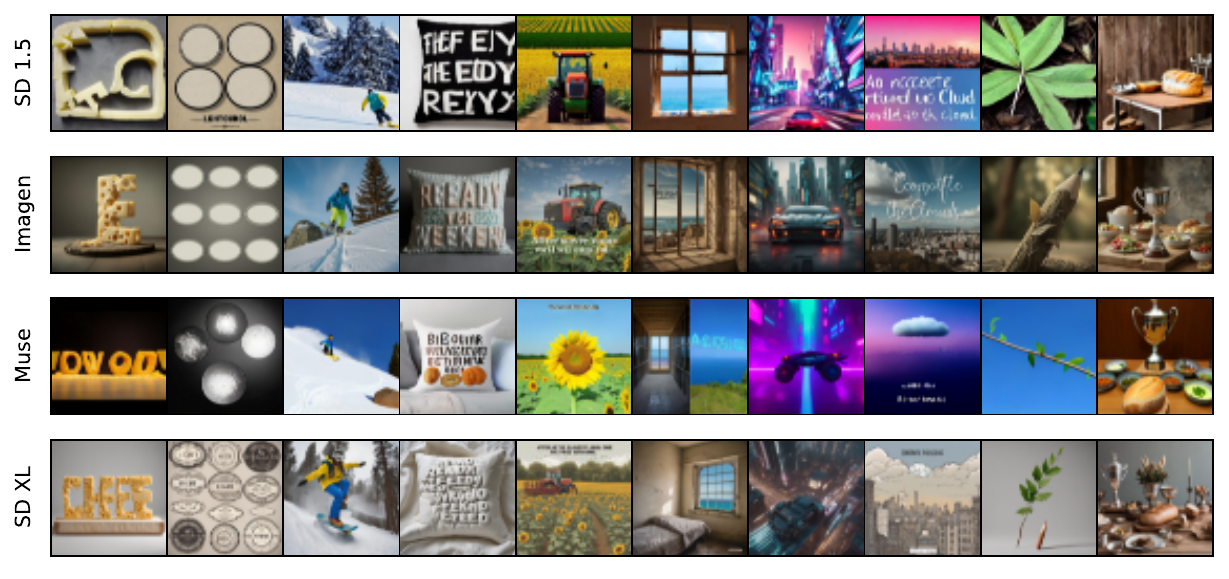}
  \caption{Images generated from Gecko(R) prompts with high level of disagreement among raters. Prompts are listed in  Table~\ref{tab:prompts_disagreement_r}.}
  \label{fig:prompts_disagreement_r}
\end{figure}

\begin{figure}[h]
  \centering
  \includegraphics[width=0.8\textwidth]{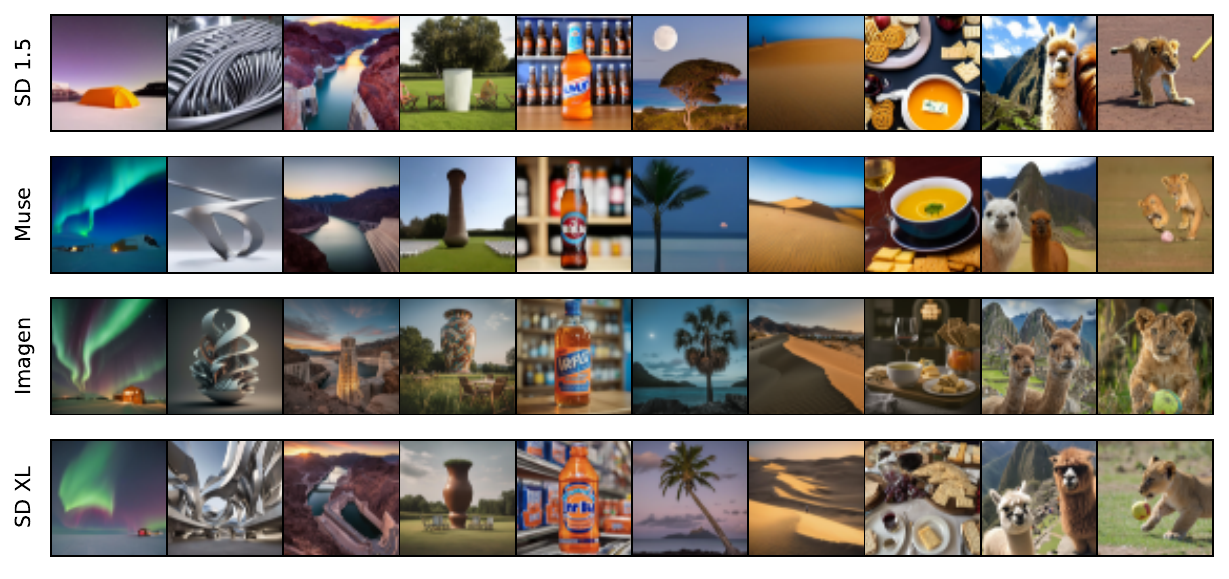}
  \caption{Images generated from Gecko(S) prompts with high levels of disagreement among raters. Prompts are listed in Table~\ref{tab:prompts_disagreement_s}.}
  \label{fig:prompts_disagreement_s}
\end{figure}

To find prompts with high agreement across raters for all templates and all models, for each model-template combination we pick a subset of responses with low variance.
Low variance is defined as the mean variance of a prompt-image pair for a model-template pair being below a certain threshold.
The threshold is set as 10\% of the maximum variance for that model-template set of ratings for both Gecko(R) and Gecko(S).
Analogously, we also find a set of prompts with high disagreement; for this we find prompts that have mean variance above 1\% of the maximum variance for a given template and for prompts from Gecko(R) and Gecko(S).
The specific threshold value here is relevant only insofar as it captures at least 10 prompt-image pairs which we are interested in visualising.
Then, we find prompts with high agreement by intersecting all model-template prompt sets where prompts have been selected based on the threshold. 
The procedure is analogous for low agreement prompts.
For Gecko(R), both sets, namely the set of prompts with high agreement as well as the set of prompts with high disagreement have 34 prompts each.
For Gecko(S), the set of prompts with high agreement has 62 prompts, while the set of prompts with high disagreement contains 85 prompts.
A subset of 10 prompts for all different combinations is listed in 
Tables~\ref{tab:prompts_agreement_r}-\ref{tab:prompts_disagreement_s} and corresponding images are shown in the Figure~\ref{fig:prompts_agreement_r}-\ref{fig:prompts_disagreement_s}.

Based on the analyses of such subsets, we observe several interesting trends.
First, for Gecko(R) the prompts with higher agreement tend to be significantly shorter in length $(\mu=54.32,\sigma=32.18)$ as measured by the number of characters, compared to the length of prompts with high disagreement $(\mu=173.35,\sigma=86.35,$ Welch's t-test $t(41.99)=-7.42$ (p$<$0.001). The same observation holds for Gecko(S), where high agreement prompts were also significantly shorter $(\mu=20.77, \sigma=18.03)$, than high disagreement prompts $(\mu=82.48, \sigma=95.17,$ Welch's t-test $t(92.20)=-5.80$ (p$<$0.001).
We further observe that prompts where raters tend to agree more are highly specific (\ie they refer to one or just a few objects with few attributes), whereas prompts with high disagreement tend to describe more complex scenes with visual descriptors and often mentioning named entities or text rendering. Intuitively, this makes sense as longer prompts are more likely to require several skills.

\begin{table}
\centering
\begin{tabular}{ll}
\toprule
 & \textbf{High Agreement Prompts (Gecko(R))} \\
\midrule
1 & three men riding horses through a grassy field \\
2 & a small bathroom with a shower and a toilet \\
3 & a wooden table with four wooden chairs in front of two windows \\
4 & a large colgate clock is by the water \\
5 & a slice of chocolate cake is on a small plate \\
6 & a black cat sitting in a field of grass \\
7 & The Statue of Liberty made of gold \\
8 & a man riding skis down a snow covered slope \\
9 & a vast, grassy field with animals in the distance \\
10 & a plastic bento box filled with rice, vegetables and fresh fruit \\
\bottomrule
\end{tabular}
\caption{Selected prompts with a high level of agreement in scores among raters for Gecko(R).}
\label{tab:prompts_agreement_r}
\end{table}

\begin{table}[h]
\centering
\begin{tabularx}{.9\textwidth}{Xp{.85\textwidth}}
\toprule
 & \textbf{High Agreement Prompts (Gecko(S))} \\
\midrule
1 & A star-shaped cookie \\
2 & a pastel coloured train passing through the station. \\
3 & a green boat. \\
4 & the cat wears a gray shirt and holds a frisbee \\
5 & a red motorcycle. \\
6 & a black fish. \\
7 & a pink bottle. \\
8 & two mushrooms. \\
9 & five cats. \\
10 & a dog named Balto is running on a beach. \\
\bottomrule
\end{tabularx}
\caption{Selected prompts with a high level of agreement in scores among raters for Gecko(S).}
\label{tab:prompts_agreement_s}
\end{table}
\begin{table}[h]
\centering
\begin{tabularx}{.9\textwidth}{Xp{.85\textwidth}}
\toprule
 & \textbf{High Disagreement Prompts (Gecko(R))} \\
\midrule
1 & Studio shot of sculpture of text 'cheese' made from cheese, with cheese frame. \\
2 & vintage light monochrome six round and oval label set  Illustration \\
3 & There is a person snow boarding down a hill. There are tracks in the snow all around the snowboarder. There is a large rock in the snow next to them. There is a green pine tree in front of the snowboarder. The snowboarder is wearing blue ski pants and a blue and yellow jacket. They have a yellow snowboard on their feet. \\
4 & pillow in the shape of words 'ready for the weekend', letterism, funny jumbled letters, [ closeup ]!!, breads, author unknown, flat art, swedish, diaper-shaped, 2000, white clay, surreal object photography \\
5 & a sunflower field with a tractor about to run over a sunflower, with the caption 'after the sunflowers they will come for you' \\
6 & a photo of a prison cell with a window and a view of the ocean, and the word 'freedom' painted on the glass \\
7 & vehicle flying through a cyberpunk city 4 k, hyper detailed photograph \\
8 & a scene with a city in the background, and a single cloud in the foreground, with the text 'contemplate the clouds' in rounded cursive \\
9 & A pencil made of a tree branch with leaves \\
10 & A wooden table that has a silver trophy in the middle of it. In front of the trophy are several bowls and dishes containing food. There is a loaf of bread on a block of wood at the front of the table. \\
\bottomrule
\end{tabularx}
\caption{Selected prompts with a high level of disagreement in scores among raters for Gecko(R).}
\label{tab:prompts_disagreement_r}
\end{table}
\begin{table}[t]
\centering
\begin{tabularx}{.9\textwidth}{Xp{.85\textwidth}}
\toprule
 & \textbf{High Disagreement Prompts (Gecko(S))} \\
\midrule
1 & the amazing view from the Halley Research Station in Antarctica on a clear night, the full Moon is rising and the sky is ablaze with the aurora australis, or polar lights. \\
2 & a futuristic sculpture made of smooth metal \\
3 & time lapse of sunrise over at the Hoover Dam \\
4 & A huge vase in the middle of a field towering over the lawn chairs. \\
5 & a bottle of Irn-Bru is sitting on a shelf. \\
6 & a lord howe island palm tree with a moon rising in the distance \\
7 & a long exposure image of the golden dunes at Playa del Ingles on the Canary Island, with a lone tourist \\
8 & The soup is behind the cheese platter, to the left of the wine glasses, and below the crackers. \\
9 & An alpaca and Chewbacca pose for a selfie at Machu Picchu \\
10 & the lion cub named Simba is catching a ball. \\
\bottomrule
\end{tabularx}
\caption{Selected prompts with a high level of disagreement in scores among raters for Gecko(S).}
\label{tab:prompts_disagreement_s}
\end{table}

\clearpage

\subsection{Human evaluation templates: Challenging cases}\label{app:human_eval_challenges_templates}
\begin{figure}[h]
\renewcommand{\arraystretch}{1.3}
\centering
\resizebox{0.65\textwidth}{!}{
\begin{tabular}{cc}
\toprule
Image             & Ratings                                                         \\ \bottomrule 
\multirow{3}{*}{\includegraphics[width=0.1\textwidth]{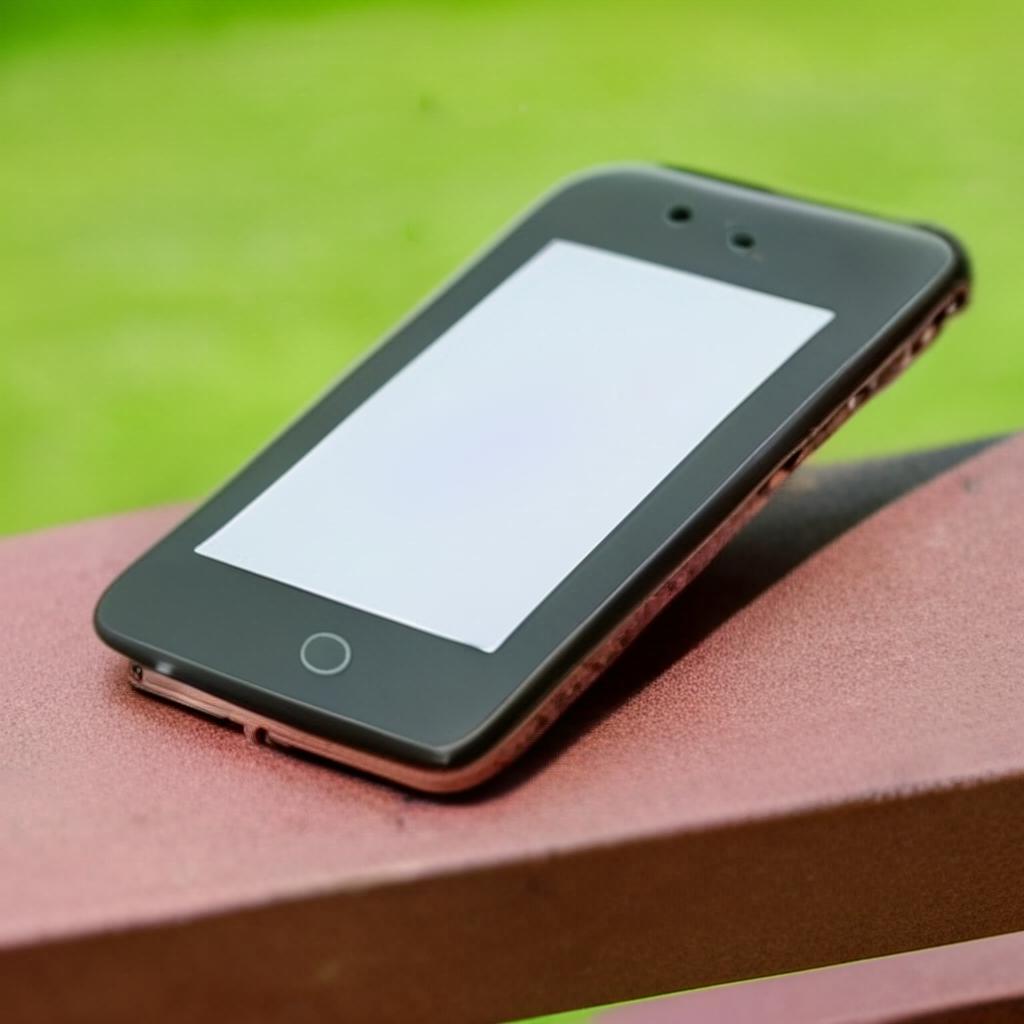}} & {\color{Green}A} {\color{Yellow}Nexus One} {\color{Green}is placed on a} {\color{Yellow}bench.}                                \\
                  & {\color{Red}A Nexus One} {\color{Green}is placed} {\color{Yellow}on a bench.}                                \\
                  & {\color{Green}A} {\color{Yellow}Nexus One} {\color{Green}is placed on a} {\color{Yellow}bench.} \\ \midrule
\multirow{3}{*}{\includegraphics[width=0.1\textwidth]{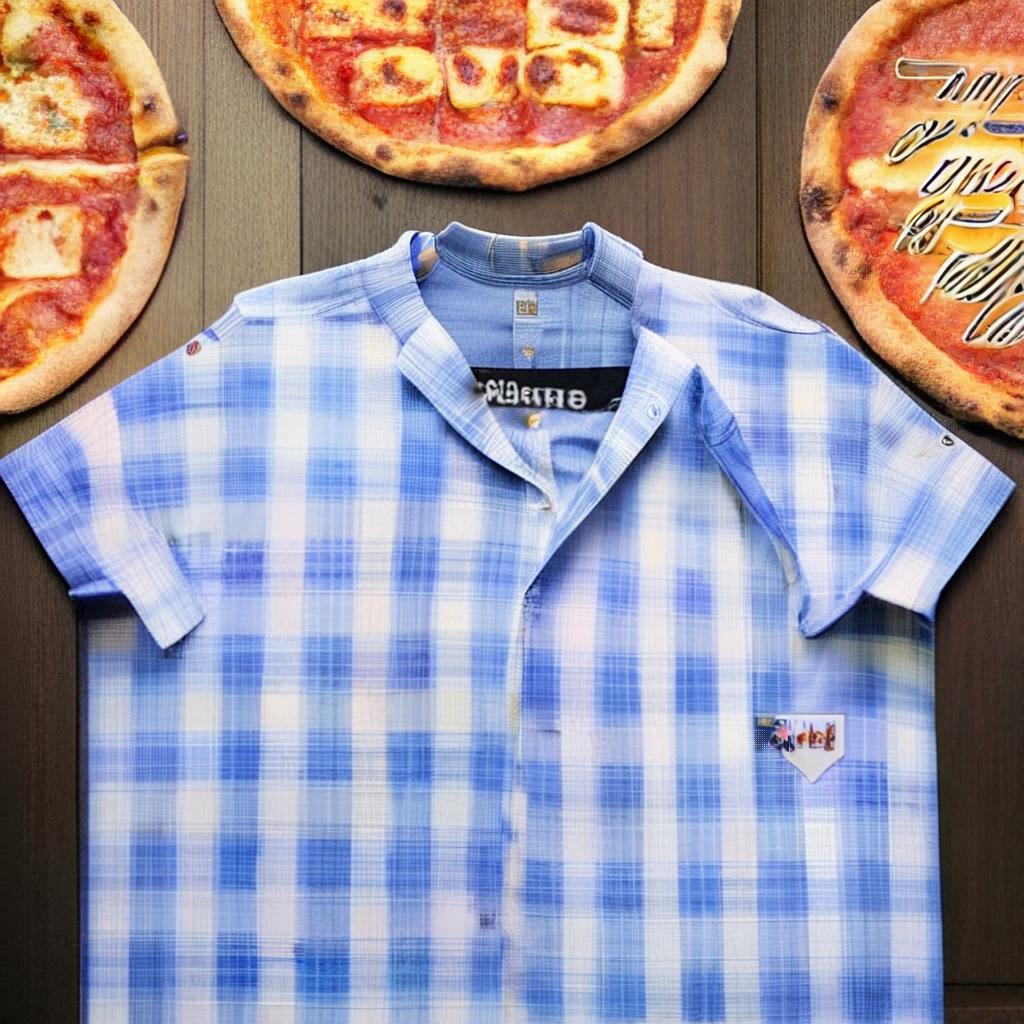}} & {\color{Red}Some shirts} {\color{Green}and some pizzas. There are} {\color{Red}more shirts than} {\color{Green}pizzas.} \\
                  & {\color{Red}Some} {\color{Green}shirts and some pizzas. There are} {\color{Red}more} {\color{Green}shirts than pizzas.} \\
                  & {\color{Green}Some} {\color{Red}shirts} {\color{Green}and some pizzas. There are} {\color{Red}more} {\color{Green}shirts than pizzas.} \\ \bottomrule
\end{tabular}}
\caption{{\bf Examples of challenges for WL.} We show two examples of evaluated images, respective prompts and annotations from three raters. Each word is coloured according to the score given by the rater: {\color{Green}{green}} indicates \textit{Aligned}, {\color{Red}{red}} \textit{Not aligned}, and {\color{Yellow}{yellow}} \textit{Unsure}. Both examples show that WL can be sensitive to words that are not relevant to the alignment evaluation. {\bf Top:} All raters seem to agree it is not possible to tell whether a bench is represented in the image (hence the word is evaluated as \textit{Unsure}). In spite of that, one of the raters disagrees on how to rate the ``on a'' preposition. {\bf Bottom:} All raters seem to agree the quantity of shirts in the image does not reflect the prompt, but their ratings vary in terms of which words are rated as not aligned.}
\label{fig:human_eval_wl_challenging}
\end{figure}

\begin{figure}[h]
\renewcommand{\arraystretch}{1.3}
\centering
\resizebox{0.8\textwidth}{!}{
\begin{tabular}{ccccc}
\toprule
Image & Prompt & Rater 1 & Rater 2 & Rater 3 \\
\bottomrule
\multirow{3}{*}{\includegraphics[width=0.1\textwidth]{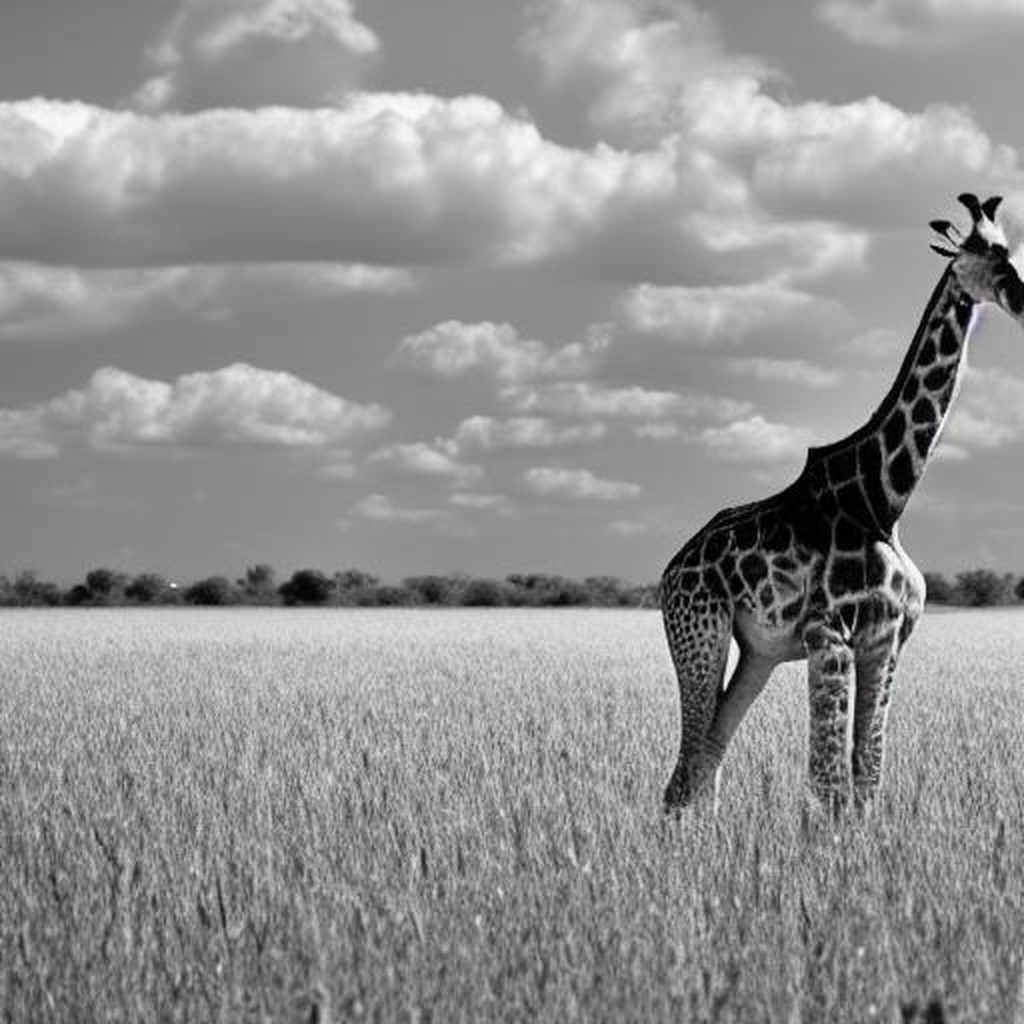}} &                                &                       &                       &                     \\
                 & A giraffe stands in the field. & 4-Mostly consistent   & 5-Consistent          & 5-Consistent        \\
                 &                                &                       &                       &                     \\
\multirow{3}{*}{\includegraphics[width=0.1\textwidth]{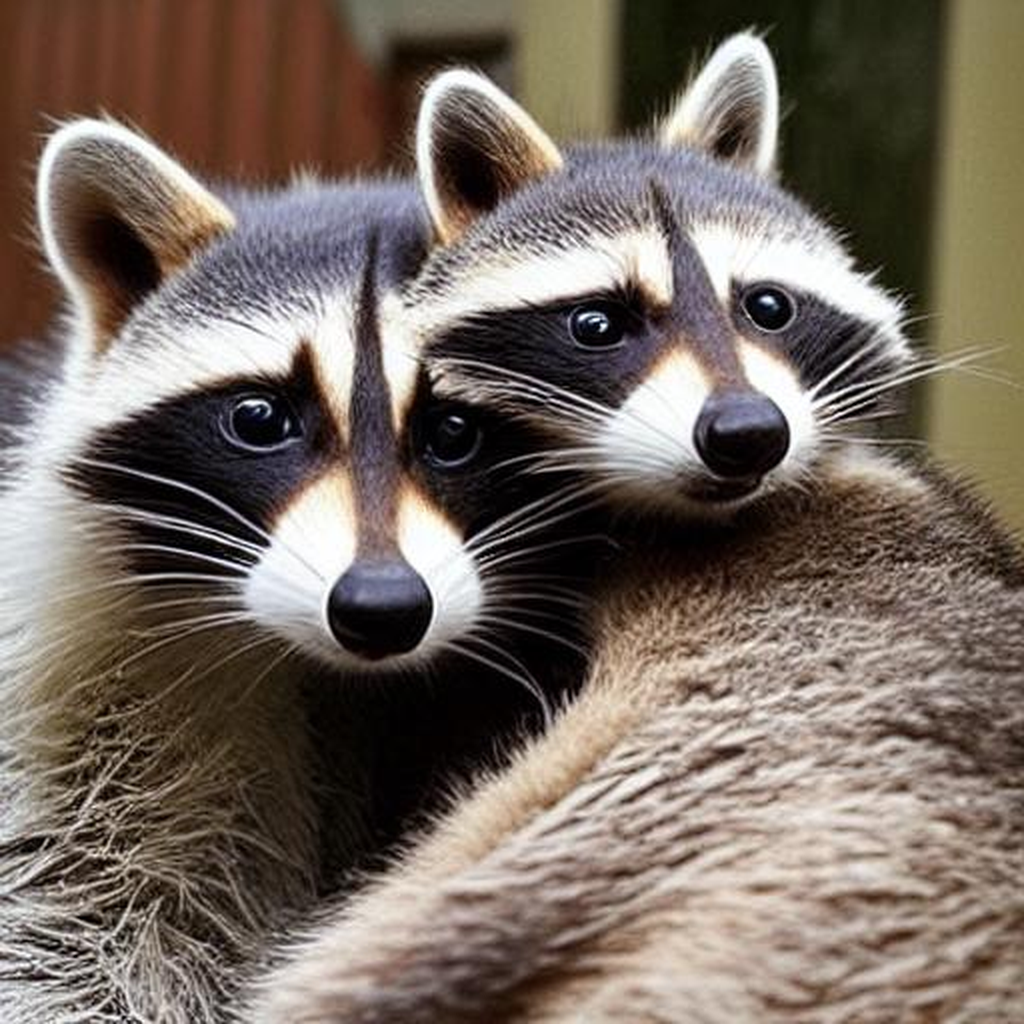}} &                                &                       &                       &                     \\
                 & The raccoon holds the cat.     & 2-Mostly inconsistent & 3-Somewhat consistent & 4-Mostly consistent \\ 
                 &                                &                       &                       & \\ \bottomrule
\end{tabular}}       
\caption{{\bf Examples of challenges for Likert.} {\bf Top:} Raters might take into account other aspects of the images besides alignment when evaluating a prompt-image pair. In this example, although the image is perfectly consistent with the prompt, one of the raters penalised its score. We hypothesise they took into account the fact the generated image is in grey scale. {\bf Bottom:} ``Uncalibrated'' scores across raters. The scores of all three raters reflect the imperfect consistency between prompt and image, but each rater penalised the score with different intensity.}
\label{fig:human_eval_likert_challenging}
\end{figure}

\begin{figure}[h]

\renewcommand{\arraystretch}{1.3}
\centering
\resizebox{0.7\textwidth}{!}{
\begin{tabular}{ccc}
\toprule
Image             & Prompt                      & Questions                         \\ \bottomrule
\multirow{3}{*}{\includegraphics[width=0.1\textwidth]{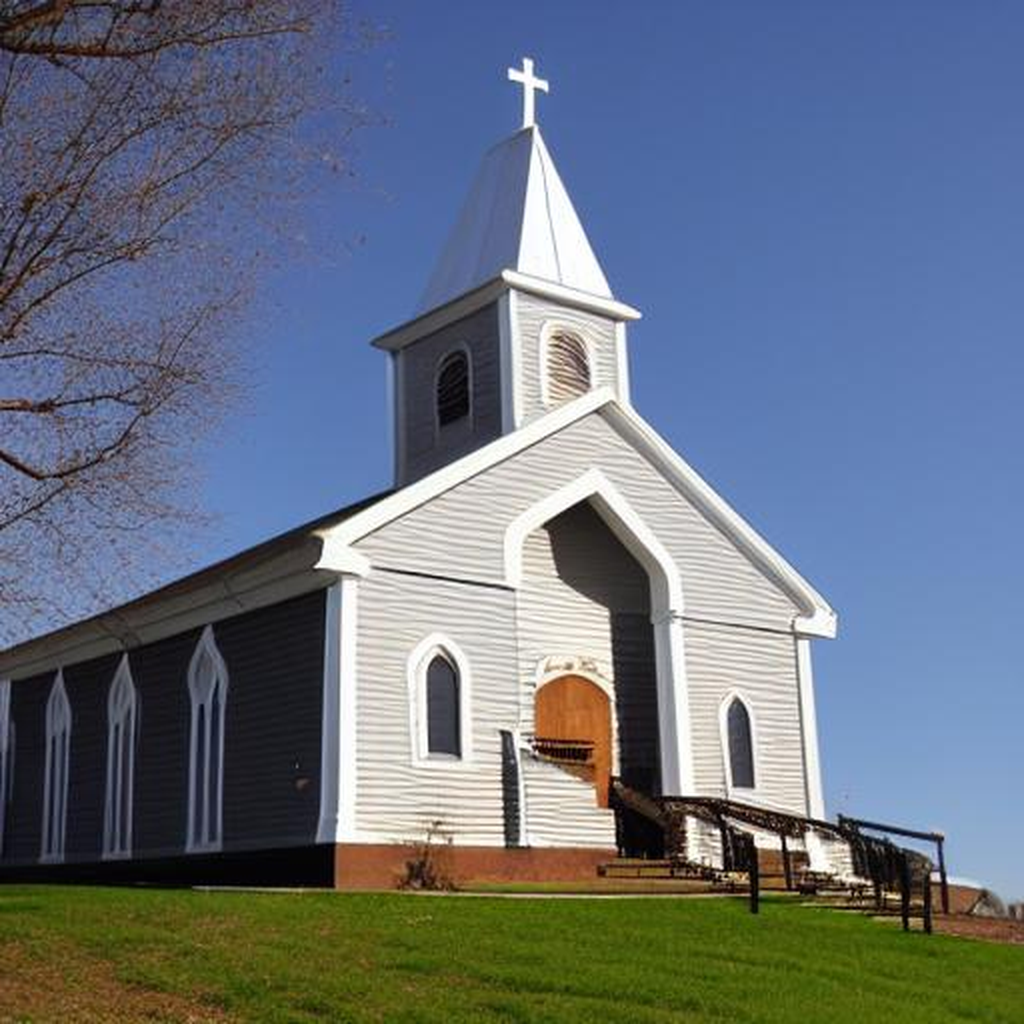}} &                             & Is there a church?                \\
                  & A church without a steeple. & Does the church have a steeple?   \\
                  &                             & Is the steeple missing?           \\ \midrule
\multirow{3}{*}{\includegraphics[width=0.1\textwidth]{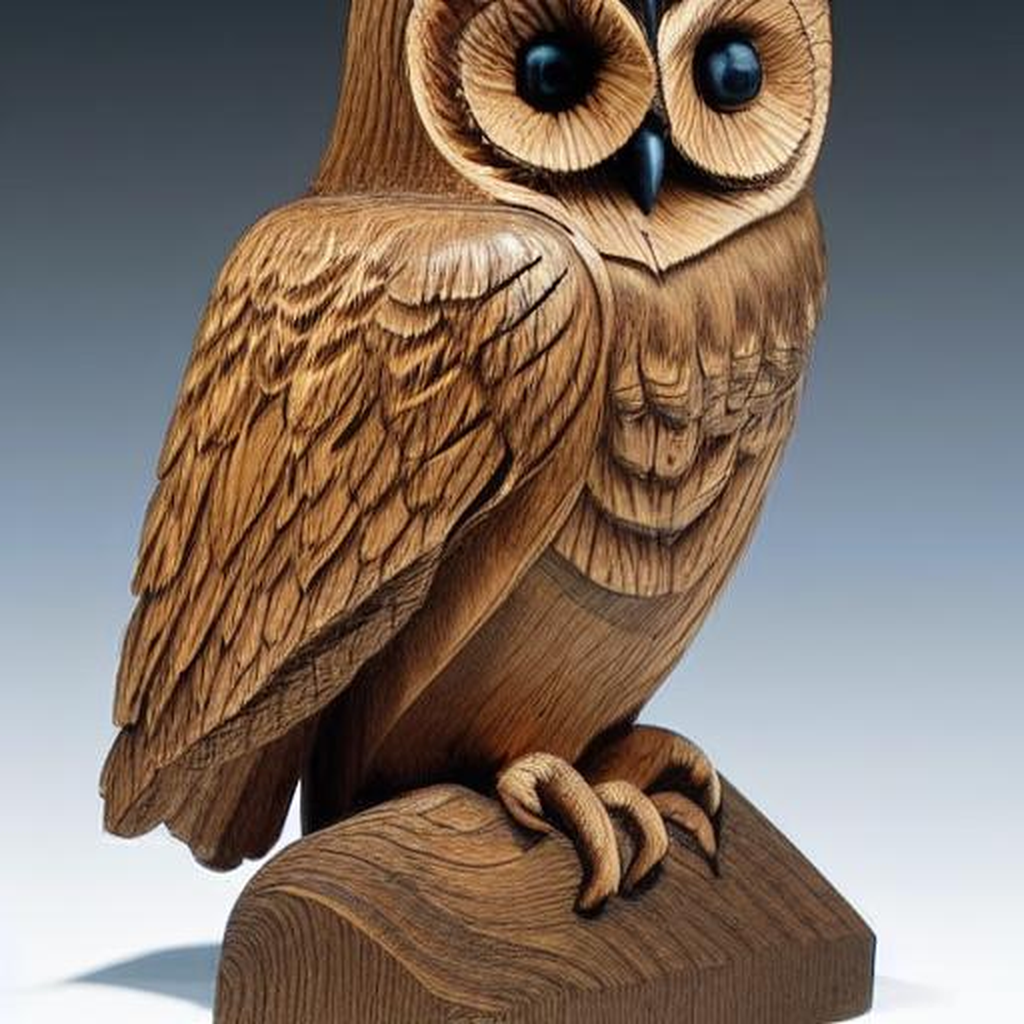}} &                             & Is there an owl?                  \\
                  & A wood carving of an owl.   & Is there a wood carving?          \\
                  &                             & Is the wood carving made of wood? \\ \bottomrule
\end{tabular}}
\caption{{\bf Examples of challenges for DSG(H).} {\bf Top:} Language complexity--Negation. As also shown in \Figure{fig:resultssynthetic}, the question generation is confused by the negation (asking if the church {\em has} a steeple as opposed to {\em does not have} a steeple). {\bf Bottom:} Coverage. The question generation fails to capture that the owl should be represented as a wood carving.}
\label{fig:human_eval_dsgh_challenging}
\end{figure}

In Figs. \ref{fig:human_eval_wl_challenging}, \ref{fig:human_eval_likert_challenging}, and \ref{fig:human_eval_dsgh_challenging} we show examples of challenging cases for the absolute comparison templates.


\clearpage

\section{T2I Models: Additional comparisons}

\subsection{Analysing model ratings per skill}
\label{app:t2imodelpercategory}
\begin{figure}
    \centering
    \includegraphics[width=\linewidth]{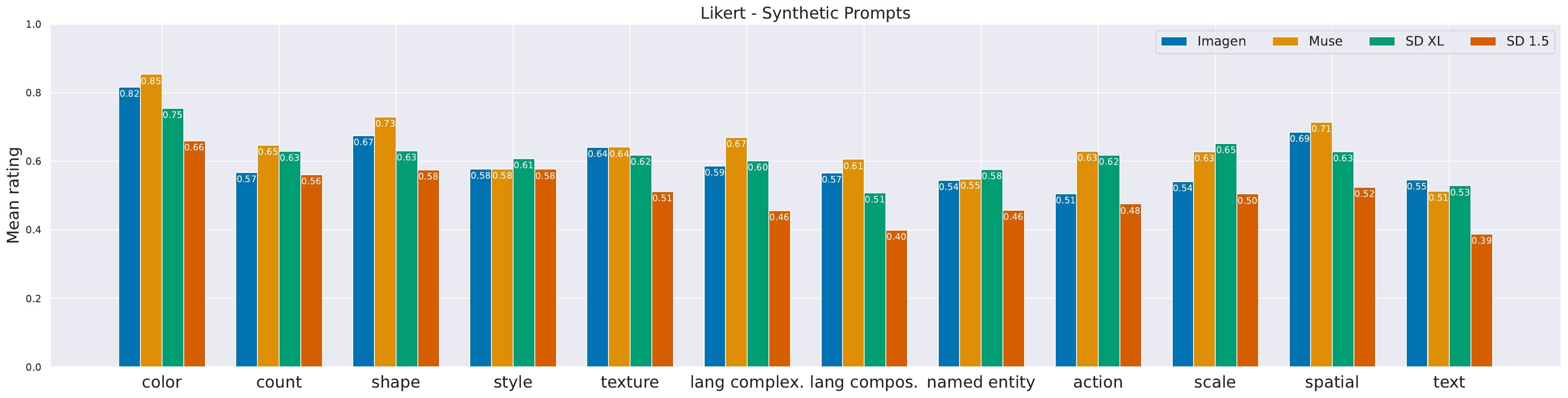}
    \caption{{\bf Per skill results - Likert.} Muse scores the best in nine out of the twelve categories, and SD1.5 performs the worst in all categories. Focusing on Muse, SDXL, and Imagen, the models score above 0.5 on all categories. Recalling that the Likert scale is symmetric (0.0 being inconsistent, and 1.0 being consistent), we see that these three models are more consistent than inconsistent on average (albeit only slightly for skills such as `lang compos.', `named entity' and `text').}
    \label{fig:results_by_category_mean_likert}
\end{figure}

\begin{figure}
    \centering
    \includegraphics[width=\linewidth]{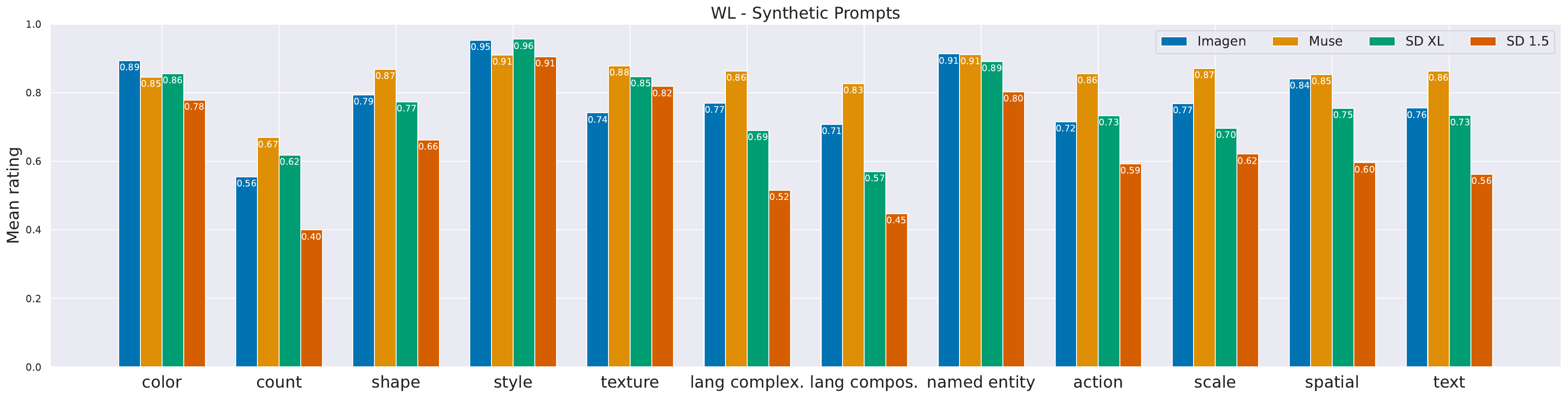}
    \caption{{\bf Per skill results - WL.} Muse scores the best in ten out of the twelve skills, and SD1.5 performs the worst in all skills. Moreover, Muse scores higher than the other models by a noticeable margin ($\geq 0.1)$ for the skills `lang compos.', `action', `scale' and `text'. In this case, analysing the results by skill shows that we can contribute Muse's higher average score (over the whole prompt set) mostly to these skills.}
    \label{fig:results_by_category_mean_wl}
\end{figure}

\begin{figure}
    \centering
    \includegraphics[width=\linewidth]{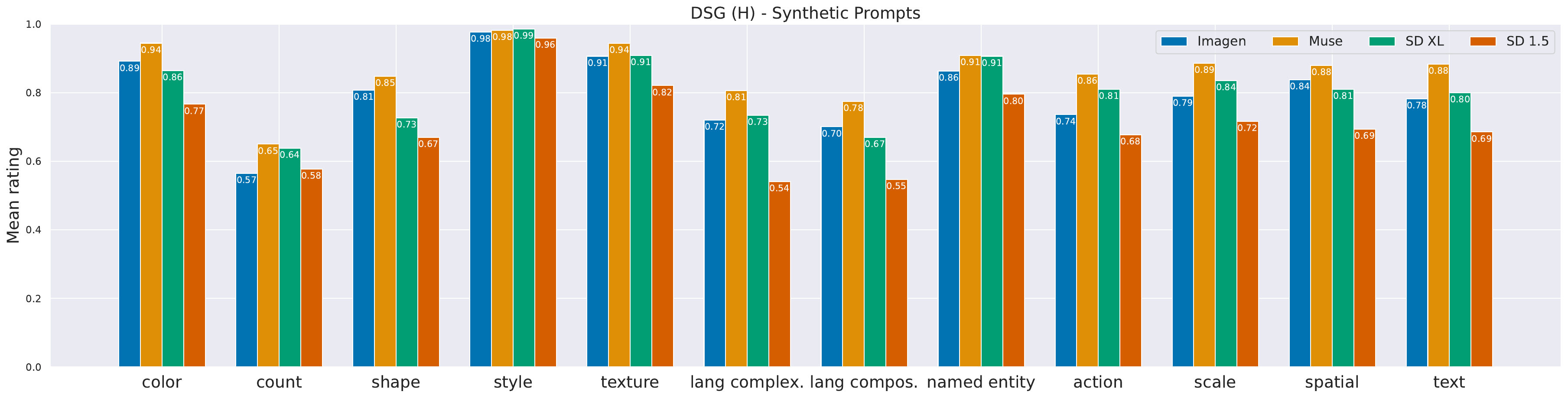}
    \caption{{\bf Per skill results - DSG(H).} Muse performs well across the skills, being the best eleven out of twelve times (scoring very close to the top for `style'). On the other hand, SD1.5 scores the worst in all the skills. This is consistent with the average scores on the overall prompt set. We see that counting is the most difficult skill for Muse, SDXL, and Imagen.  Aside from counting, the hardest skills for Muse are the language ones (`lang complex.' and `lang compos.'). This relative skill deficiency is not evident from the Likert and WL ratings, and therefore, the DSG ratings are better able to capture model shortcomings for prompts with more complex linguistic structure.}
    \label{fig:results_by_category_mean_dsg}
\end{figure}

In Figures~\ref{fig:results_by_category_mean_likert}, \ref{fig:results_by_category_mean_wl} and \ref{fig:results_by_category_mean_dsg}, we plot the mean ratings in different skills for Likert, WL and DSG(H), respectively. We focus on Gecko(S) because we have a skill/sub-skill label for each prompt. Our goal is to understand how the trends in model performance on the whole prompt set relate to their performance in individual skills. We provide an overview of the results in the captions for the plots. Overall, the results broken down by skill are consistent with the averages over the whole prompt set. In other words, if a model is better or worse on the full prompt set, this is generally true for the individual categories as well. Another observation is that \textsc{counting} and \textsc{complex language} seem to be the most difficult skills judging by WL and DSG, but this is not as clear from Likert (where many categories seem just as difficult).

\begin{figure}
    \centering
    \includegraphics[width=\linewidth]{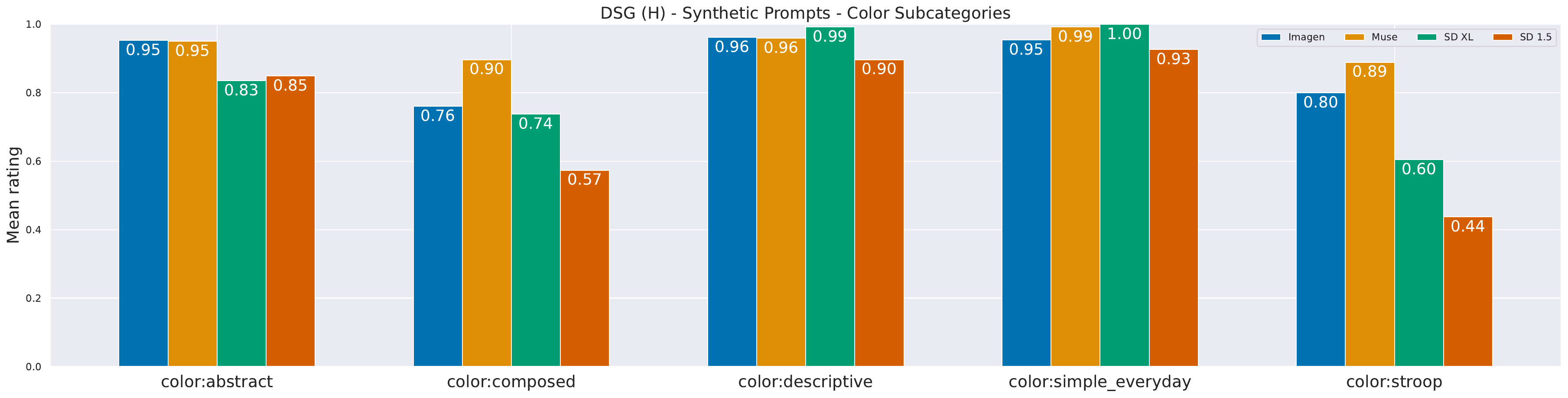}
    \caption{{\bf Color sub-skill results - DSG (H).} We further break down the prompts in the color skill into five sub-skills. One observations is that two of the sub-skills (`composed' and `stroop') are noticeably more difficult for all the models. This can be explained by the fact that they combine multiple skills: `composed' includes prompts with multiple colors/objects, and `stroop' includes text rendering in a certain color. Hence, while models may perform well on a skill overall, the sub-skills can illuminate where they struggle in generating images aligned with more complex prompts. On the other hand, models perform well with both abstract and everyday colors, likely because these are more commonly seen during training.}
    \label{fig:results_by_category_mean_dsg_color}
\end{figure}

\paragraph{\bf Further Breaking Down Skills.}
We can gain more insight into the skills of the models by looking at variation within a skill.  Figure~\ref{fig:results_by_category_mean_dsg_color} shows sub-skills of the \textsc{colour} prompts. We find that two sub-skills are more challenging, corresponding to prompts that require the models to combine multiple skills when generating the image (i.e., colour plus either composition or text rendering). For example, the `colour:composed' sub-skill (\textsc{composed expressions}) includes prompts such as \textit{`A brown vase, a white plate, and a red fork.'} with variations in the colors/objects. The sub-skill `color:stroop' (\textsc{stroop}) contains prompts like \textit{`Text saying "green" in white letters.'} where the word in quotes differs from the color of the letters.

\clearpage

\subsection{Model comparisons with TIFA160}
\label{app:tifa160_dataset_experiments}

We augment the experiment from \Figure{fig:h_eval_wilcoxon} in \Section{sec:human_eval_comparing_models} by performing a similar analysis with TIFA160. We generate images with this set of prompts and perform human evaluation following the same protocol as for Gecko2K and its subsets. In \Figure{fig:h_eval_wilcoxon_tifa}, we show the results of pairwise model comparisons with TIFA160 carried out with the Wilcoxon signed-rank test with $p<0.001$. Results show that all three versions of Gecko prompts are able to better distinguish models by finding more significant comparisons between them.

\begin{figure}[h]
\vspace{-0.5mm}
\tiny
\begin{tikzpicture}
  \matrix(D)[matrix of nodes,nodes in empty cells,
             row sep=-\pgflinewidth,column sep=-\pgflinewidth,
             nodes={anchor=center},
             row 1/.style={minimum height=0.5cm},
             row 2/.style={minimum height=0.5cm},
             row 3/.style={minimum height=0.5cm},
             row 4/.style={minimum height=0.5cm},
             row 5/.style={minimum height=0.5cm},
             row 6/.style={minimum height=0.2cm},
             row 7/.style={minimum height=0.5cm},
             column 1/.style={minimum width=0.5cm},
             column 2/.style={minimum width=0.5cm},
             column 3/.style={minimum width=0.5cm},
             column 4/.style={minimum width=0.5cm},
             column 5/.style={minimum width=0.5cm},
            ]
  {%
    &  &  & \\
    &  & \rotatebox{90}{WL} & \rotatebox{90}{L} & \rotatebox{90}{D(H)}\\
    & G(S)-rel & |[draw,fill=red!20]|$<$ & |[draw,fill=red!20]|$<$ & |[draw,fill=red!20]|$<$ \\
    & G(R)-rel & |[draw,fill=red!20]|$<$ & |[draw,fill=red!20]|$<$& |[draw,fill=blue!20]|=\\
    & G2K-rel & |[draw,fill=red!20]|$<$ & |[draw,fill=red!20]|$<$& |[draw,fill=red!20]|$<$\\
    \\
    & TIFA160 & |[draw,fill=blue!20]|= & |[draw,fill=blue!20]|=& |[draw,fill=blue!20]|=\\
  };
  \node at ($(D-1-4)!0.2!(D-2-4)$) {\scriptsize Muse};
  \node at ($(D-4-1)!0.2!(D-5-2)$) {\scriptsize \rotatebox{90}{Imagen}};
\end{tikzpicture}
\begin{tikzpicture}
  \matrix(D)[matrix of nodes,nodes in empty cells,
             row sep=-\pgflinewidth,column sep=-\pgflinewidth,
             nodes={anchor=center},
             row 1/.style={minimum height=0.5cm},
             row 2/.style={minimum height=0.5cm},
             row 3/.style={minimum height=0.5cm},
             row 4/.style={minimum height=0.5cm},
             row 5/.style={minimum height=0.5cm},
             row 6/.style={minimum height=0.2cm},
             row 7/.style={minimum height=0.5cm},
             column 1/.style={minimum width=0.5cm},
             column 2/.style={minimum width=0.5cm},
             column 3/.style={minimum width=0.5cm},
             column 4/.style={minimum width=0.5cm},
             column 5/.style={minimum width=0.5cm},
            ]
  {%
    &  & \\
    \rotatebox{90}{WL} & \rotatebox{90}{L} & \rotatebox{90}{D(H)}\\
    |[draw,fill=blue!20]|= & |[draw,fill=blue!20]|= & |[draw,fill=blue!20]|= \\
    |[draw,fill=red!20]|$<$ & |[draw,fill=red!20]|$<$& |[draw,fill=green!20]|$>$\\
    |[draw,fill=blue!20]|= & |[draw,fill=red!20]|$<$& |[draw,fill=blue!20]|=\\
    \\
    |[draw,fill=blue!20]|= & |[draw,fill=blue!20]|=& |[draw,fill=green!20]|$>$\\
  };
  \node at ($(D-1-2)!0.2!(D-2-2)$) {\scriptsize SDXL};
\end{tikzpicture}
\begin{tikzpicture}
  \matrix(D)[matrix of nodes,nodes in empty cells,
             row sep=-\pgflinewidth,column sep=-\pgflinewidth,
             nodes={anchor=center},
             row 1/.style={minimum height=0.5cm},
             row 2/.style={minimum height=0.5cm},
             row 3/.style={minimum height=0.5cm},
             row 4/.style={minimum height=0.5cm},
             row 5/.style={minimum height=0.5cm},
             row 6/.style={minimum height=0.2cm},
             row 7/.style={minimum height=0.5cm},
             column 1/.style={minimum width=0.5cm},
             column 2/.style={minimum width=0.5cm},
             column 3/.style={minimum width=0.5cm},
             column 4/.style={minimum width=0.5cm},
             column 5/.style={minimum width=0.5cm},
            ]
  {%
    &  & \\
    \rotatebox{90}{WL} & \rotatebox{90}{L} & \rotatebox{90}{D(H)}\\
    |[draw,fill=green!20]|$>$ & |[draw,fill=green!20]|$>$ & |[draw,fill=green!20]|$>$ \\
    |[draw,fill=red!20]|$<$ & |[draw,fill=red!20]|$<$& |[draw,fill=green!20]|$>$\\
    |[draw,fill=green!20]|$>$ & |[draw,fill=green!20]|$>$& |[draw,fill=green!20]|$>$\\
    \\
    |[draw,fill=green!20]|$>$ & |[draw,fill=green!20]|$>$& |[draw,fill=green!20]|$>$\\
  };
  \node at ($(D-1-2)!0.2!(D-2-2)$) {\scriptsize SD1.5};
\end{tikzpicture}
\begin{tikzpicture}
  \matrix(D)[matrix of nodes,nodes in empty cells,
             row sep=-\pgflinewidth,column sep=-\pgflinewidth,
             nodes={anchor=center},
             row 1/.style={minimum height=0.5cm},
             row 2/.style={minimum height=0.5cm},
             row 3/.style={minimum height=0.5cm},
             row 4/.style={minimum height=0.5cm},
             row 5/.style={minimum height=0.5cm},
             row 6/.style={minimum height=0.2cm},
             row 7/.style={minimum height=0.5cm},
             column 1/.style={minimum width=0.5cm},
             column 2/.style={minimum width=0.5cm},
             column 3/.style={minimum width=0.5cm},
             column 4/.style={minimum width=0.5cm},
             column 5/.style={minimum width=0.5cm},
            ]
  {%
    &  &  & \\
    &  \rotatebox{90}{WL} & \rotatebox{90}{L} & \rotatebox{90}{D(H)}\\
    & |[draw,fill=green!20]|$>$ & |[draw,fill=green!20]|$>$ & |[draw,fill=green!20]|$>$ \\
    & |[draw,fill=blue!20]|= & |[draw,fill=blue!20]|=& |[draw,fill=green!20]|$>$\\
    & |[draw,fill=green!20]|$>$ & |[draw,fill=blue!20]|=& |[draw,fill=green!20]|$>$\\
    \\
    & |[draw,fill=blue!20]|= & |[draw,fill=blue!20]|=& |[draw,fill=green!20]|$>$\\
  };
  \node at ($(D-1-3)!0.2!(D-2-3)$) {\scriptsize SDXL};
  \node at ($(D-4-1)!0.2!(D-5-2)$) {\scriptsize \rotatebox{90}{Muse}};
\end{tikzpicture}
\begin{tikzpicture}
  \matrix(D)[matrix of nodes,nodes in empty cells,
             row sep=-\pgflinewidth,column sep=-\pgflinewidth,
             nodes={anchor=center},
             row 1/.style={minimum height=0.5cm},
             row 2/.style={minimum height=0.5cm},
             row 3/.style={minimum height=0.5cm},
             row 4/.style={minimum height=0.5cm},
             row 5/.style={minimum height=0.5cm},
             row 7/.style={minimum height=0.5cm},
             row 6/.style={minimum height=0.2cm},
             column 1/.style={minimum width=0.5cm},
             column 2/.style={minimum width=0.5cm},
             column 3/.style={minimum width=0.5cm},
             column 4/.style={minimum width=0.5cm},
             column 5/.style={minimum width=0.5cm},
            ]
  {%
    &  & \\
    \rotatebox{90}{WL} & \rotatebox{90}{L} & \rotatebox{90}{D(H)}\\
    |[draw,fill=green!20]|$>$ & |[draw,fill=green!20]|$>$ & |[draw,fill=green!20]|$>$ \\
    |[draw,fill=green!20]|$>$ & |[draw,fill=blue!20]|=& |[draw,fill=green!20]|$>$\\
    |[draw,fill=green!20]|$>$ & |[draw,fill=green!20]|$>$& |[draw,fill=green!20]|$>$\\
    \\
    |[draw,fill=blue!20]|= & |[draw,fill=green!20]|$>$& |[draw,fill=green!20]|$>$\\
  };
  \node at ($(D-1-2)!0.2!(D-2-2)$) {\scriptsize SD1.5};
\end{tikzpicture}
\begin{tikzpicture}
  \matrix(D)[matrix of nodes,nodes in empty cells,
             row sep=-\pgflinewidth,column sep=-\pgflinewidth,
             nodes={anchor=center},
             row 1/.style={minimum height=0.5cm},
             row 2/.style={minimum height=0.5cm},
             row 3/.style={minimum height=0.5cm},
             row 4/.style={minimum height=0.5cm},
             row 5/.style={minimum height=0.5cm},
             row 6/.style={minimum height=0.2cm},
             row 7/.style={minimum height=0.5cm},
             column 1/.style={minimum width=0.5cm},
             column 2/.style={minimum width=0.5cm},
             column 3/.style={minimum width=0.5cm},
             column 4/.style={minimum width=0.5cm},
             column 5/.style={minimum width=0.5cm},
            ]
  {%
    &  &  & \\
    &  \rotatebox{90}{WL} & \rotatebox{90}{L} & \rotatebox{90}{D(H)}\\
    & |[draw,fill=green!20]|$>$ & |[draw,fill=green!20]|$>$ & |[draw,fill=green!20]|$>$ \\
    & |[draw,fill=blue!20]|= & |[draw,fill=blue!20]|=& |[draw,fill=green!20]|$>$\\
    & |[draw,fill=green!20]|$>$ & |[draw,fill=blue!20]|=& |[draw,fill=green!20]|$>$\\
    \\
    & |[draw,fill=green!20]|$>$ & |[draw,fill=green!20]|$>$& |[draw,fill=blue!20]|=\\
  };
  \node at ($(D-1-3)!0.2!(D-2-3)$) {\scriptsize SD1.5};
  \node at ($(D-4-1)!0.2!(D-5-2)$) {\scriptsize \rotatebox{90}{SDXL}};
\end{tikzpicture}
\caption{\textbf{Comparing models using human annotations.} We compare model rankings on the reliable subsets of Gecko(S) (G(S)-rel), Gecko(R) (G(R)-rel), both subsets (G2K-rel), and TIFA160.  We perform the Wilcoxon signed-rank test for all pairs of models (p$<$0.001) and post-hoc comparison based on average ratings. 
Each grid represents a comparison between two models. Entries in the grid depict results for WL, Likert (L), and DSG(H) (D(H)) scores. The $>$ sign indicates the left-side model is better, worse ($<$), or not significantly different (=) than the model on the top. All three versions of Gecko prompts are able to better distinguish models by finding more significant comparisons between them.}
\label{fig:h_eval_wilcoxon_tifa}
\end{figure}

\section{Auto-eval metrics: Additional experiments on T2I}
\label{app:additional_experimental_results}

\subsection{Intuitive explanation of each task}\label{app:intuitive_explanation_task}
\label{app:intuitive_explanation_tasks}
We use this section to give an intuitive explanation of each task -- {\em \pointscore, \pairscore, \paircompare} -- as well as how they can lead to different outcomes in terms of metric ranking on a toy setting.
We evaluate the different auto-eval metrics on our actual dataset in \Section{sec:experiments}, where we find that {\em in practice} metrics do achieve different rankings on these tasks.

{\bf \Pointscore.} \Pointscore~evaluates how well a metric ranks generations for a single model. Assume we have a set of generations for Model $A$ over a prompt set $\mathcal{P}$ and metric $m$ that calculates scores $m(A(p), p)$ for a generation $A(p)$ for a prompt $p \in \mathcal{P}$. We also have a human rating $h(A(p), p)$ on some numerical scale between $[0, 1]$ where $1$ indicates a perfect generation and $0$ a terrible one. This evaluation compares how well the metric scores correlate with the human ones over that prompt set for Model $A$. If we have multiple models, we average the correlation coefficients obtained for each model.

{\bf \Pairscore.} \Pairscore~compares two metrics given a pair of generations for a prompt. Assume we have two generations $A(p), B(p)$ for the prompt $p$ for Model $A$ and Model $B$. We also have a human preference (e.g.~Model $A$ $>$ Model $B$), and a metric $m$ which gives a score (e.g. $m(A(p), p)$
) for a generation. Pairwise instance scoring would evaluate whether the relationship between the scores (i.e.~$m(A(p), p) > m(B(p), p)$) matches that of the human rating. If the relationship matches, then we say $m$ is correct for that example, else it is incorrect. To get an average accuracy for a given metric within a dataset, we count the number of examples for which $m$ predicts the correct relationship for all prompts / model pairs and divide by the number of comparisons.

{\bf \Paircompare.} For \paircompare, we compare two metrics over a set of prompts and corresponding generations. We average the scores for a metric across all prompts to get 
$m_{\textit{avg}}(A) = \sum_p m(A(p), p) / |\mathcal{P}|$. We then evaluate (for all model pairs) how often the model ordering from human evaluation matches that obtained when comparing a given model pair: e.g.~$m_{\textit{avg}}(A)$ vs $m_{\textit{avg}}(B)$ for Model $A$ and Model $B$.

{\bf Toy example.}
Why these different evaluation procedures can give different results for a metric is subtle. Consider the following toy setting. We have two models: $A$ and $B$, and 3 prompts. For prompt $p_1$, Model $A$ is clearly much better than Model $B$, which is terrible, but for prompts $p_2, p_3$, Model $A$ is ever so slightly worse but both are reasonable. Intuitively, Model $A$ is better than Model $B$ on this prompt set.

\begin{enumerate}
\item {\em Example 1}: Now imagine we have a metric that can do a good job of doing pairwise instance scoring but is not well calibrated across prompts (e.g. a score does not indicate an overall notion of alignment – 0.7 for one prompt could correspond to a poor generation but a score of 0.5 on another denotes a high quality generation). So in this toy example, that metric gives the following scores for Model $A$: $p_1 = 0.1, p_2=0.1, p_3=0.1$ and for Model $B$: $p_1=0.05, p_2=0.8, p_3=0.8$. The comparisons are all right, but the average score for Model $A$ is $0.1$ and Model $B$ is $0.55$, which does not match the intuition that Model $A$ is actually better than Model $B$.
\item {\em Example 2}: We can conversely have a metric that is well calibrated across prompts but not able to reliably pick up on subtle differences. Such a metric could have the following scores for Model $A$: $p_1=0.8, p_2=0.7,p_3=0.72$ whereas Model $B$: $p_1=0.1, p_2=0.69, p_3=0.71$. While the model ordering is correct, the pairwise comparisons for prompts $p_2, p_3$ are not.
\end{enumerate}

These examples demonstrate how a metric can be good at pairwise comparison (e.g. a metric like CLIP) but be poor at model ordering, i.e.~a metric can give scores that are not well calibrated across prompts. Similarly, a metric can be good at model ordering but bad at pairwise instance scoring because it does not capture subtle differences. We can use a similar logic to understand why the \pointscore~and \pairscore~tasks can achieve differing results as well as the \paircompare~and \pointscore~tasks.

\subsection{Model-ordering Evaluation}\label{app:model_ordering_eval}
We provide detailed results for the model ordering experiment. The goal is to determine the each auto-eval metric can predict the significant relations found from human annotation.
We use G2K-rel as it is the largest subset with agreement among annotation templates across models.
We take the majority vote to determine the relationship and note that there is no template that disagrees with this vote but one template may find a significant relation where the others do not, or vice versa.
For each model pair, we compare distributions of auto-eval metric predictions using the same Wilcoxon signed rank test to get the relationship predicted by the metric.
We plot in \Table{tab:app:modelorderingresults} the results.
CLIP performs poorly, confusing wins with ties.
However, no other auto-eval metric confuses a win with a tie showing that these auto-eval metrics are already robust for this task.
Of these metrics, we see that TIFA and Gecko are able to get the most number of significant relationships right.

\begin{figure}[t]
\vspace{-0.44mm}
\scriptsize
\begin{tikzpicture}
  \matrix(D)[matrix of nodes,nodes in empty cells,
             row sep=-\pgflinewidth,column sep=-\pgflinewidth,
             nodes={anchor=center},
             row 1/.style={minimum height=0.5cm},
             row 2/.style={minimum height=0.5cm},
             row 3/.style={minimum height=0.5cm},
             row 4/.style={minimum height=0.5cm},
             row 5/.style={minimum height=0.5cm},
             row 6/.style={minimum height=0.5cm},
             row 7/.style={minimum height=0.5cm},
             column 1/.style={minimum width=1.2cm},
             column 2/.style={minimum width=1.2cm},
             column 3/.style={minimum width=1.2cm},
             column 4/.style={minimum width=1.2cm},
             column 5/.style={minimum width=1.2cm},
             column 6/.style={minimum width=1.2cm},
             column 7/.style={minimum width=1.2cm},
             column 8/.style={minimum width=1.2cm},
             column 9/.style={minimum width=1.2cm},
             column 10/.style={minimum width=1.2cm},
             column 11/.style={minimum width=1.2cm},
            ]
  {%
    {\bf Model 1} & {\bf Model 2}  & & `GT' & & \model & DSG & TIFA & CLIP & VNLI & VQAScore \\
    Imagen & Muse & &  |[draw,fill=red!20]| $<$ & & |[draw,fill=red!20]| $<$ & |[draw,fill=red!20]| $<$ & |[draw,fill=red!20]| $<$  & |[draw,fill=red!20]| $<$   & |[draw,fill=red!20]| $<$   & |[draw,fill=red!20]| $<$   \\
    Imagen & SDXL & &  |[draw,fill=yellow!20]| $--$ & & |[draw,fill=red!20]| $<$ & |[draw,fill=red!20]| $<$ & |[draw,fill=yellow!20]| $--$  & |[draw,fill=red!20]| $<$   & |[draw,fill=yellow!20]| $--$   & |[draw,fill=red!20]| $<$   \\
    Imagen & SD1.5 & &  |[draw,fill=green!20]| $>$ & & |[draw,fill=green!20]| $>$ & |[draw,fill=green!20]| $>$ & |[draw,fill=green!20]| $>$  & |[draw,fill=red!20]| $<$   & |[draw,fill=green!20]| $>$   & |[draw,fill=green!20]| $>$   \\
    Muse & SDXL & &  |[draw,fill=green!20]| $>$ & & |[draw,fill=green!20]| $>$ & |[draw,fill=yellow!20]| $--$ & |[draw,fill=yellow!20]| $--$  & |[draw,fill=red!20]| $<$   & |[draw,fill=yellow!20]| $--$    & |[draw,fill=yellow!20]| $--$   \\
    Muse & SD1.5 & &  |[draw,fill=green!20]| $>$ & & |[draw,fill=green!20]| $>$ & |[draw,fill=green!20]| $>$ & |[draw,fill=green!20]| $>$  & |[draw,fill=green!20]| $>$   & |[draw,fill=green!20]| $>$   & |[draw,fill=green!20]| $>$   \\
    SDXL & SD1.5 & &  |[draw,fill=green!20]| $>$ & & |[draw,fill=green!20]| $>$ & |[draw,fill=green!20]| $>$ & |[draw,fill=green!20]| $>$  & |[draw,fill=green!20]| $>$   & |[draw,fill=green!20]| $>$   & |[draw,fill=green!20]| $>$   \\
  };
\end{tikzpicture} 
\caption{\textbf{Comparing model ordering obtained from humans and auto-eval metrics on G2K-rel.} We show the `GT' human ordering and the predicted ones for auto-eval metrics. $<$ means Model 1 $<$ Model 2, $>$ Model 1 $>$ Model 2 and $--$ that no significant relation was found. While CLIP performs poorly, mistaking wins with losses, no other metric confuses a win with a loss.}
\label{tab:app:modelorderingresults}
\end{figure}

\subsection{Additional \pairscore~and \pointscore~results on Gecko2K, Gecko-Rel}
\label{app_sec:corr_pearson}

\begin{table}[t]
\renewcommand{\arraystretch}{1.2}
\setlength{\tabcolsep}{3pt}

\centering
\resizebox{\textwidth}{!}{%
\begin{tabular}{c|c|cccc|cccc|cccc|cccc}
\hline
                                  &                                       & \multicolumn{4}{c|}{\textbf{Gecko(R)}}                                                                                                     & \multicolumn{4}{c|}{\textbf{Gecko(S)}}                                                                                                                                & \multicolumn{4}{c|}{\textbf{Gecko(R)-Rel}}                                                                                                          & \multicolumn{4}{c}{\textbf{Gecko(S)-Rel}}                                                                                                                    \\ \cline{3-18} 
                                  &                                       & WL                          & Likert                      & \multicolumn{1}{c|}{DSG(H)}                      & SxS                         & WL                          & Likert                               & \multicolumn{1}{c|}{DSG(H)}                               & SxS                         & WL                          & Likert                      & \multicolumn{1}{c|}{DSG(H)}                      & SxS                         & WL                          & Likert                               & \multicolumn{1}{c|}{DSG(H)}                      & SxS                         \\ \cline{3-18} 
\multirow{-3}{*}{\textbf{Metrics}} & \multirow{-3}{*}{\textbf{FT}} & \multicolumn{3}{c|}{SpearmanR}                                                                               & Acc                         & \multicolumn{3}{c|}{SpearmanR}                                                                                                 & Acc                         & \multicolumn{3}{c|}{SpearmanR}                                                                               & Acc                         & \multicolumn{3}{c|}{SpearmanR}                                                                                        & Acc                         \\ \hline
{\em Interpretable (QA/VQA)} & & & & \multicolumn{1}{c|}{} & & & & \multicolumn{1}{c|}{} & & & & \multicolumn{1}{c|}{} & & & & \multicolumn{1}{c|}{} & \\
TIFA$_{\texttt{PALM-2/PALI}}$                                & \xmark                                 & 0.26                        & 0.34                        & \multicolumn{1}{c|}{0.28}                        & 41.7                        & 0.39                        & 0.32                                 & \multicolumn{1}{c|}{0.39}                                 & 53.2                        & 0.34                        & 0.37                        & \multicolumn{1}{c|}{0.33}                        & 47.3                        & 0.41                        & 0.36                                 & \multicolumn{1}{c|}{0.40}                        & 52.8                        \\
DSG$_{\texttt{PALM-2/PALI}}$                              & \xmark                                 & 0.35                        & 0.47                        & \multicolumn{1}{c|}{0.42}                        & 49.6                        & 0.45                        & 0.45                                 & \multicolumn{1}{c|}{{0.45}}                        & 58.1                        & 0.47                        & 0.50                        & \multicolumn{1}{c|}{0.50}                        & 53.9                        & 0.48                        & 0.47                                 & \multicolumn{1}{c|}{0.48}                        & 56.9                        \\ 
Gecko$_{\texttt{PALM-2/PALI}}$                              & \xmark                                 & \underline{0.41}               & \underline{0.55}               & \multicolumn{1}{c|}{\underline{0.46}}               & \underline{62.1}                        & \underline{0.47}               & \underline{0.52}                        & \multicolumn{1}{c|}{\underline{0.45}}                        & \underline{74.6}               & \underline{0.52}               & \underline{0.58}               & \multicolumn{1}{c|}{\underline{0.53}}               & \underline{71.3}               & \underline{0.52}               & \underline{0.54}                        & \multicolumn{1}{c|}{\underline{0.49}}               & \underline{75.2}               \\ \hline
Gecko$_{\texttt{Gemini Flash}}$                      & \xmark                                 & \underline{\textbf{0.43}}               & \underline{\textbf{0.58}}               & \multicolumn{1}{c|}{\underline{\textbf{0.48}}}               & \underline{72.2}                        & \underline{\textbf{0.54}}               & \underline{\textbf{0.59}}                        & \multicolumn{1}{c|}{\underline{\textbf{0.56}}}                        & \underline{\textbf{78.8}}               & \underline{\textbf{0.56}}               & \underline{\textbf{0.62}}               & \multicolumn{1}{c|}{\underline{\textbf{0.58}}}               & \underline{\textbf{74.0}}               & \underline{\textbf{0.57}}               & \underline{\textbf{0.63}}                        & \multicolumn{1}{c|}{\underline{\textbf{0.57}}}               & \underline{\textbf{80.0}}               \\ \hline
{\em Uninterpretable (single score)} & & & & \multicolumn{1}{c|}{} & & & & \multicolumn{1}{c|}{} & & & & \multicolumn{1}{c|}{} & & & & \multicolumn{1}{c|}{} & \\

CLIP                               & \xmark                                 & 0.14                        & 0.16                        & \multicolumn{1}{c|}{0.13}                        & 54.4                        & 0.25                        & 0.18                                 & \multicolumn{1}{c|}{0.26}                                 & 67.2                        & 0.11                        & 0.09                        & \multicolumn{1}{c|}{0.08}                        & 59.7                        & 0.24                        & 0.19                                 & \multicolumn{1}{c|}{0.25}                        & 67.1                        \\
PyramidCLIP                        & \xmark                                 & 0.26                        & 0.27                        & \multicolumn{1}{c|}{0.26}                        & {64.3}               & 0.22                        & 0.25                                 & \multicolumn{1}{c|}{0.23}                                 & 70.7                        & 0.26                        & 0.26                        & \multicolumn{1}{c|}{0.23}                        & 65.8                        & 0.21                        & 0.26                                 & \multicolumn{1}{c|}{0.22}                        & 71.0                        \\
VQAScore$_\texttt{Gemini Flash}$                        & \xmark                                 & \underline{0.42}                        & \underline{0.54}                       & \multicolumn{1}{c|}{\underline{0.45}}                        & \underline{\textbf{73.1}}               & \underline{0.51}                        & \underline{0.57}                                 & \multicolumn{1}{c|}{\underline{0.49}}                                 & \underline{76.5}                        & \underline{0.51}                        & \underline{0.59}                        & \multicolumn{1}{c|}{\underline{0.52}}                        & \underline{73.9}                        & \underline{0.54}                        & 0.60                                 & \multicolumn{1}{c|}{\underline{0.51}}                        & \underline{77.0}                        \\
{\color[HTML]{9B9B9B} VNLI}        & {\color[HTML]{9B9B9B} \cmark}          & {\color[HTML]{9B9B9B} 0.37} & {\color[HTML]{9B9B9B} 0.49} & \multicolumn{1}{c|}{{\color[HTML]{9B9B9B} 0.42}} & {\color[HTML]{9B9B9B} 54.4} & {\color[HTML]{9B9B9B} 0.45} & {\color[HTML]{9B9B9B} {0.55}} & \multicolumn{1}{c|}{{\color[HTML]{9B9B9B} {0.45}}} & {\color[HTML]{9B9B9B} 72.7} & {\color[HTML]{9B9B9B} 0.49} & {\color[HTML]{9B9B9B} 0.57} & \multicolumn{1}{c|}{{\color[HTML]{9B9B9B} 0.46}} & {\color[HTML]{9B9B9B} 65.6} & {\color[HTML]{9B9B9B} 0.50} & {\color[HTML]{9B9B9B} {\underline 0.61}} & \multicolumn{1}{c|}{{\color[HTML]{9B9B9B} 0.48}} & {\color[HTML]{9B9B9B} 72.7} \\ \hline
Gecko+VQAScore$_{\texttt{Gemini Flash}}$                        & \xmark                                 & --                       & --                       & \multicolumn{1}{c|}{--}                        & {81.0}               & --                       & --                                & \multicolumn{1}{c|}{--}                                 & 86.3                        & --                       & --                        & \multicolumn{1}{c|}{--}                        & 82.7                        & --                        & --                                 & \multicolumn{1}{c|}{--}                        & 87.4                        \\ \hline
\end{tabular}%
}
\caption{\textbf{Correlation between VQA-based, contrastive, and fine-tuned (FT) auto-eval metrics and human ratings across annotation templates on Gecko2K and Gecko2K-Rel.}  We observe a similar trend in Gecko2k and Gecko2K-Rel:~\model~performs the best across the board, and it can be improved by using a better language/VQA backend.}
\label{tab:corr-full}
\end{table}

\begin{table}[h]
\renewcommand{\arraystretch}{1.2}
\setlength{\tabcolsep}{3pt}

\centering
\resizebox{\textwidth}{!}{%
\begin{tabular}{c|c|cccc|cccc|cccc|cccc}
\hline
                                   &                                       & \multicolumn{4}{c|}{\textbf{Gecko(R)}}                                                                                                     & \multicolumn{4}{c|}{\textbf{Gecko(S)}}                                                                                                                                & \multicolumn{4}{c|}{\textbf{Gecko(R)-Rel}}                                                                                                          & \multicolumn{4}{c}{\textbf{Gecko(S)-Rel}}                                                                                                                    \\ \cline{3-18} 
                                   &                                       & WL                          & Likert                      & \multicolumn{1}{c|}{DSG(H)}                      & SxS                         & WL                          & Likert                               & \multicolumn{1}{c|}{DSG(H)}                               & SxS                         & WL                          & Likert                      & \multicolumn{1}{c|}{DSG(H)}                      & SxS                         & WL                          & Likert                               & \multicolumn{1}{c|}{DSG(H)}                      & SxS                         \\ \cline{3-18} 
\multirow{-3}{*}{\textbf{Metrics}} & \multirow{-3}{*}{\textbf{FT}} & \multicolumn{3}{c|}{Pearson}                                                                               & Acc                         & \multicolumn{3}{c|}{
Pearson}                                                                                                 & Acc                         & \multicolumn{3}{c|}{Pearson}                                                                               & Acc                         & \multicolumn{3}{c|}{Pearson}                                                                                        & Acc                         \\ \hline
{\em Interpretable (QA/VQA)} & & & & \multicolumn{1}{c|}{} & & & & \multicolumn{1}{c|}{} & & & & \multicolumn{1}{c|}{} & & & & \multicolumn{1}{c|}{} & \\
TIFA$_{\texttt{PALM-2/PALI}}$                                & \xmark                                 & 0.21                        & 0.32                        & \multicolumn{1}{c|}{0.25}                        & 41.7                        & 0.39                        & 0.32                                 & \multicolumn{1}{c|}{0.39}                                 & 53.2                        & 0.27                        & 0.35                        & \multicolumn{1}{c|}{0.27}                        & 47.3                        & 0.43                        & 0.35                                & \multicolumn{1}{c|}{0.41}                        & 52.8                        \\
DSG$_{\texttt{PALM-2/PALI}}$                              & \xmark                                 & 0.28                        & 0.41                        & \multicolumn{1}{c|}{0.38}                        & 49.6                        & 0.43                        & 0.42                                 & \multicolumn{1}{c|}{0.44}                        & 58.1                        & 0.39                        & 0.43                        & \multicolumn{1}{c|}{0.44}                        & 53.9                        & 0.45                        & 0.43                                 & \multicolumn{1}{c|}{0.46}                        & 56.9                        \\ 
Gecko$_{\texttt{PALM-2/PALI}}$                               & \xmark                                 & {0.38}               & {0.51}               & \multicolumn{1}{c|}{{0.42}}               & 62.1                        & {0.46}               & {0.48}                        & \multicolumn{1}{c|}{{0.46}}                        & {74.6}               & {0.50}               & {0.55}               & \multicolumn{1}{c|}{{0.51}}               & {71.3}               & {0.52}               & {0.52}                        & \multicolumn{1}{c|}{{0.50}}               & {75.2}               \\ 

Gecko$_{\texttt{Gemini Flash}}$                               & \xmark                                 & \underline{\textbf{0.40}}               & \underline{\textbf{0.53}}               & \multicolumn{1}{c|}{\underline{\textbf{0.47}}}               & \textbf{72.7}                        & \underline{\textbf{0.52}}               & \underline{\textbf{0.58}}                        & \multicolumn{1}{c|}{\underline{\textbf{0.56}}}                        & \underline{\textbf{77.9}}               & \underline{\textbf{0.53}}               & \underline{\textbf{0.57}}               & \multicolumn{1}{c|}{\underline{\textbf{0.54}}}               & \textbf{74.0}               & \underline{\textbf{0.56}}               & \underline{\textbf{0.61}}                        & \multicolumn{1}{c|}{\underline{\textbf{0.56}}}               & \underline{\textbf{80.0}}               \\ \hline
{\em Uninterpretable (single score)} & & & & \multicolumn{1}{c|}{} & & & & \multicolumn{1}{c|}{} & & & & \multicolumn{1}{c|}{} & & & & \multicolumn{1}{c|}{} & \\
CLIP                               & \xmark                                 & 0.15                        & 0.18                        & \multicolumn{1}{c|}{0.16}                        & 54.4                        & 0.26                        & 0.19                                 & \multicolumn{1}{c|}{0.25}                                 & 67.2                        & 0.13                        & 0.12                        & \multicolumn{1}{c|}{0.10}                        & 59.7                        & 0.26                        & 0.19                                 & \multicolumn{1}{c|}{0.25}                        & 67.1                        \\
PyramidCLIP                        & \xmark                                 & 0.29                        & 0.30                        & \multicolumn{1}{c|}{0.26}                        & {64.3}               & 0.28                        & 0.27                                 & \multicolumn{1}{c|}{0.25}                                 & 70.7                        & 0.31                        & 0.28                        & \multicolumn{1}{c|}{0.25}                        & 65.8                        & 0.28                        & 0.28                                 & \multicolumn{1}{c|}{0.26}                        & 71.0                        \\
VQAScore$_\texttt{Gemini Flash}$                        & \xmark                                 & \textbf{0.35}                        & {0.44}                       & \multicolumn{1}{c|}{{0.36}}                        & \underline{\textbf{73.1}}               & \textbf{0.42}                        & {0.53}                                 & \multicolumn{1}{c|}{\textbf{0.41}}                                 & \textbf{76.5}                        & \textbf{0.41}                        & \textbf{0.47}                        & \multicolumn{1}{c|}{\textbf{0.42}}                        & \underline{\textbf{73.9}}                        & \textbf{0.46}                        & \textbf{0.56}                                 & \multicolumn{1}{c|}{\textbf{0.42}}                        & \textbf{77.0}                        \\
{\color[HTML]{9B9B9B} VNLI}        & {\color[HTML]{9B9B9B} \cmark}          & {\color[HTML]{9B9B9B} 0.34} & {\color[HTML]{9B9B9B} \textbf{0.48}} & \multicolumn{1}{c|}{{\color[HTML]{9B9B9B} \textbf{0.39}}} & {\color[HTML]{9B9B9B} 54.4} & {\color[HTML]{9B9B9B} 0.41} & {\color[HTML]{9B9B9B} \textbf{0.55}} & \multicolumn{1}{c|}{{\color[HTML]{9B9B9B} 0.42}} & {\color[HTML]{9B9B9B} 72.7} & {\color[HTML]{9B9B9B} 0.25} & {\color[HTML]{9B9B9B} 0.41} & \multicolumn{1}{c|}{{\color[HTML]{9B9B9B} 0.22}} & {\color[HTML]{9B9B9B} 65.6} & {\color[HTML]{9B9B9B} 0.35} & {\color[HTML]{9B9B9B}0.49} & \multicolumn{1}{c|}{{\color[HTML]{9B9B9B} 0.34}} & {\color[HTML]{9B9B9B} 72.7} \\ \hline
Gecko+VQAScore$_{\texttt{Gemini Flash}}$                        & \xmark                                 & --                       & --                       & \multicolumn{1}{c|}{--}                        & {81.0}               & --                       & --                                & \multicolumn{1}{c|}{--}                                 & 86.3                        & --                       & --                        & \multicolumn{1}{c|}{--}                        & 82.7                        & --                        & --                                 & \multicolumn{1}{c|}{--}                        & 87.4                        \\ \hline
\end{tabular}%
}
\vspace{3mm}
\caption{\textbf{Pearson correlation between VQA-based, contrastive, and fine-tuned (FT) auto-eval metrics and human ratings across annotation templates on Gecko2K and Gecko2K-Rel.} Similar to the comparisons on Spearman correlation, Gecko again outperforms other auto-eval metrics (both QA/VQA and single score)  with higher overall Pearson correlation. Swapping for the GeminiFlash backend leads to consistent performance improvement across templates. }
\label{tab:corr_pearson}
\vspace{-5mm}
\end{table} 
We report the Spearman Rank correlation of different auto-eval metrics on Gecko2K in \Table{tab:corr}. Here we report the Pearson correlation in \Table{tab:corr_pearson} as well and both Spearman and Pearson on Gecko-Rel in \Table{tab:corr-full}. Results follow those in the paper: the \model~metric is consistently best though VQAScore is a strong baseline. While DSG/TIFA perform better than CLIP on the absolute templates, CLIP performs better on SxS.

In SxS comparison, to investigate how the interpretable and uninterpretable metrics can be combined to achieve better results, we also take the samples on which \model~and VQAScore agree, and compute the accuracy of prediction on them. We found that we can improve agreement to  $>$80\% for this subset.

\subsection{Results for Additional CLIP Metrics on the Gecko Benchmark} \label{app:all_correlations}

\begin{table}[]
\renewcommand{\arraystretch}{1.4}
\centering
\resizebox{\textwidth}{!}{%
\begin{tabular}{l|cc|cc|cc|cc|cc|cc}
\toprule
\multicolumn{1}{l|}{\multirow{3}{*}{\textbf{Metrics}}} & \multicolumn{6}{c|}{\textbf{Gecko (R)}}        & \multicolumn{6}{c}{\textbf{Gecko (S)}}        \\
\multicolumn{1}{c|}{} &
  \multicolumn{2}{c|}{WL} &
  \multicolumn{2}{c|}{Likert} &
  \multicolumn{2}{c|}{DSG(H)} &
  \multicolumn{2}{c|}{WL} &
  \multicolumn{2}{c|}{Likert} &
  \multicolumn{2}{c}{DSG(H)} \\
\cline{2-13}
\multicolumn{1}{c|}{} &
  \multicolumn{1}{c}{Pearson} &
  \multicolumn{1}{c|}{Spearman-R} &
  \multicolumn{1}{c}{Pearson} &
  \multicolumn{1}{c|}{Spearman-R} &
  \multicolumn{1}{c}{Pearson} &
  \multicolumn{1}{c|}{Spearman-R} &
  \multicolumn{1}{c}{Pearson} &
  \multicolumn{1}{c|}{Spearman-R} &
  \multicolumn{1}{c}{Pearson} &
  \multicolumn{1}{c|}{Spearman-R} &
  \multicolumn{1}{c}{Pearson} &
  \multicolumn{1}{c}{Spearman-R} \\
\midrule
BLIP-2$_\text{ITM}$ &
  0.25 &
  0.22 &
  0.24 &
  0.19 &
  0.23 &
  0.21 &
  0.28 &
  0.23 &
  0.13 &
  0.16 &
  0.25 &
  0.23 \\
CLIP-B/32 &
0.15                        & \multicolumn{1}{c|}{0.14}                        & 0.18                        & \multicolumn{1}{c|}{0.16}                                 & 0.16                        & \multicolumn{1}{c|}{0.13}                                               & 0.26                        & \multicolumn{1}{c|}{0.25}                        & 0.19                        & \multicolumn{1}{c|}{0.18}                                 & 0.25                        & \multicolumn{1}{c}{0.26}                                           \\
CLIP-B/32$_\text{LAION-2B}$ &
  0.26 &
  0.21 &
  0.24 &
  0.22 &
  0.26 &
  0.21 &
  0.28 &
  0.24 &
  0.23 &
  0.22 &
  0.26 &
  0.24 \\
CLIP-B/16 &
  0.16 &
  0.11 &
  0.15 &
  0.12 &
  0.17 &
  0.12 &
  0.26 &
  0.22 &
  0.15 &
  0.14 &
  0.24 &
  0.22 \\
CLIP-L/14 &
  0.18 &
  0.16 &
  0.16 &
  0.14 &
  0.18 &
  0.15 &
  0.26 &
  0.23 &
  0.17 &
  0.16 &
  0.25 &
  0.24 \\
CLIP-H/14$_\text{LAION-2B}$ &
  0.29 &
  0.24 &
  0.25 &
  0.22 &
  0.27 &
  0.23 &
  0.29 &
  0.24 &
  0.23 &
  0.22 &
  0.27 &
  0.25 \\
CLIP-g/14$_\text{LAION-2B}$ &
  \underline{0.30} &
  0.24 &
  0.26 &
  0.23 &
  0.28 &
  0.23 &
  \underline{0.30} &
  0.25 &
  0.24 &
  0.23 &
  0.28 &
  0.25 \\
CLIP-G/14$_\text{LAION-2B}$ &
  0.29 &
  0.23 &
  0.25 &
  0.21 &
  0.27 &
  0.23 &
  \underline{0.30} &
  0.24 &
  0.25 &
  0.23 &
  0.28 &
  0.25 \\
CoCa-L/14 &
  0.28 &
  0.23 &
  0.26 &
  0.22 &
  0.26 &
  0.21 &
  0.29 &
  0.25 &
  0.24 &
  0.22 &
  0.28 &
  0.26 \\
EVA-02-CLIP-L/14 &
  0.27 &
  0.24 &
  0.23 &
  0.21 &
  0.24 &
  0.22 &
  \underline{0.30} &
  \underline{0.26} &
  0.21 &
  0.20 &
  0.28 &
  \underline{0.27} \\
EVA-02-CLIP-E/14 &
  0.28 &
  0.23 &
  0.24 &
  0.20 &
  0.27 &
  0.22 &
  0.28 &
  0.23 &
  0.23 &
  0.22 &
  0.26 &
  0.24 \\
EVA-02-CLIP-E/14+ &
  \underline{0.30} &
  0.24 &
  0.24 &
  0.21 &
  0.27 &
  0.23 &
  0.29 &
  0.24 &
  0.25 &
  0.23 &
  0.28 &
  0.25 \\
SigLIP-B/16 &
  0.26 &
  0.21 &
  0.22 &
  0.18 &
  0.27 &
  0.21 &
  0.29 &
  0.25 &
  0.22 &
  0.21 &
  \underline{0.29} &
  0.26 \\
SigLIP-L/16 &
  0.28 &
  0.24 &
  0.26 &
  0.22 &
  \underline{0.29} &
  0.25 &
  0.29 &
  \underline{0.26} &
  0.23 &
  0.22 &
  \underline{0.29} &
  \underline{0.27} \\
PyramidCLIP-B/16 &
  0.29 &
  \underline{0.26} &
  \underline{0.30} &
  \underline{0.27} &
  \underline{0.29} &
  \underline{0.26} &
  0.28 &
  0.22 &
  \underline{0.27} &
  \underline{0.25} &
  0.25 &
  0.23 \\
X-VLM$_\text{16M}$ &
  0.17 &
  0.11 &
  0.21 &
  0.09 &
  0.25 &
  0.15 &
  0.26 &
  0.23 &
  0.23 &
  0.16 &
  0.24 &
  0.23 \\
 \midrule
 
TIFA$_{\texttt{PALM-2/PALI}}$                                                   & 0.21                        & \multicolumn{1}{c|}{0.26}                        & 0.32                        & \multicolumn{1}{c|}{0.34}                                 & 0.25                        & \multicolumn{1}{c|}{0.28}                                             & 0.39                        & \multicolumn{1}{c|}{0.39}                        & 0.32                        & \multicolumn{1}{c|}{0.32}                                 & 0.39                        & \multicolumn{1}{c}{0.39}                                              \\
DSG$_{\texttt{PALM-2/PALI}}$                                               & 0.28                        & \multicolumn{1}{c|}{0.35}                        & 0.41                        & \multicolumn{1}{c|}{0.47}                                 & 0.38                        & \multicolumn{1}{c|}{0.42}                                              & {0.43}               & \multicolumn{1}{c|}{0.45}                        & 0.42 & \multicolumn{1}{c|}{0.45}                                 & 0.44                        & \multicolumn{1}{c}{{0.45}}               \\ 
Gecko$_{\texttt{PALM-2/PALI}}$                                                & {0.38}               & \multicolumn{1}{c|}{{0.41}}               & {0.51}               & \multicolumn{1}{c|}{{0.55}}                        & {0.42}               & \multicolumn{1}{c|}{{0.46}}                                    & {0.46}                        & \multicolumn{1}{c|}{{0.47}}               & {0.48}               & \multicolumn{1}{c|}{{0.52}}                        & {0.46}               & \multicolumn{1}{c}{{0.45}}                                               \\ 
Gecko$_{\texttt{Gemini Flash}}$                                                & \textbf{0.40}               & \multicolumn{1}{c|}{\textbf{0.42}}               & \textbf{0.53}               & \multicolumn{1}{c|}{\textbf{0.57}}                        & \textbf{0.47}               & \multicolumn{1}{c|}{\textbf{0.47}}                                    & \textbf{0.52}                        & \multicolumn{1}{c|}{\textbf{0.54}}               & \textbf{0.58}               & \multicolumn{1}{c|}{\textbf{0.60}}                        & \textbf{0.56}               & \multicolumn{1}{c}{\textbf{0.56}}                                               \\ 
\hline
VQAScore$_{\texttt{Gemini Flash}}$                                                & {0.35}               & \multicolumn{1}{c|}{\textbf{0.42}}               & {0.44}               & \multicolumn{1}{c|}{{0.54}}                        & {0.36}               & \multicolumn{1}{c|}{{0.45}}                                    & {0.42}                        & \multicolumn{1}{c|}{{0.51}}               & {0.53}               & \multicolumn{1}{c|}{{0.57}}                        & {0.41}               & \multicolumn{1}{c}{{0.49}}                                               \\ 
{\color[HTML]{9B9B9B} VNLI}                & {\color[HTML]{9B9B9B} 0.34} & \multicolumn{1}{c|}{{\color[HTML]{9B9B9B} 0.37}} & {\color[HTML]{9B9B9B} 0.48} & \multicolumn{1}{c|}{{\color[HTML]{9B9B9B} 0.49}} & {\color[HTML]{9B9B9B} 0.39} & \multicolumn{1}{c|}{{\color[HTML]{9B9B9B} 0.42}} & {\color[HTML]{9B9B9B} 0.41} & \multicolumn{1}{c|}{{\color[HTML]{9B9B9B}{ 0.45}}} & {\color[HTML]{9B9B9B} {0.55}} & \multicolumn{1}{c|}{{\color[HTML]{9B9B9B} {0.55}}} & {\color[HTML]{9B9B9B} 0.42} & \multicolumn{1}{c}{{\color[HTML]{9B9B9B} 0.45}}  \\ 

\bottomrule
\end{tabular}%
}
\vspace{1mm}
\caption{\textbf{Correlation between auto-eval metrics and human ratings across three annotation templates on Gecko2K.} Best results per model type are \underline{underlined}; best results are in \textbf{bold}.}
\label{tab:gecko_corr_all}
\end{table}

We compare the \model{} metric with several score-based auto-eval metrics \citep{li2023blip2,ilharco_gabriel_2021_5143773,yu2022coca,sun2023eva,zhai2023sigmoid,gao2022pyramidclip,zeng2022xvlm}, as well as QA-based metrics such as TIFA \citep{hu2023tifa} and DSG \citep{cho2023davidsonian}, and VNLI models \citep{yarom2024you} in \Table{tab:gecko_corr_all}.

Generally, our \model~metric outperforms the others and shows a higher correlation with most of the human annotation templates.
DSG is the second best metric, except on SxS where it ranks third. It outperforms TIFA by a clear margin but falls behind \model.
Finally, we note that \model~even shows higher correlation than the supervised VNLI model.
By using a stronger, Gemini Flash backend, \model~performs best by a significant margin consistently.

Looking at efficient, score-based metrics, we find that PyramidCLIP achieves competitive correlations.
Moreover, a larger pre-training corpus leads to better metrics; \eg, as seen by comparing CLIP-B/32 (trained on 400M images) and CLIP-B/32$_\text{LAION-2B}$ (trained on 2B images).
Finally, larger models are often better (\eg, SigLIP-L/16 vs. SigLIP-B/16), although these trends are less consistent (\eg, EVA-02-CLIP-L/14+ $\approx$ EVA-02-CLIP-E/14).

\subsection{Analysing Auto-Eval Metric results per skill.}
\label{app:metricpercategory}
\begin{figure*}
    \centering
    \begin{subfigure}[t]{0.33\linewidth}
        \includegraphics[width=\linewidth]{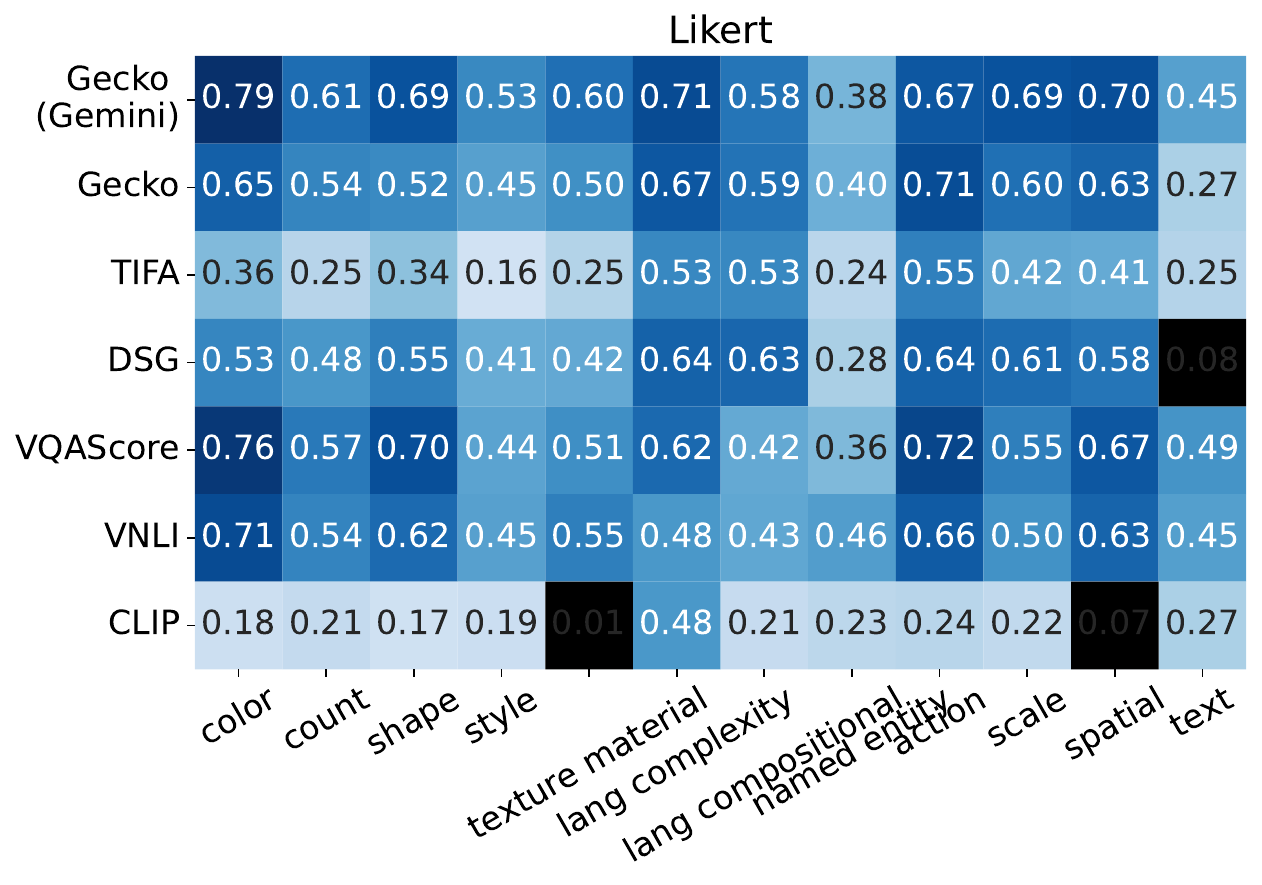}
        \caption{Likert}
    \end{subfigure}%
    ~
    \begin{subfigure}[t]{0.33\linewidth}
        \centering
        \includegraphics[width=\linewidth]{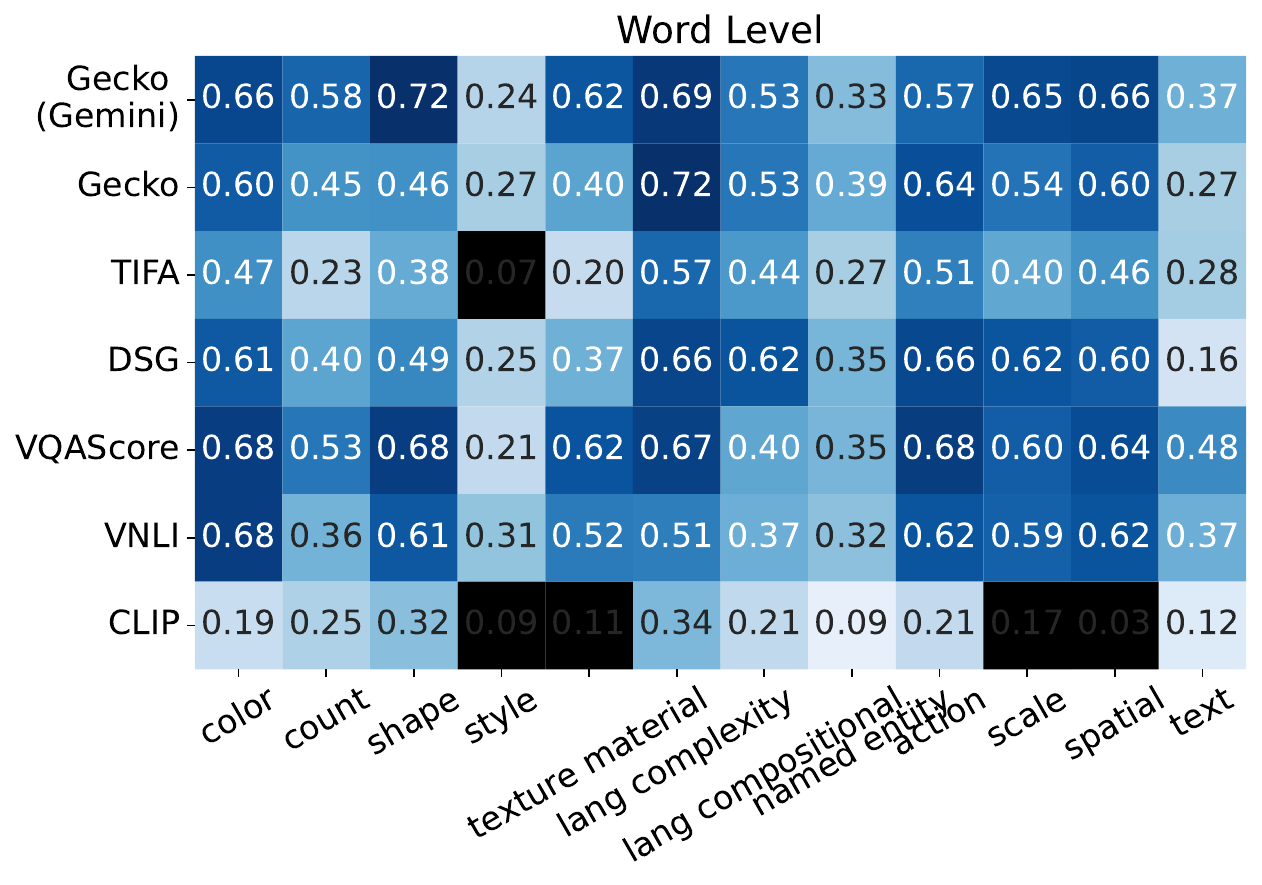}
        \caption{Word-level}
    \end{subfigure}%
    \begin{subfigure}[t]{0.33\linewidth}
        \centering
        \includegraphics[width=\linewidth]{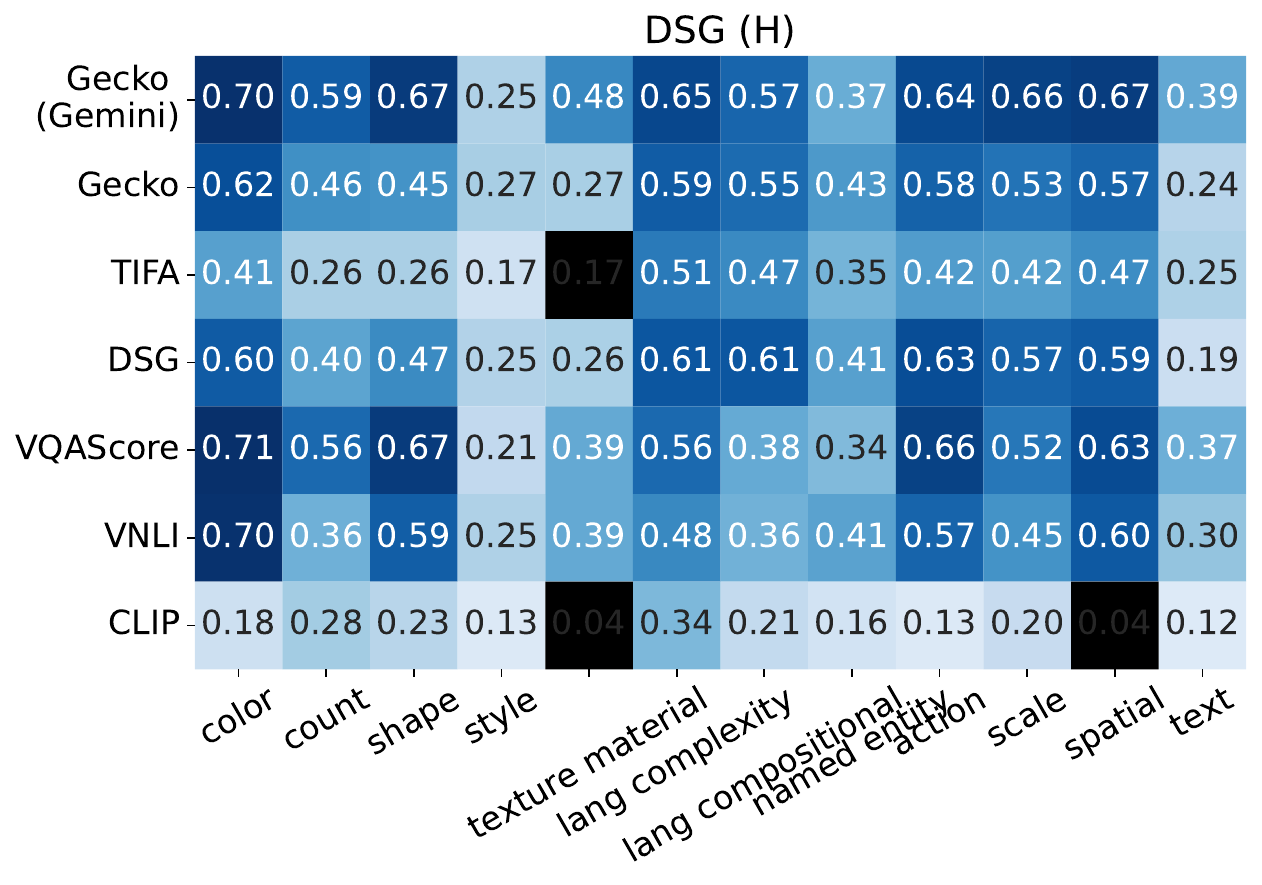}
        \caption{DSG(H)}
    \end{subfigure}%

    \caption{{\bf Per skill results by metric.} We visualise the correlation for each skill. Where p-values are $>0.05$, we color the square black.}
    \label{fig:all_results_by_category_high_level}
\end{figure*}

We present the per-skill Spearman Ranked Correlation between different auto-eval metrics and human annotation templates in \Figure{fig:all_results_by_category_high_level}. We observe a similar trend across the three plots, as we discussed in \Section{sec:experiments}: Gecko is the best on handling prompts with ``language complexity'', which can be attributed to the coverage tagging and filtering steps in its pipeline that make Gecko less prone to errors when processing long and complicated prompts.
DSG is better on ``compositional prompts'', as it can leverage its utilization of dependency graphs. VNLI and VQAScore demonstrate advantages in assessing ``shape'', ``color'', and ``text'' (\eg, \textsc{text rendering}) prompts though we note they are both worse on the more complex prompts (\eg, ``compositional'' and ``language complexity'').
When leveraging a better QA/VQA model (\eg, GeminiFlash) for \model, we see improvements across the board; \model~with GeminiFlash performs consistently the same or better than VNLI/VQAScore. 
TIFA and CLIP consistently perform poorly and worse than the other metrics.
It is also worth noting that all metrics exhibit relatively poor performance on ``text'', ``style'', and ``named identity'', highlighting the current lack of OCR and named recognition ability in existing contrastive, NLI and VQA models.

\subsection{Additional visualisations.}
\label{app:additionalvisualisations}
\begin{figure}[t]
\tiny
\centering
\renewcommand{\arraystretch}{0.8}
\vspace{3mm}
\begin{tabular}{c|cccc|cccc}
    {\bf Prompt:} & \multicolumn{4}{c}{A snake is on the elephant.} & \multicolumn{4}{p{5cm}}{a bird flying over the mountains in the sunset, with the text "bla bla bla, this is the sound of a helicopter"} \\
    & \includegraphics[width=0.075\textwidth]{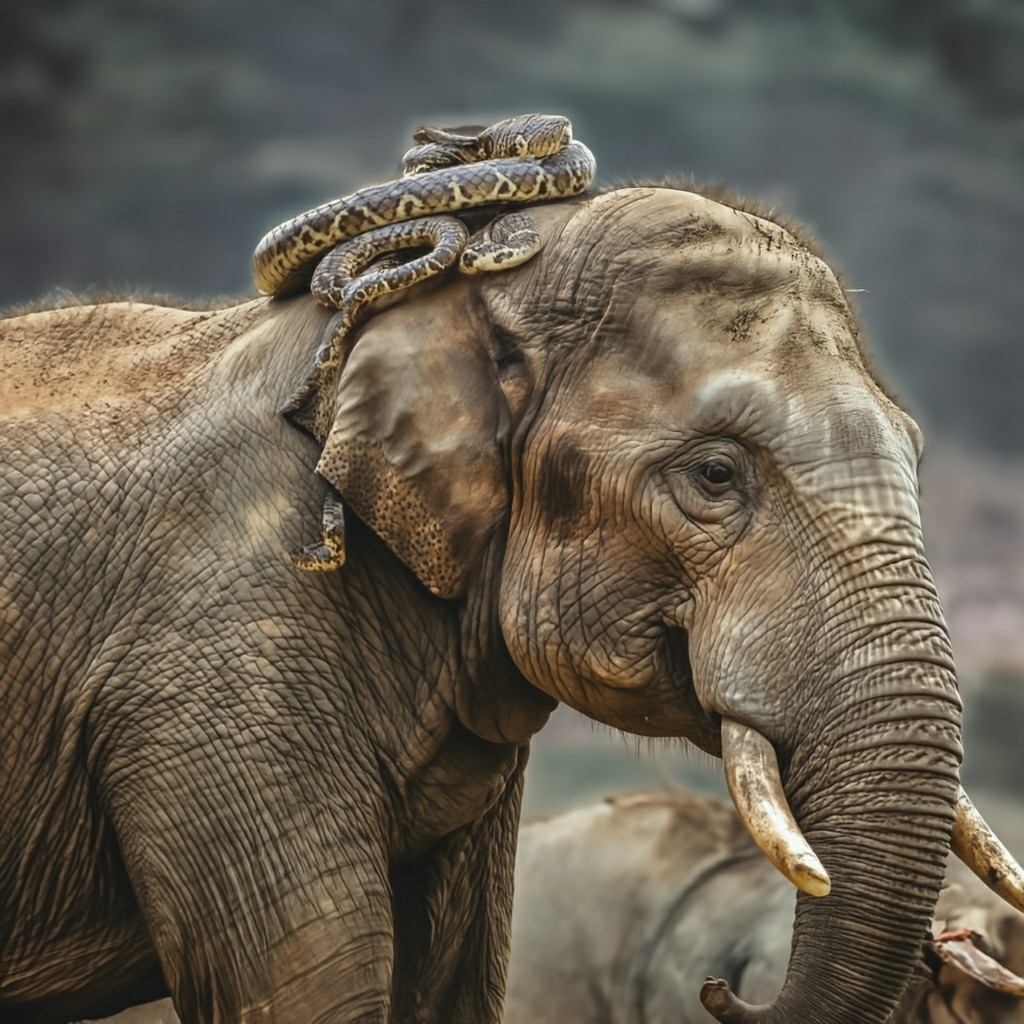} & \includegraphics[width=0.075\textwidth]{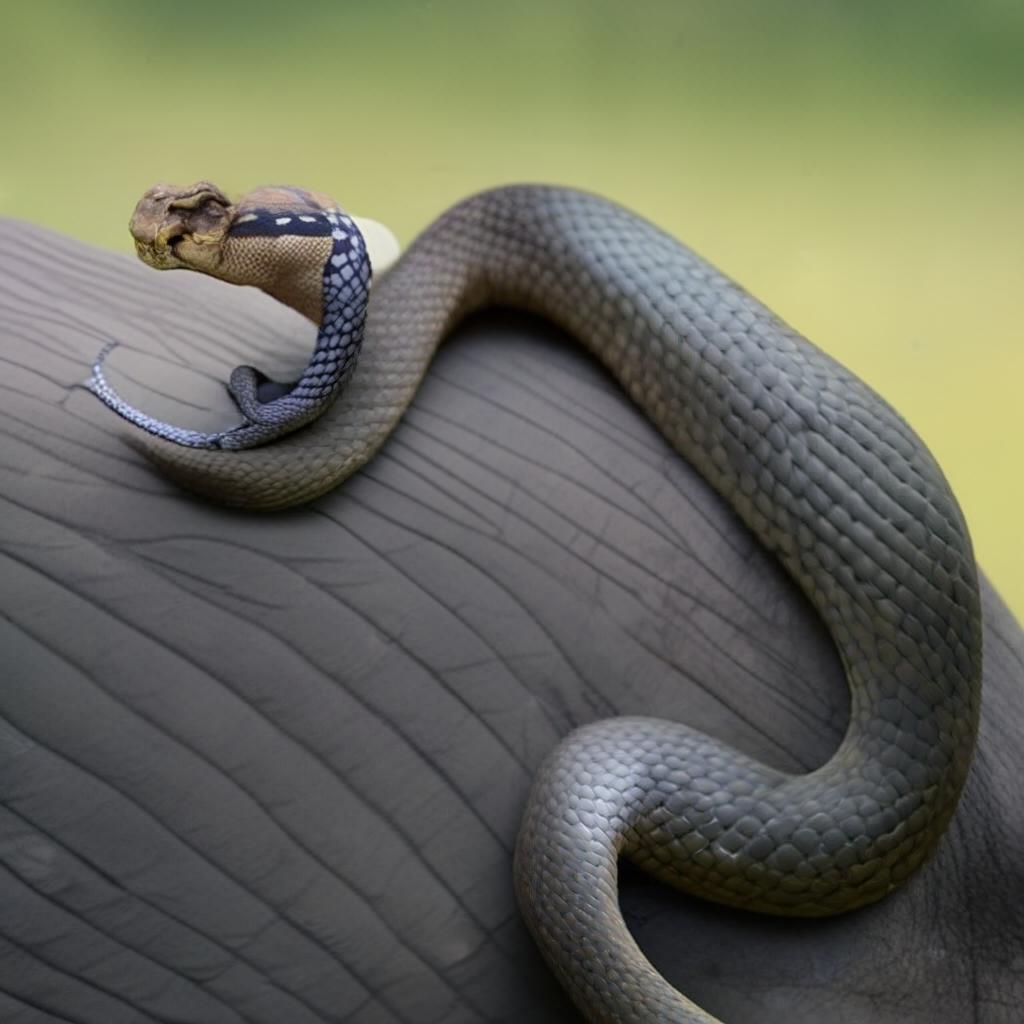}  & \includegraphics[width=0.075\textwidth]{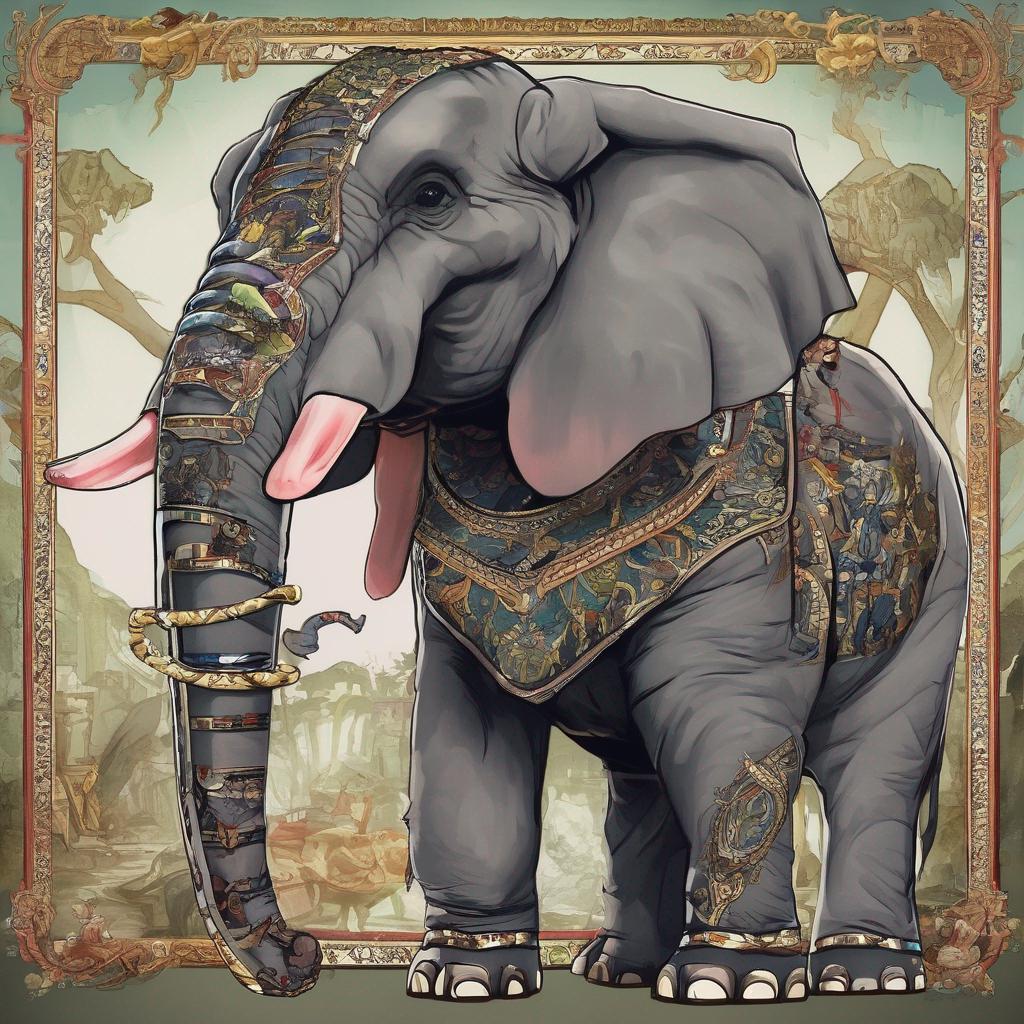} & \includegraphics[width=0.075\textwidth]{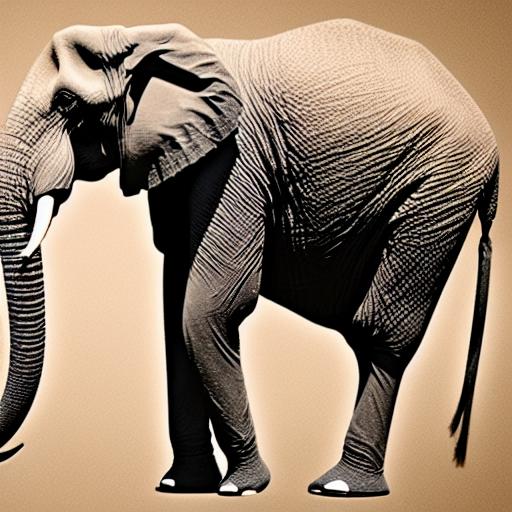} & \includegraphics[width=0.075\textwidth]{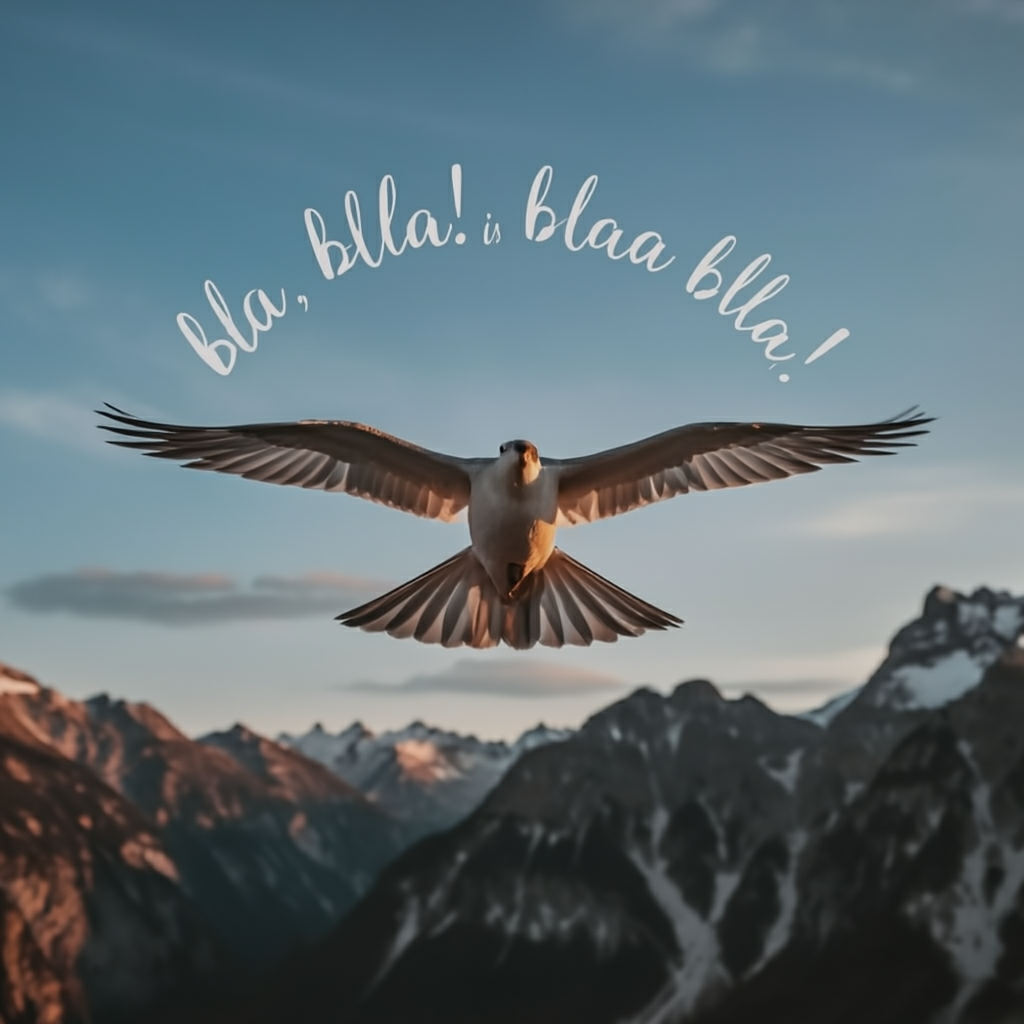} & \includegraphics[width=0.075\textwidth]{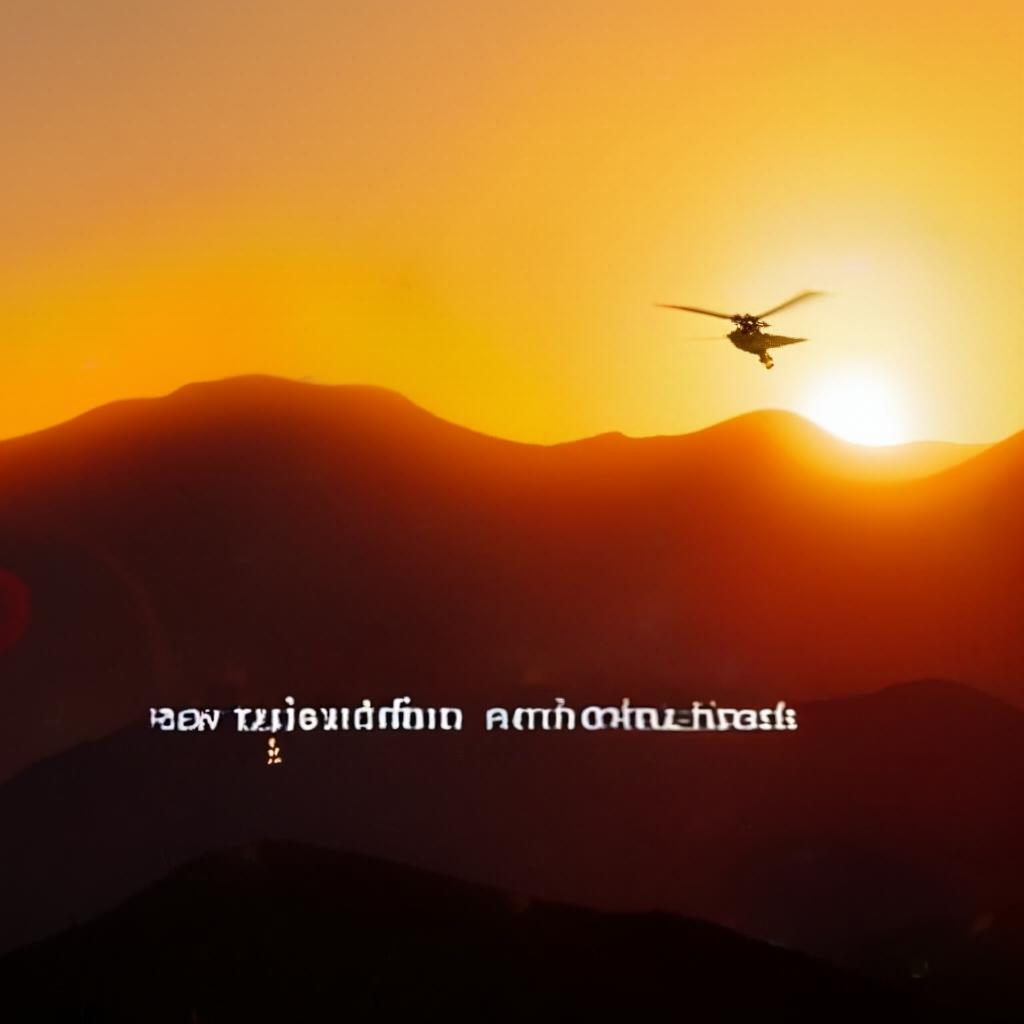}  & \includegraphics[width=0.075\textwidth]{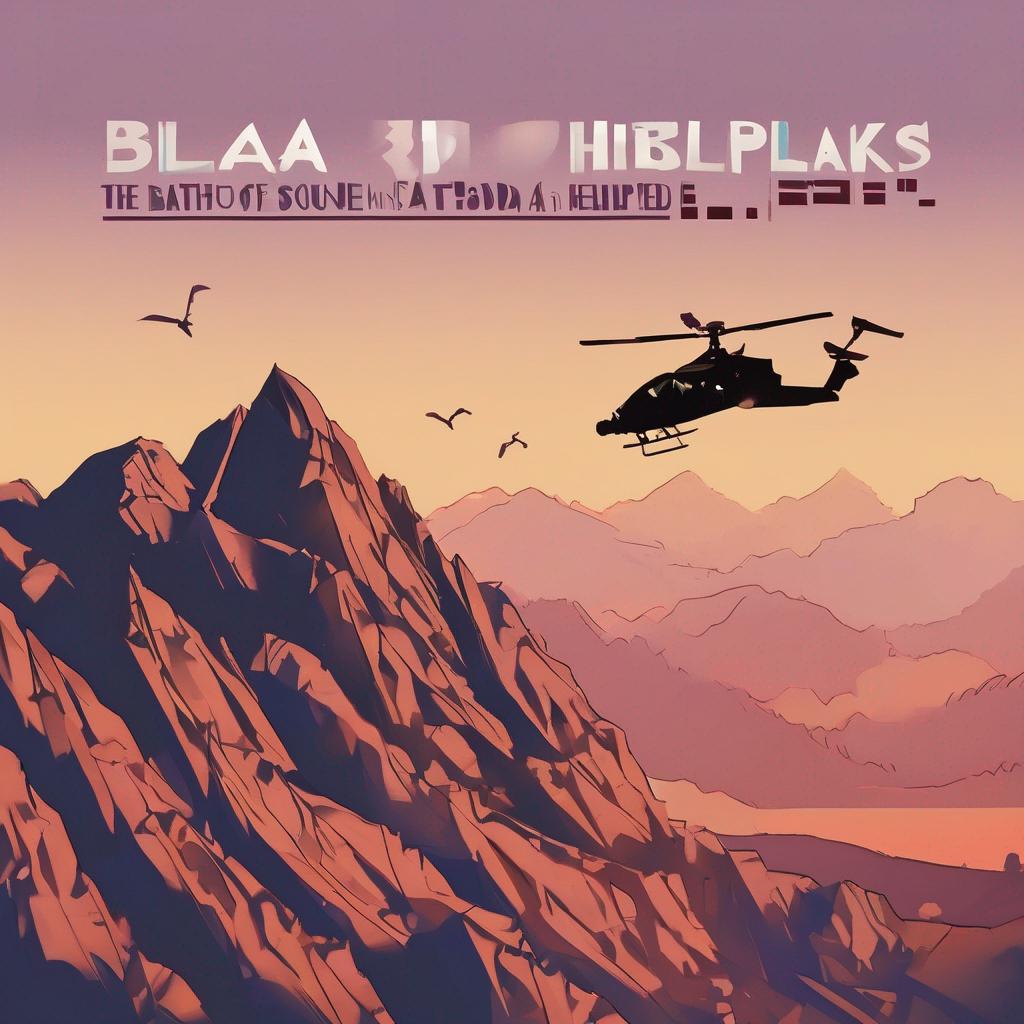} & \includegraphics[width=0.075\textwidth]{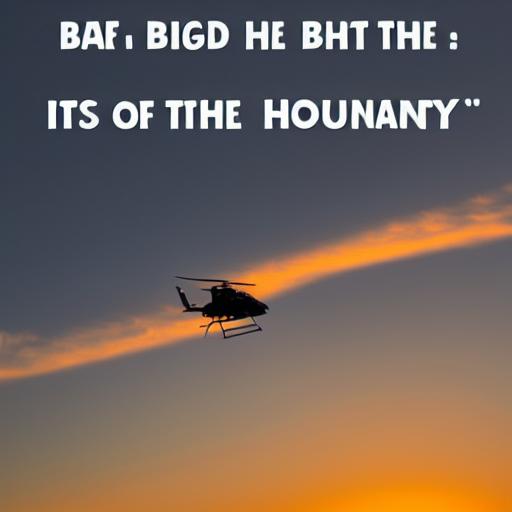} \\
    & {Imagen} & Muse & SDXL & SD1.5 & {Imagen} & Muse & SDXL & SD1.5 \\ \toprule
    APIC: & 1. & 1.0 & 0.44 & 0.33 & 0.53 & 0.83 & 0.47 & 0.21 \\
    Likert: & 1. & 0.80 & 0.6 & 0.4 & 0.6 & 0.6 & 0.6 & 0.4\\
    DSG (H): & 1. & 0.50 & 1. & 0.5 & 0.92 & 0.8 & 0.71 & 0.25 \\ \midrule
    Gecko: & 0.98 & 0.84 & 0.73 & 0.26 & 0.85 & 0.58 & 0.86 & 0.21 \\
    DSG: & 1.0 & 1.0 & 1.0 & 0.33 & 0.67 & 0.50 & 0.89 & 0.11 \\
    VNLI: & 0.94 & 0.93 & 0.14 & 0.23 & 0.19 & 0.30 & 0.34 & 0.27 \\
    \bottomrule
\end{tabular}

\vspace{2em}

\begin{tabular}{c|cccc|cccc}
    {\bf Prompt:} & \multicolumn{4}{p{5cm}}{There are 5 apples, two of them are yellow and two are black but none are red.} & \multicolumn{4}{c}{the dog who wears a white shirt holds a beer} \\
    & \includegraphics[width=0.075\textwidth]{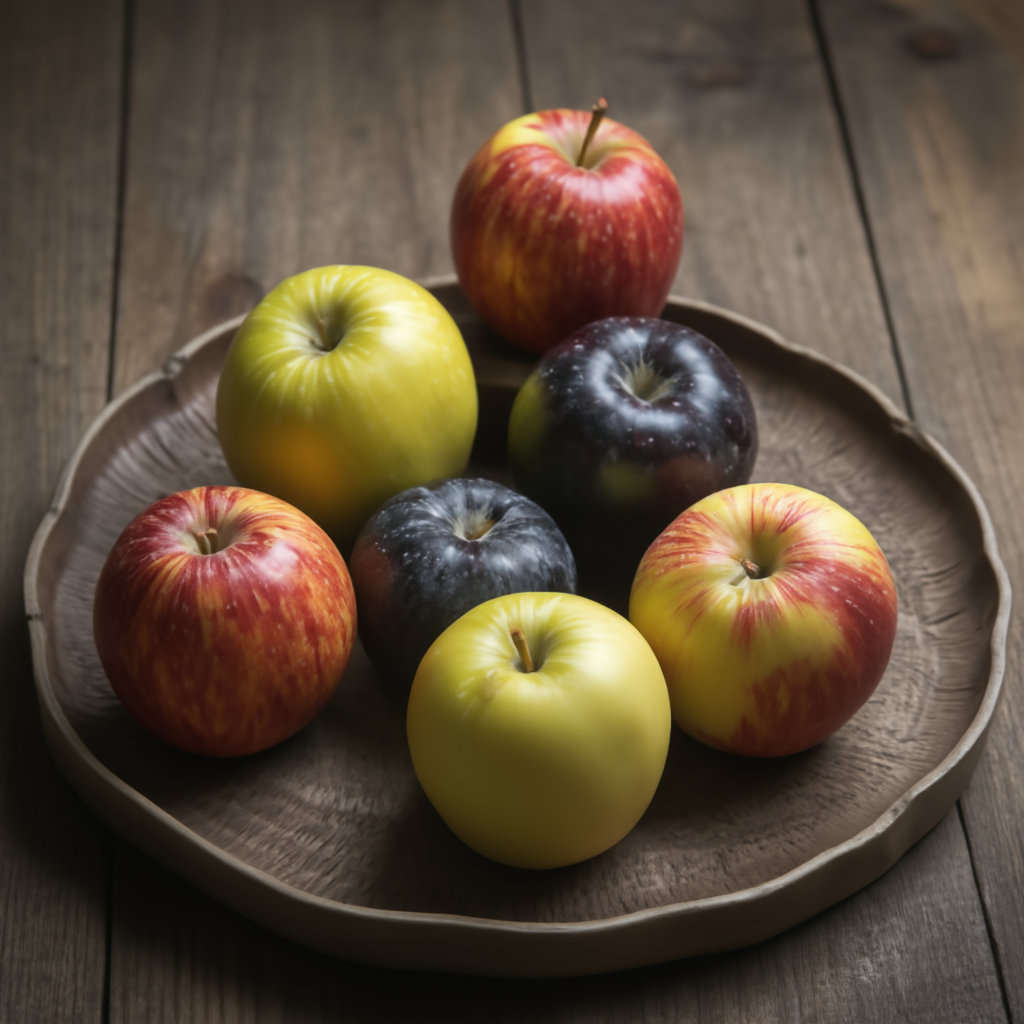} & \includegraphics[width=0.075\textwidth]{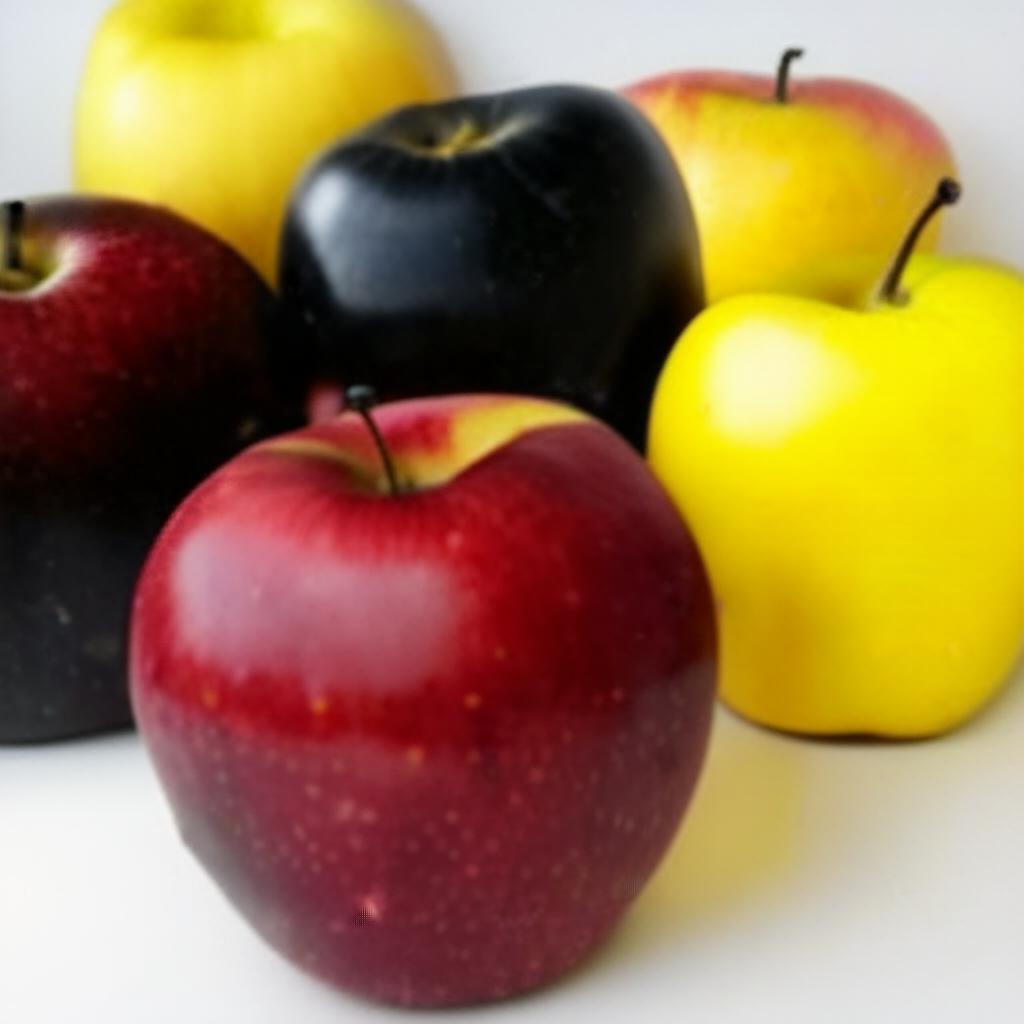}  & \includegraphics[width=0.075\textwidth]{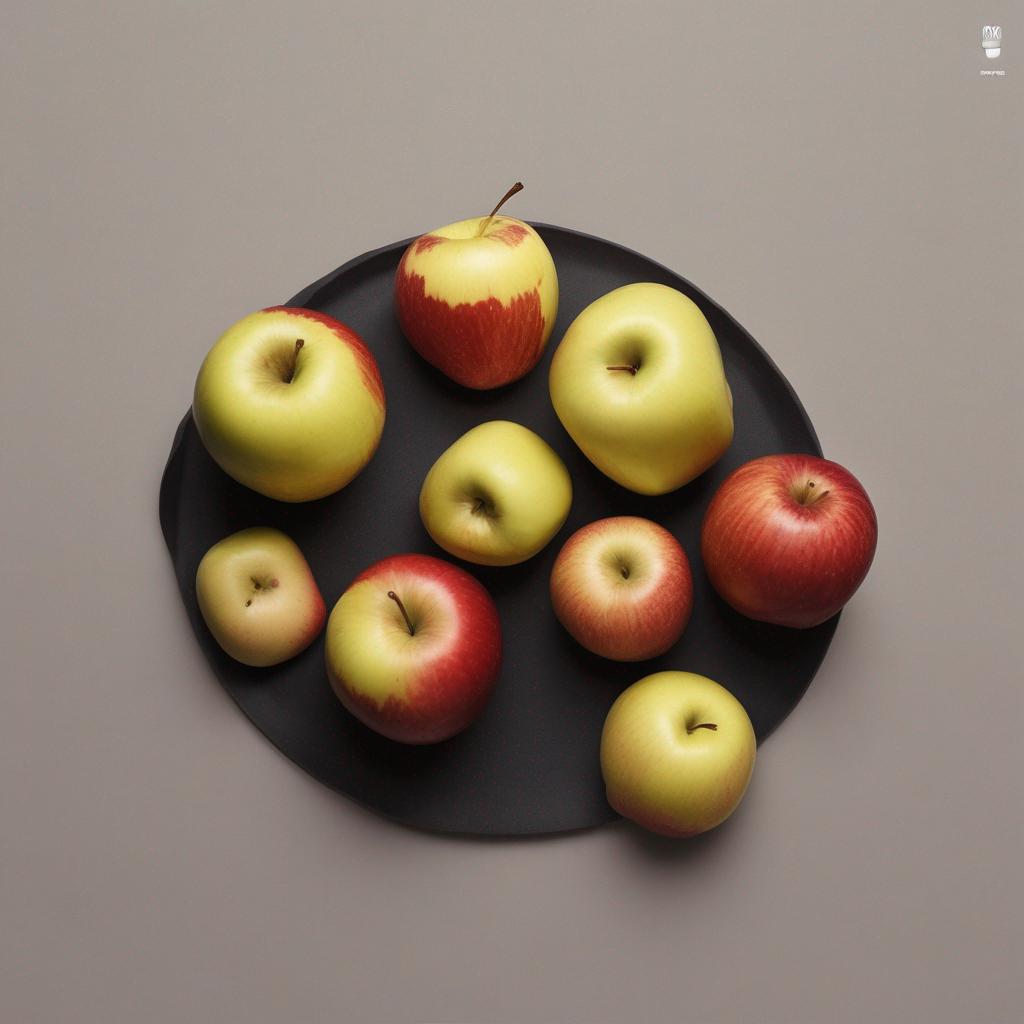} & \includegraphics[width=0.075\textwidth]{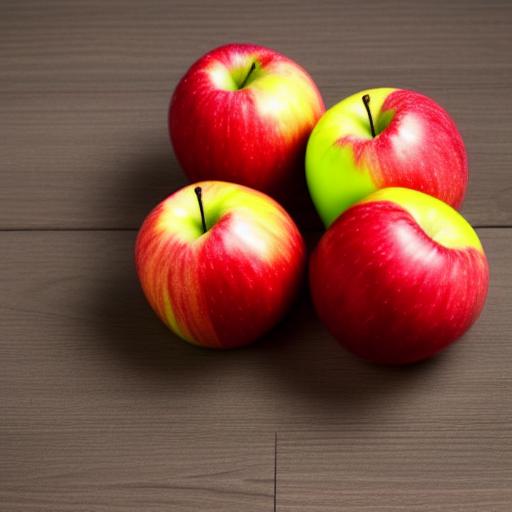} & \includegraphics[width=0.075\textwidth]{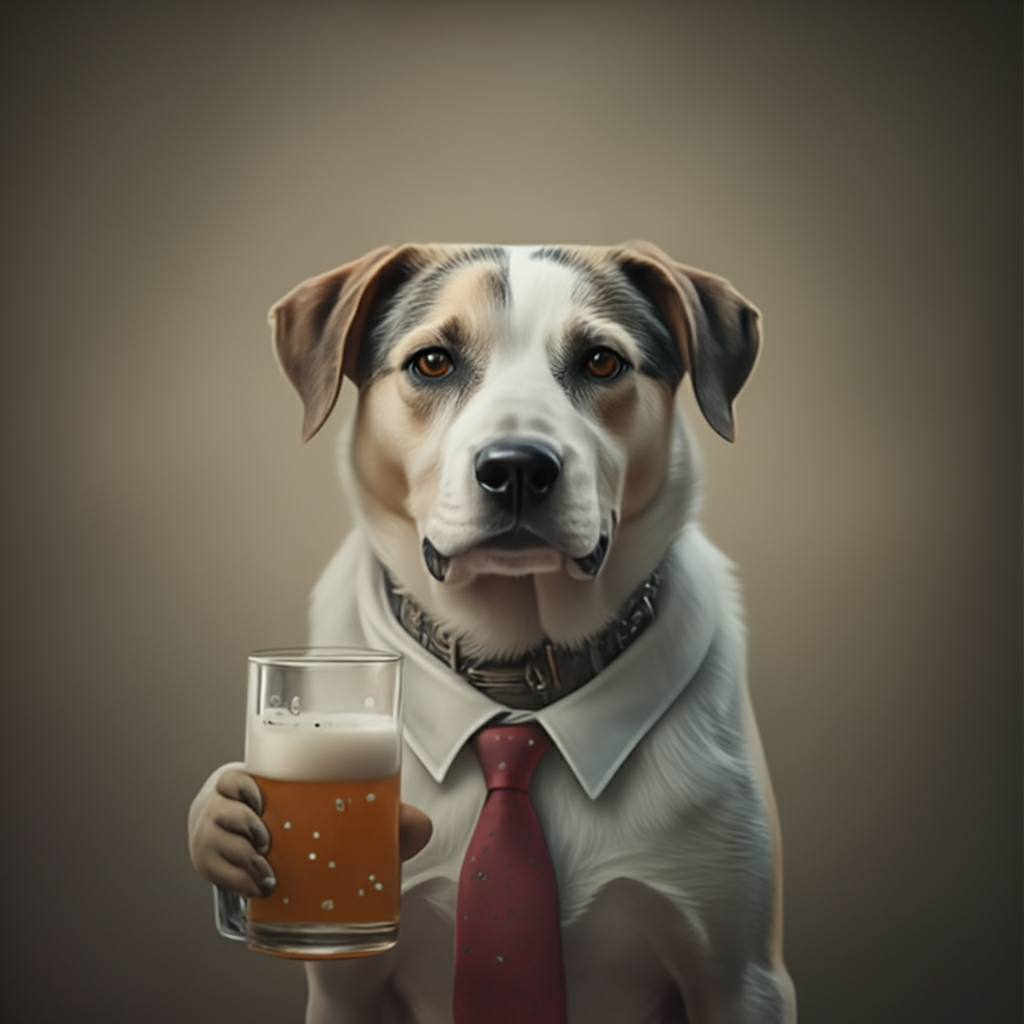} & \includegraphics[width=0.075\textwidth]{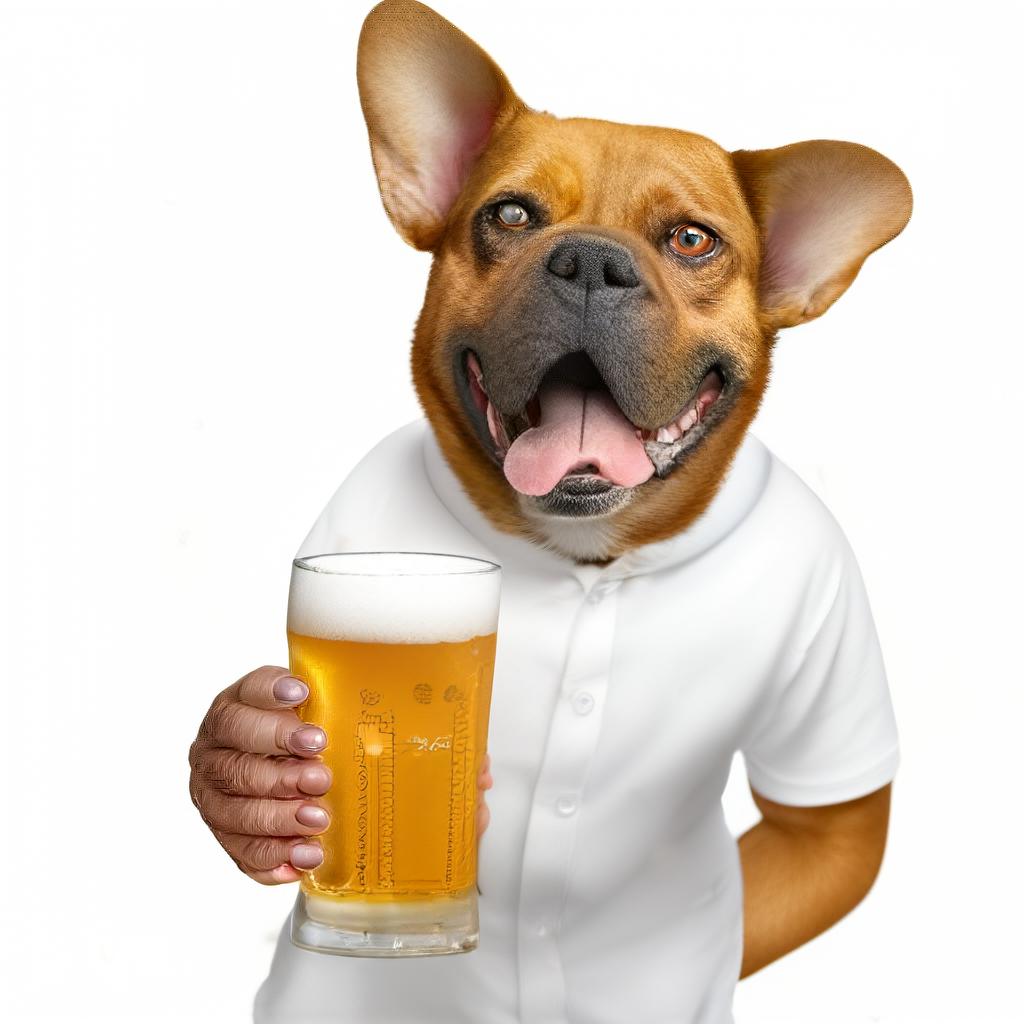}  & \includegraphics[width=0.075\textwidth]{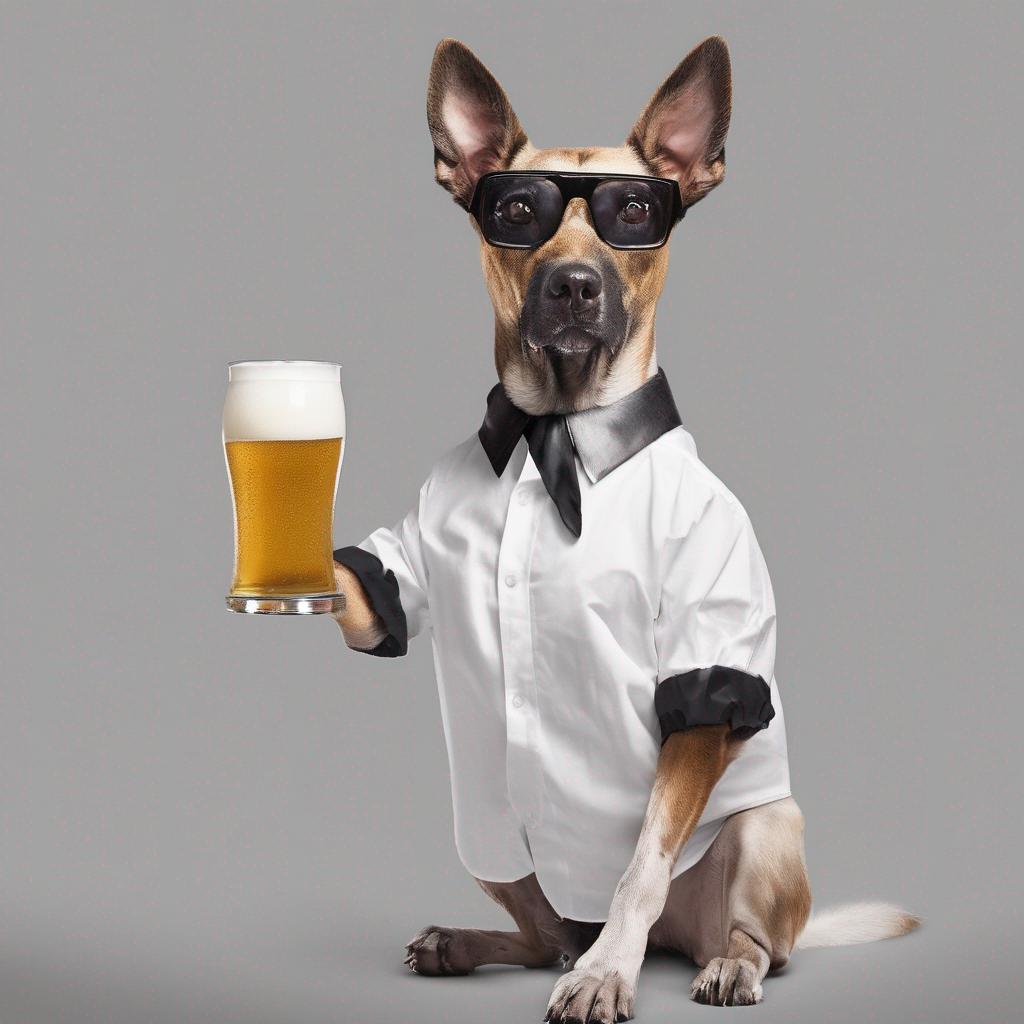} & \includegraphics[width=0.075\textwidth]{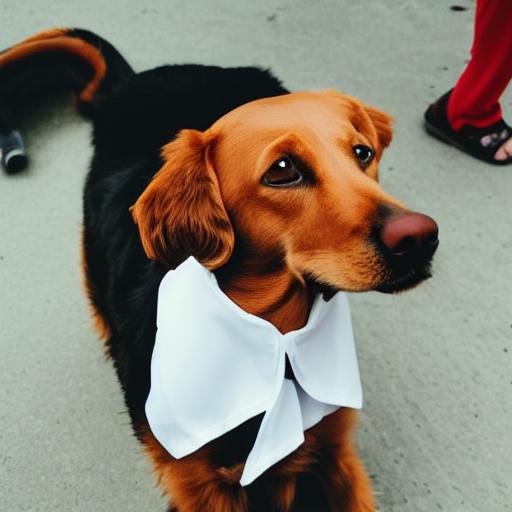} \\
    & {Imagen} & Muse & SDXL & SD1.5 & {Imagen} & Muse & SDXL & SD1.5 \\ \toprule
    APIC: & 0.88 & 0.73 & 0.57 & 0.61 & 1.0 & 1.0 & 1. & 0.4 \\
    Likert: & 0.73 & 0.60 & 0.53 & 0.60 & 0.8 & 0.93 & 1. & 0.53 \\
    DSG (H): & 0.80 & 0.53 & 0.5 & 0.40 &  0.75 & 1.0 & 1. & 0.25 \\ \midrule
    Gecko: & 0.88 & 0.91 & 0.84 & 0.69 & 0.98 & 0.97 & 0.97 & 0.7  \\
    DSG: & 1.0 & 1.0 & 0.9 & 0.8 & 1. & 0.83 & 1. & 0.17 \\
    VNLI: & 0.23 & 0.20 & 0.19 & 0.16 & 0.9 & 0.95 & 0.94 & 0.17 \\
    \bottomrule
\end{tabular}

\vspace{2em}

\begin{tabular}{c|cccc|cccc}
    {\bf Prompt:} & \multicolumn{4}{p{5cm}}{A fortune cookie that has the fortune "the best way to predict the future is to create it."} & \multicolumn{4}{c}{A brown glass salad bowl on a grey metal table.} \\
    & \includegraphics[width=0.075\textwidth]{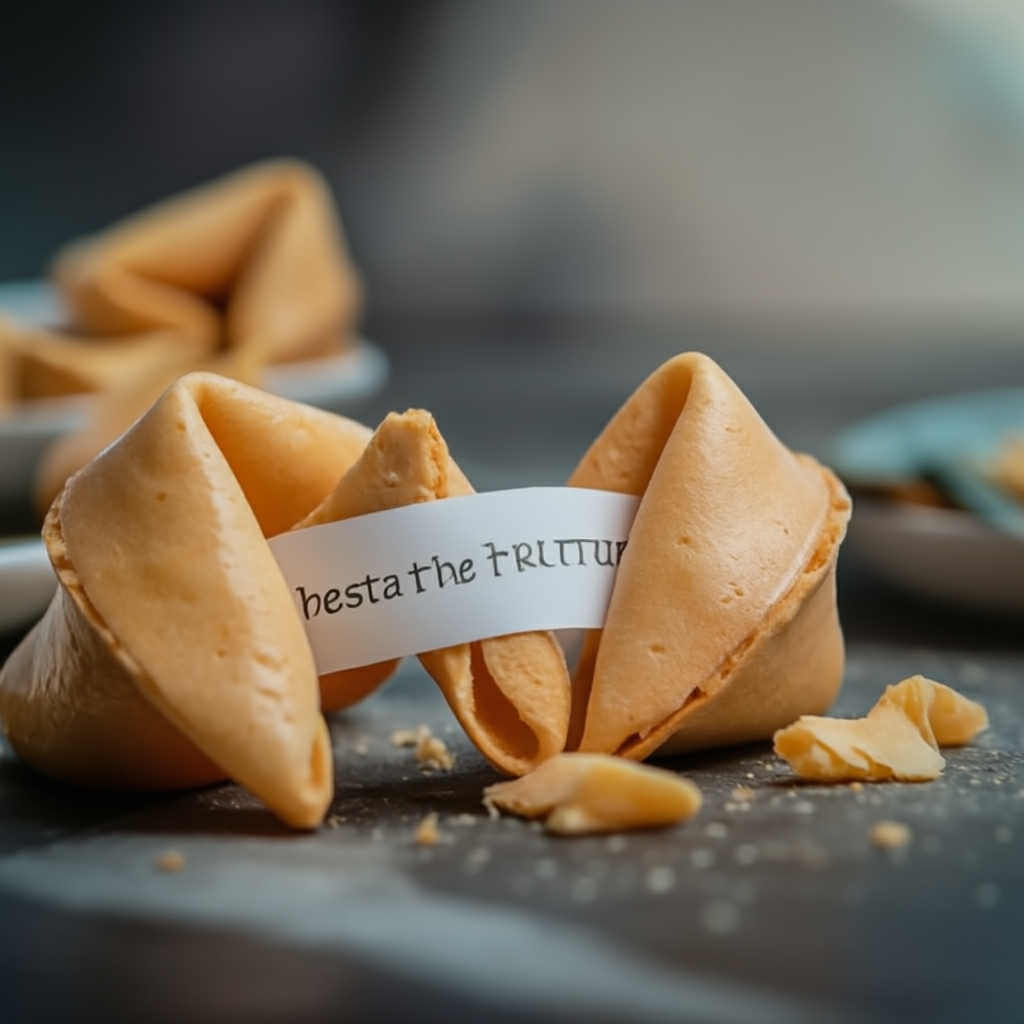} & \includegraphics[width=0.075\textwidth]{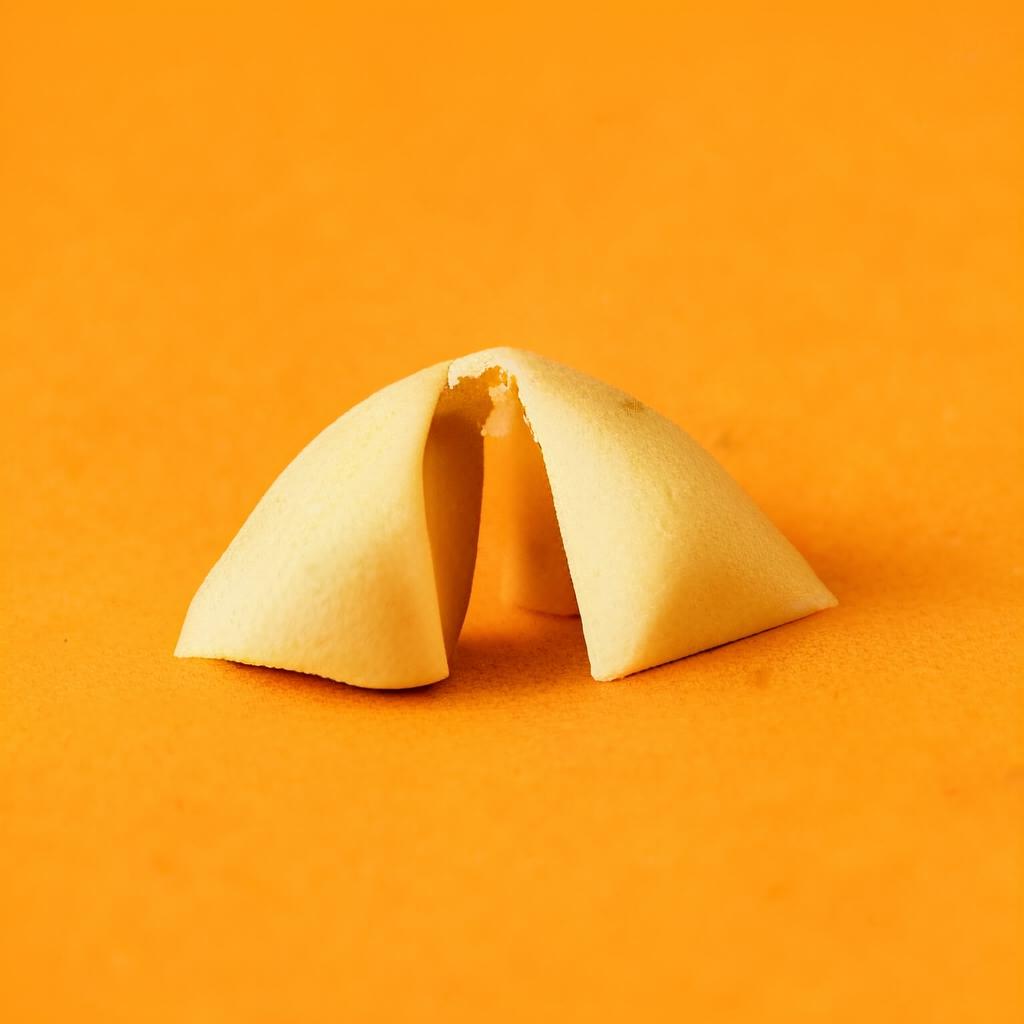}  & \includegraphics[width=0.075\textwidth]{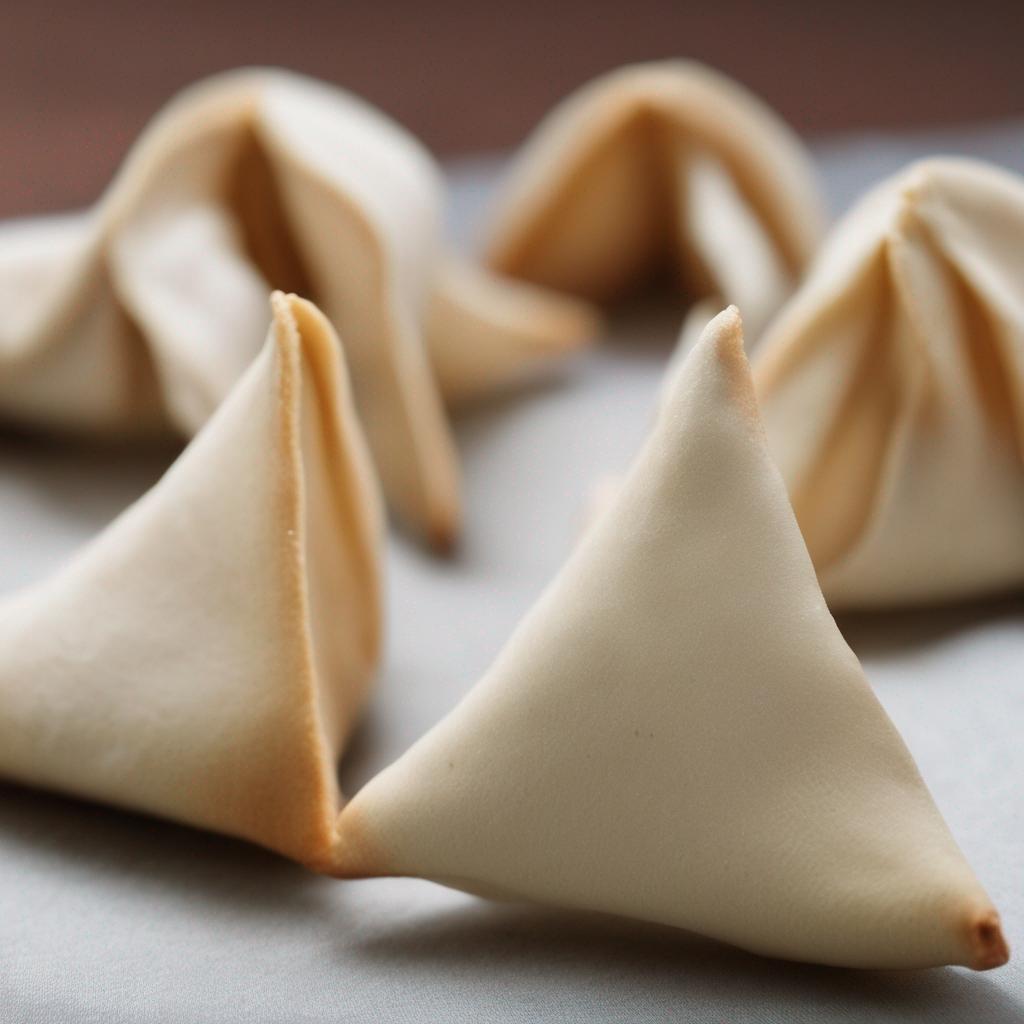} & \includegraphics[width=0.075\textwidth]{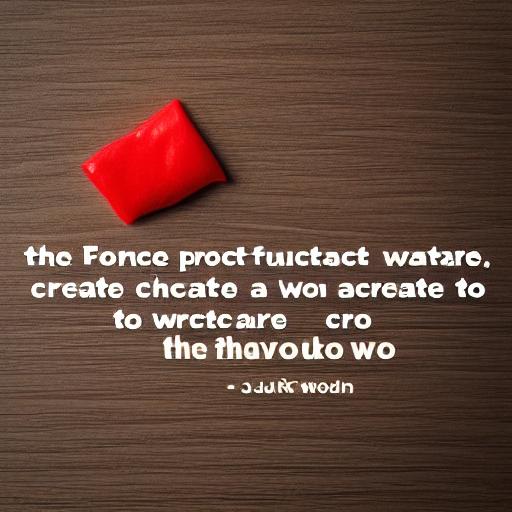} & \includegraphics[width=0.075\textwidth]{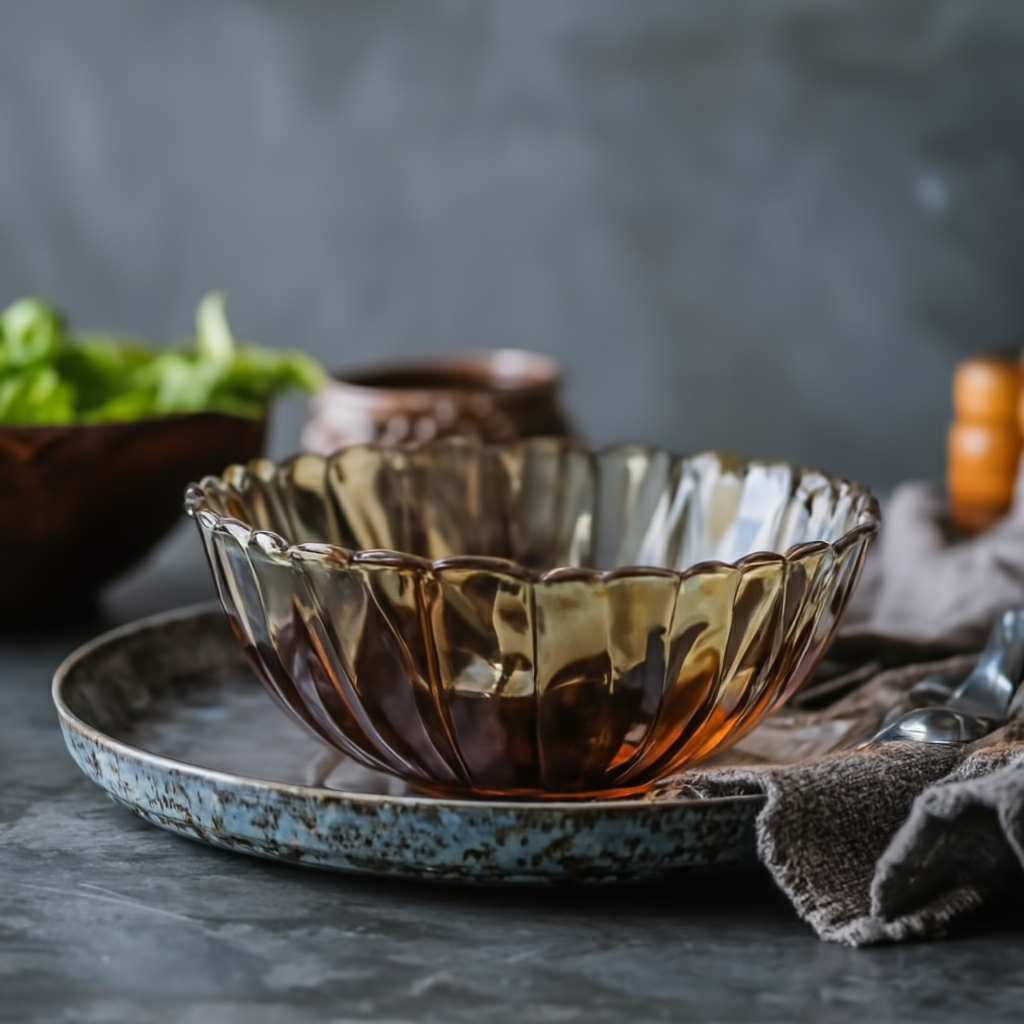} & \includegraphics[width=0.075\textwidth]{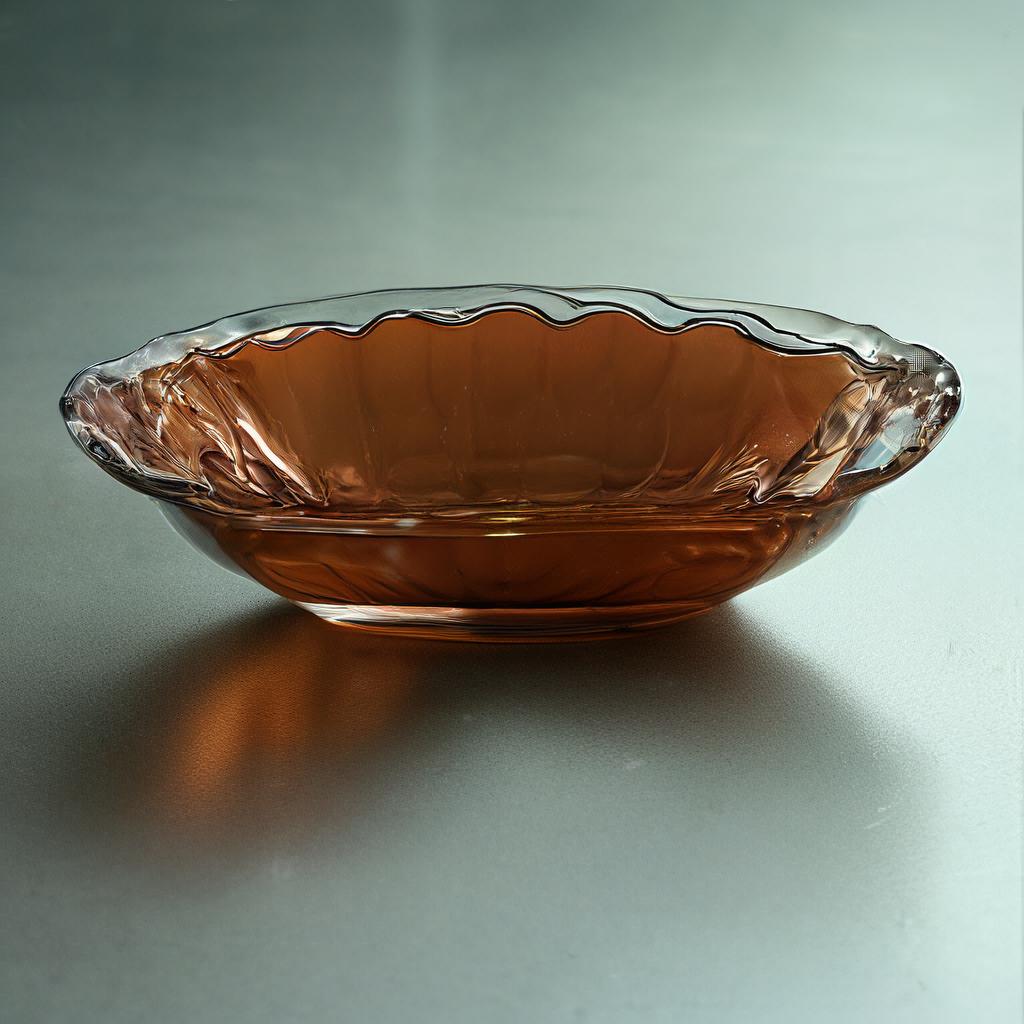}  & \includegraphics[width=0.075\textwidth]{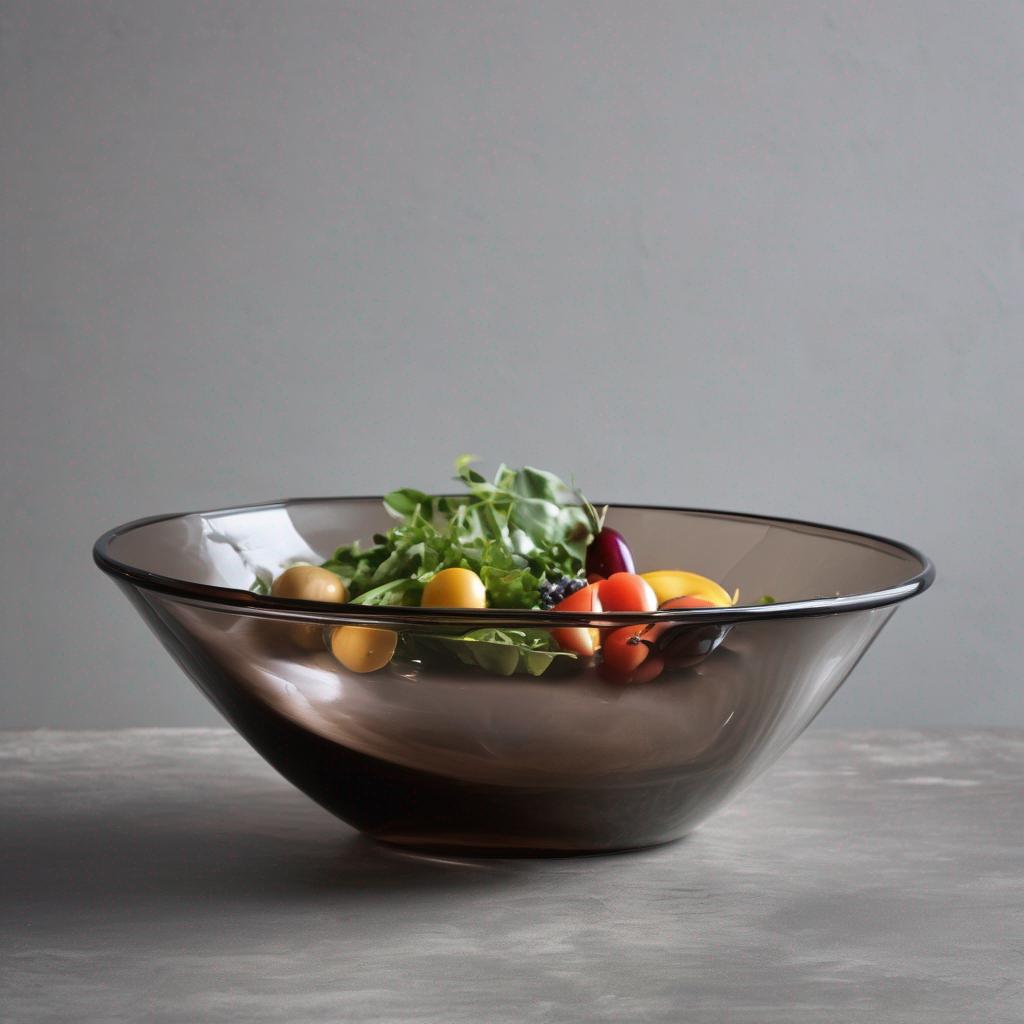} & \includegraphics[width=0.075\textwidth]{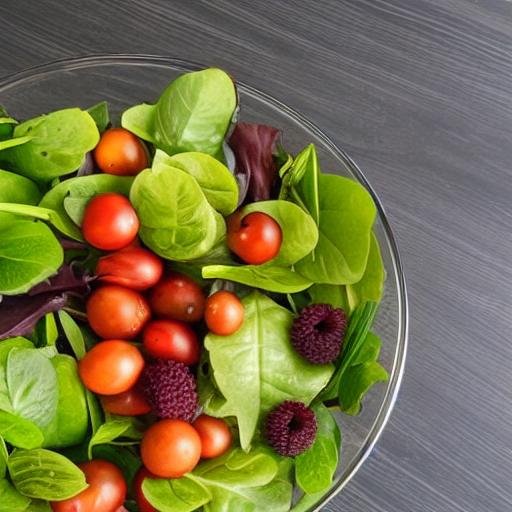} \\
    & {Imagen} & Muse & SDXL & SD1.5 & {Imagen} & Muse & SDXL & SD1.5 \\ \toprule
    APIC: & 0.35 & 0.39 & 0. & 0.02 & 1.0 & 0.88 & 1.0 & 0.63 \\
    Likert: & 0.53 & 0.47 & 0.4 & 0.27 & 0.73 & 0.67 & 0.73 & 0.53 \\
    DSG(H): & 1.0 & 0.5 & 0.67 & 0.17 & .83 &  1.0 & 0.83 & 0.83 \\ \midrule
    Gecko: & 0.96 & 0.79 & 0.75 & 0.62 & 0.92 & 0.88 & 0.87 & 0.70 \\
    DSG: & 0.88 & 0.25 & 1.0 & 0.25 & 0.93 & 0.93 & 0.86 & 0.79 \\
    VNLI: & 0.37 & 0.27 & 0.25 & 0.35 & 0.81 & 0.65 & 0.55 & 0.31  \\
    \bottomrule
\end{tabular}
\caption{\textbf{Additional visualisations of scores from different auto-eval metrics.} We show the image generations by the four generative models given two prompts from Gecko(S), with the alignment scores predicted by human annotators and auto-eval metrics respectively.}

\label{app:additionalresultssynthetic}
\end{figure}

We visualise additional examples in \Figure{app:additionalresultssynthetic} for different categories (spatial, counting, text rendering, linguistic complexity, etc.).
These examples demonstrate both differences arising from different annotation templates and also different metrics.

\subsection{Results per Word for WL}
\label{app:results_per_word}
Here we evaluate how well the Gecko metric (with the PALI/PALM-2 backend) can identify whether words are or are not grounded as rated by WL. This experiment can {\em only} be done for Gecko as {\em no other} metric gives word level annotations that can be traced back to the original words in the prompt. We note that this is much more challenging than giving an overall image rating. In order to perform this experiment, we first parse the coverage prediction to ensure we can match words in the original prompt with those in the coverage prediction.
For example, if we have the original prompt `a red-colored dog' and `a \{1\}[red-colored] \{2\}[dog]' as the generated coverage one, we can map from the word index (\eg `\{1\}' to the phrase `red-colored'). 

We then take all word level predictions where all annotators either annotated the word as grounded or not grounded (we removed those for which a subset of annotators annotated the word as `unsure'). 
For these words, we take the ones where the coverage model predicts that it should be covered (\eg in the example above, even if `a' was always annotated as grounded, we would ignore it as it is not considered groundable by the coverage step).
Given this final set of words, we look at whether the VQA prediction was accurate and compare this to whether the annotators thought that the word was grounded or not.

We report three numbers in \Table{app:table_word_level_results}: (1) the number of words we are left with, (2) the accuracy and (3) the error consistency $\kappa$ \citep{geirhos2020beyond}, equation (1),(3). We report error consistency as many of the words ($\sim90\%$) are rated as grounded. Accuracy does not account for the fact that a metric which predicts `grounded' $\sim90\%$ of the time would actually get $\sim80\%$ accuracy by chance. Error consistency takes this into account such that $\kappa=-1$ means that two sets of results never agree, $\kappa=0$ that the overlap is explained by chance and $\kappa=1$ means results agree perfectly.
As shown by the results, \model~is able to predict grounding at the word level with reasonable accuracy. Moreover, results are not simply explained by chance (as $\kappa > 0.$); the error consistency results indicate in general some but not substantial agreement.

\begin{table}[t]
\vspace{-6mm}
\centering
\tiny
\begin{tabular}{cP{1.5cm}cP{1.5cm}cP{1.5cm}cP{1.5cm}} \toprule
     & \multicolumn{3}{c}{\textbf{Gecko(R)}} &  \quad \quad & \multicolumn{3}{c}{\textbf{Gecko(S)}} \\ \midrule
     & \# Words evaluated & Accuracy & Error Consistency ($\kappa$) &  \quad \quad  & \# Words  evaluated & Accuracy & Error Consistency ($\kappa$) \\ \midrule
    SD1.5 & 3727 & 75.5 & 0.20 & & 2729 & 77.5 & 0.52 \\ 
    SDXL & 3901 & 80.0 & 0.13 & & 3304 & 79.5 & 0.34\\
    Muse & 2792 & 82.5 & 0.24 & & 3298 & 82.0 & 0.19 \\
    Imagen & 2549 & 76.5 & 0.29 & & 3472 & 80.2 & 0.39 \\ \bottomrule
\end{tabular}
\caption{{\bf Word level results comparing Gecko to the WL annotation template.} We can see that Gecko achieves high accuracy but also that results are not simply explained by chance, as $\kappa > 0$ and in general indicates fair agreement.}
\label{app:table_word_level_results}
\end{table}

\section{Extending Gecko to More Modalities:\\ Text-to-Video Generation}
\label{app:video_results}



Video evaluations are more challenging, in that there are multiple aspects that are encapsulated in the text-to-video consistency, including stylistic, temporal, semantic and overall fidelity. In order to extend our approach, we follow a similar two step process, in which question-answer pairs are generated using a few-shot prompt that outlines which aspects of the video need to be covered by those pairs. The corresponding few-shot prompt used for question-answer pair generation with sufficient coverage of the groundable words in the prompt is shown in~\Listing{alg:few_shot_prompt_llm_qa_video}.

\begin{lstlisting}[language=Python,caption={Sample LLM template for generating QAs with coverage for videos.},label={alg:few_shot_prompt_llm_qa_video}]
""" 
Given a video description and the groundable words in it, generate multiple-choice questions that verify if the video description is correct.
The goal is to ask questions about entities, objects, attributes, actions, colors, spatial relations, temporal relations, styles and scenes, when these are present in the description.
Make sure that all options are substantially different from each other and only one option can be the correct one based on the description. Do not include other parts of the description as a non correct option.
Justify why the other options cannot be true based on the description and question. Also, make sure that the question cannot be answered correctly only based on common sense and without reading the description.
Each generated question should be independent of the other ones and it should be able to be understood without knowing the other questions; avoid refering to entities/objects/places from previous questions.
Finally, avoid asking very general questions, such as 'What is in the video?', or 'Name a character in the video'.
Generate the multiple-choice questions in the exact same format as the examples that follow. Do not add asterisks, white spaces, or any other reformatting and explanation that deviate from the formatting of the following examples.

Description:
A fat rabbit wearing a purple robe walking through a fantasy landscape.
The visual-groundable words and their scores are labelled below:
A {1}[fat, attribute, 1.0] {2}[rabbit, entity, 1.0] {3}[wearing a {4}[purple, color, 1.0] robe, attribute, 1.0] {5}[walking, action, 1.0] through a {6}[fantasy landscape, scene, 1.0].
Generated questions and answers are below:
About {1}:
Q: What is the most appropriate description for the animal of the video?
Choices: thin, regular, slim, fat
A: fat
Justification: the rabbit in the video is fat ({1}). The options thin and slim are oposite of the attribute mentioned in the description and the regular adjective checks whether it is obvious that the rabbit has a weight above normal.
About {2}:
Q: Who wears a robe in the video?
Choices: rabbit, hare, squirrel, rat
A: rabbit
Justification: the rabbit is the animal that wears a robe in the video ({2}). Hare is an animal very similar to rabbit, and the other two options (squirrel and rat) are also similar but not true according to the description.
About {3}:
Q: What is the rabbit wearing in the video?
Choices: nothing, dress, robe, jumpsuit
A: robe
Justification: the rabbit is wearing a robe ({3}). Nothing is what normally an animal is wearing, and the options dress and jumpsuit are similar to the robe but not true according to the description.
About {4}:
Q: What is the color of the clothing that the rabbit wears in the video?
Choices: purple, blue, pink, green
A: purple
Justification: the rabbit is wearing a purple robe ({4}). the options blue, pink and green are colors similar to purple.
About {6}:
Q: What is the rabbit doing in the video?
Choices: running, walking, standing, jumping
A: walking
Justification: the rabbit is walking through a fantasy landscape ({5}, {6}). The options running and standing are similar to walking, and jumping is an action that could be performed by a rabbit, but not true according to the description.
About {7}:
Q: Where is the video taking place?
Choices: fields, countryside, fantasy landscape, mountains
A: fantasy landscape
Justification: the rabbit is walking through a fantasy landscape ({6}). The options fields, countryside, and mountains are different types of landscapes, but they are real-world scenes instead of fantasy ones.

Description:
A beautiful coastal beach in spring, waves lapping on sand by Hokusai, in the style of Ukiyo
The visual-groundable words and their scores are labelled below:
A {1}[beautiful coastal beach, scene, 1.0] {2}[in spring, temporal relation, 1.0], {3}[waves, scene, 1.0] {4}[lapping, action, 1.0] {5}[on sand, spatial relation, 1.0] {6}[by Hokusai, style, 1.0], {7}[in the style of Ukiyo, style, 1.0]
Generated questions and answers are below:
About {1}:
Q: Where is the video taking place?
Choices: cliffs, harbor, coastal park, coastal beach
A: coast beach
Justification: the main scene is a beautiful coastal beach ({1}). The options cliffs, harbor, and coastal park are similar to coastal beach but not true according to the description.
About {2}:
Q: Which season is most likely during the video?
Choices: spring, summer, autumn, winter
A: spring
Justification: the video shows a coastal beach in spring ({2}). The options summer, autumn and winter are other seasons that are not true according to the description.
About {3}:
Q: What is the level of movement of the sea during the video?
Choices: calm, wavy, slightly moving, ripply
A: wavy
Justification: the sea is wavy ({3}). The options calm, slightly moving, and ripply are different levels of movement of the sea and they are all different enough from wavy.
About {4}:
Q: What is the movement of the sea during the video?
Choices: gentle waves are coming to the shore, there is a tide, waves are lapping on the shore, there are sea ripples
A: waves are lapping on the shore
Justification: the sea is lapping on the shore ({4}). The other provided options are either of less intesity (gentle waves are coming to the shore, there are sea ripples) or the exact opposite (there is a tide).
About {5}:
Q: Where does the sea move to during the video?
Choices: sand, rocks, cliffs, pebbles
A: sand
Justification: the waves are lapping on sand ({5}). The options pebbles, rocks, and cliffs are different types of ground typically by the sea and have different levels of solidity.
About {6}:
Q: Whose artist is the theme of the scene similar to?
Choices: Utamaro, Hokusai, Hiroshige, Yoshitoshi
A: Hokusai
Justification: the theme of the scene resembles a painting of Hokusai. The other options are other Japanese artists that are similar to Hokusai.
About {7}:
Q: Which Japanese painting style is most similar to the video?
Choices: Ukiyo, Nihonga, Sumi, ink calligraphy
A: Ukiyo
Justification: the video scene is in the style of Ukiyo ({7}). The other options are other types of Japanese painting styles that are not similar to the video according to the description.

Description:
...
"""

\end{lstlisting}

{\bf Evaluation setup.} As described in Section~\ref{main:video_results}, we choose a prompt set that is appropriate for measuring alignment between the description and the video and a set of text-to-video models to collect human annotations using different templates. We consider a subset of 94 prompts from the VBench benchmark~\citep{huang2024vbench} manually tagged with ``overall consistency'' for evaluating overall text-to-video alignment by the curators of the benchmark.
We compare the following text-to-video models: Lumiere~\citep{bar2024lumiere}, Phenaki~\citep{villegas2022phenaki} and WALT~\citep{gupta2023photorealistic}. For human evaluation, we consider both absolute comparison templates (i.e., Likert and Word Level) and a template for relative pairwise comparisons (i.e., SxS), and for automatic evaluation, we again benchmark two types of auto-eval metrics: contrastive models (i.e., VideoCLIP;~\citealt{xu2021videoclip}) and VQA-based metrics. For VQA-based metrics, since there is no prior work on text-to-video generation, we extend the VQAScore and our fine-grained Gecko metric on the video domain using Gemini Flash as our video question answering model, which can process long context multimodal inputs.

\begin{figure}[t]
\scriptsize
\begin{tikzpicture}
  \matrix(D)[matrix of nodes,nodes in empty cells,
             row sep=-\pgflinewidth,column sep=-\pgflinewidth,
             nodes={anchor=center},
             row 1/.style={minimum height=0.5cm},
             row 2/.style={minimum height=0.5cm},
             row 3/.style={minimum height=0.5cm},
             row 4/.style={minimum height=0.5cm},
             column 1/.style={minimum width=1.2cm},
             column 2/.style={minimum width=1.2cm},
             column 3/.style={minimum width=1.2cm},
             column 4/.style={minimum width=1.6cm},
             column 5/.style={minimum width=1.2cm},
             column 6/.style={minimum width=1.6cm},
             column 7/.style={minimum width=1.6cm},
             column 8/.style={minimum width=1.6cm},
            ]
  {%
    {\bf Model 1} & {\bf Model 2}  & & `GT' & & \model & VideoCLIP & VQAScore \\
    Lumiere & Phenaki & &  |[draw,fill=green!20]| $>$ & & |[draw,fill=green!20]| $>$ & |[draw,fill=yellow!20]| $--$ & |[draw,fill=yellow!20]| $--$  \\
    Lumiere & WALT & &  |[draw,fill=green!20]| $>$ & & |[draw,fill=green!20]| $>$ & |[draw,fill=yellow!20]| $--$ & |[draw,fill=green!20]| $>$  \\
    Phenaki & WALT & &  |[draw,fill=yellow!20]| $--$ & & |[draw,fill=yellow!20]| $--$ & |[draw,fill=yellow!20]| $--$ & |[draw,fill=green!20]| $>$  \\
  };
\end{tikzpicture} 
\caption{\textbf{Comparing model ordering obtained from humans and auto-eval metrics on VBench.} We show the `GT' human ordering and the predicted ones for auto-eval metrics. $<$ means Model 1 $<$ Model 2, $>$ Model 1 $>$ Model 2 and $--$ that no significant relation was found.}
\label{tab:app:video_modelorderingresults}
\end{figure}

{\bf \Paircompare.} In addition to the \Pairscore and \pointscore~results presented in Table~\ref{tab:corr_pearson_video}, we also compare model rankings per metric in Figure~\ref{tab:app:video_modelorderingresults} using the Wilcoxon signed-rank test for all pairs of models. To obtain the ground-truth (GT) model ordering, we average human preferences across the three templates (Likert, WL, SxS) via majority voting and consider valid model rankings only when pairwise differences are significant (p$<$0.05).
We find that the model ranking provided by Gecko agrees with the human rankings for all three model comparisons. In contrast, VideoCLIP, which is often used as part of tool use for text-to-video evaluation~\citep{huang2024vbench,liu2024evalcrafter}, does not provide {\em any} information about the relative performance of video models in terms of alignment. VQAScore agrees with the ground truth comparisons only for one model pair (Lumiere vs WALT).

\clearpage

\end{document}